\documentclass[10pt]{article}

\usepackage{xcolor}
\PassOptionsToPackage{numbers,sort&compress}{natbib}
 \usepackage[preprint]{tmlr}
\setcitestyle{numbers,square,comma}

\usepackage{microtype}
\usepackage{graphicx}
\usepackage{subcaption}

\usepackage{float}
\usepackage{booktabs} 
\usepackage{fontawesome5}
\usepackage{xcolor}
\usepackage{adjustbox}
\usepackage{tabularx}
\usepackage{array}
\usepackage[margin=1.5cm]{geometry}
\usepackage{setspace}   
\usepackage{tikz}
\usepackage{array}
\usepackage{multirow}
\usepackage{colortbl}
\usepackage{booktabs}
\usepackage{amssymb}
\usepackage{graphicx}
\usepackage{makecell}

\definecolor{gpink}{HTML}{D81B8C}\definecolor{gpinkf}{HTML}{FCEBF4}
\definecolor{gteal}{HTML}{1E8E82}\definecolor{gtealf}{HTML}{E6F5F3}
\definecolor{gorange}{HTML}{E2760B}\definecolor{gorangef}{HTML}{FDF1E1}
\definecolor{gpurple}{HTML}{6C45B0}\definecolor{gpurplef}{HTML}{F0EBFA}
\definecolor{ggreen}{HTML}{3B9A40}\definecolor{ggreenf}{HTML}{E9F5EA}
\definecolor{grootc}{HTML}{37474F}\definecolor{grootf}{HTML}{ECEFF1}

\definecolor{citeblue}{RGB}{20,50,180}
\usepackage[colorlinks=true,
            citecolor=citeblue,
            linkcolor=black,
            urlcolor=black,
            filecolor=black]{hyperref}
\usepackage[figuresleft]{rotating}

\definecolor{hdrbg}{RGB}{245,245,245}

\newcommand{\yes}[1]{\ensuremath{\checkmark}}

\usepackage{amsmath}
\usepackage{mathtools}
\usepackage{amsthm}
\usepackage{tcolorbox}

\usepackage[utf8]{inputenc} 
\usepackage[T1]{fontenc}    
\usepackage{url}            
\usepackage{amsfonts}       
\usepackage{nicefrac}       
\usepackage{multirow} 
\usepackage[capitalize,noabbrev,nameinlink]{cleveref}

\usetikzlibrary{positioning,calc}
\usetikzlibrary{shapes.arrows,shapes.geometric,positioning,fit}
\usetikzlibrary{arrows.meta,backgrounds}
\definecolor{resblue}{RGB}{30,60,150}
\definecolor{despurple}{RGB}{100,30,140}
\definecolor{buildgreen}{RGB}{40,110,60}
\definecolor{lightgreen}{RGB}{240,247,240}

\newcommand{\fgcitewithyear}[1]{\citeauthor{#1}~(\citeyear{#1})}
\newcommand{\customcite}[2]{%
  {\scriptsize
    \defcitealias{#1}{#2}%
    \citetalias{#1}%
  }%
}

\theoremstyle{plain}

\theoremstyle{definition}

\theoremstyle{remark}

\title{Towards Trustworthy Physical Intelligence: \\From Theory to Practice Across the Life Cycle}

\author{}

\edef\coverdate{\ifcase\month\or January\or February\or March\or April\or May\or June\or July\or August\or September\or October\or November\or December\fi\space\number\day, \number\year}

\def\month{MM}   
\def\year{YYYY}  
\def\openreview{\url{https://openreview.net/forum?id=XXXX}}

\begin{document}

%

\thispagestyle{empty}

\begingroup
\setlength{\parskip}{0pt}
\centering

{\fontsize{22}{25}\selectfont\bfseries
Towards Trustworthy Physical Intelligence:
\par}

\vspace{0.06in}

{\fontsize{20}{22}\selectfont\bfseries
From Theory to Practice Across the Life Cycle
\par}

\vspace{0.20in}

{\small
\noindent
\mbox{Wang~Yang$^{1,*}$},\hspace{0.12em}
\mbox{Hongxuan~Liu$^{2,*}$},\hspace{0.12em}
\mbox{Xinghui~Xu$^{3,*}$},\hspace{0.12em}
\mbox{Arjun~Menon$^{4,*}$},\hspace{0.12em}
\mbox{Xiaoran~Cai$^{5,*}$},\hspace{0.12em}
\mbox{Yunyu~He$^{5,*}$},\hspace{0.12em}
\mbox{Alex~Tarvo$^{6}$},\hspace{0.12em}
\mbox{Jingzong~Zhou$^{7}$},\hspace{0.12em}
\mbox{Mengzhong~Ma$^{8}$},\hspace{0.12em}
\mbox{Xinpeng~Wei$^{9}$},\hspace{0.12em}
\mbox{Yi~Yu$^{3}$},\hspace{0.12em}
\mbox{Shaobo~Wang$^{10}$},\hspace{0.12em}
\mbox{Cheng~Peng$^{11}$},\hspace{0.12em}
\mbox{Aoran~Jiao$^{12}$},\hspace{0.12em}
\mbox{Alexei~Korolev$^{4}$},\hspace{0.12em}
\mbox{Yanyan~Zhang$^{1}$},\hspace{0.12em}
\mbox{Kai~Ye$^{1}$},\hspace{0.12em}
\mbox{Xinpeng~Li$^{1}$},\hspace{0.12em}
\mbox{Chengquan~Guo$^{13}$},\hspace{0.12em}
\mbox{Jingjing~Fu$^{10}$},\hspace{0.12em}
\mbox{Nicholas~Bai$^{14}$},\hspace{0.12em}
\mbox{Yongjun~He$^{15}$},\hspace{0.12em}
\mbox{Silei~Ren$^{16}$},\hspace{0.12em}
\mbox{Mohamad~Louai~Shehab$^{17}$},\hspace{0.12em}
\mbox{Keshu~Cai$^{18}$},\hspace{0.12em}
\mbox{Nathaniel~Dennler$^{2}$},\hspace{0.12em}
\mbox{Traian~Tus$^{11}$},\hspace{0.12em}
\mbox{Gaoyue~Zhou$^{3}$},\hspace{0.12em}
\mbox{Abdullah~Garra$^{19}$},\hspace{0.12em}
\mbox{Mason~Nakamura$^{19}$},\hspace{0.12em}
\mbox{George~Ortiz$^{20}$},\hspace{0.12em}
\mbox{Marius~Urbonas$^{11}$},\hspace{0.12em}
\mbox{Junru~Ren$^{2}$},\hspace{0.12em}
\mbox{Lars~Johannsmeier$^{21}$},\hspace{0.12em}
\mbox{Rohit~Sharma$^{22}$},\hspace{0.12em}
\mbox{Suraj~Deshmukh$^{23}$},\hspace{0.12em}
\mbox{Michael~Spranger$^{24}$},\hspace{0.12em}
\mbox{Vinayak~Gupta$^{25}$},\hspace{0.12em}
\mbox{Xu~Chen$^{6}$},\hspace{0.12em}
\mbox{Ashis~G.~Banerjee$^{6}$},\hspace{0.12em}
\mbox{Yuxiang~Feng$^{26}$}
\\
\mbox{Yoshua~Bengio$^{27}$},\hspace{0.15em}
\mbox{Peng~Qi$^{28,\dagger}$}
\par}

\vspace{0.2in}
{\footnotesize
\setlength{\tabcolsep}{4pt}
\renewcommand{\arraystretch}{0.95}
\begin{tabular}{@{}p{0.31\textwidth}p{0.31\textwidth}p{0.31\textwidth}@{}}

$^{1}$Case Western Reserve University &
$^{2}$Massachusetts Institute of Technology &
$^{3}$New York University \\

$^{4}$Princeton University &
$^{5}$Columbia University &
$^{6}$University of Washington \\

$^{7}$University of California, Riverside &
$^{8}$Nanyang Technological University &
$^{9}$Georgia Institute of Technology \\

$^{10}$Harvard University &
$^{11}$University of Oxford &
$^{12}$University of Toronto \\

$^{13}$University of Chicago &
$^{14}$California Institute of Technology &
$^{15}$ETH Zürich \\

$^{16}$Cornell University &
$^{17}$University of Michigan &
$^{18}$University College London \\

$^{19}$University of Massachusetts Amherst &
$^{20}$Carnegie Mellon University &
$^{21}$NVIDIA \\

$^{22}$Salesforce &
$^{23}$CoreWeave &
$^{24}$Sony AI \\

$^{25}$Biohub &
$^{26}$Imperial College London &
$^{27}$Mila, Université de Montréal \\

$^{28}$Uniphore
\end{tabular}
}

{\footnotesize $^{*}$Equal contribution. \quad $^{\dagger}$Corresponding author.\par}

\vspace{0.05in}


\includegraphics[width=\textwidth,height=5.4in,keepaspectratio]{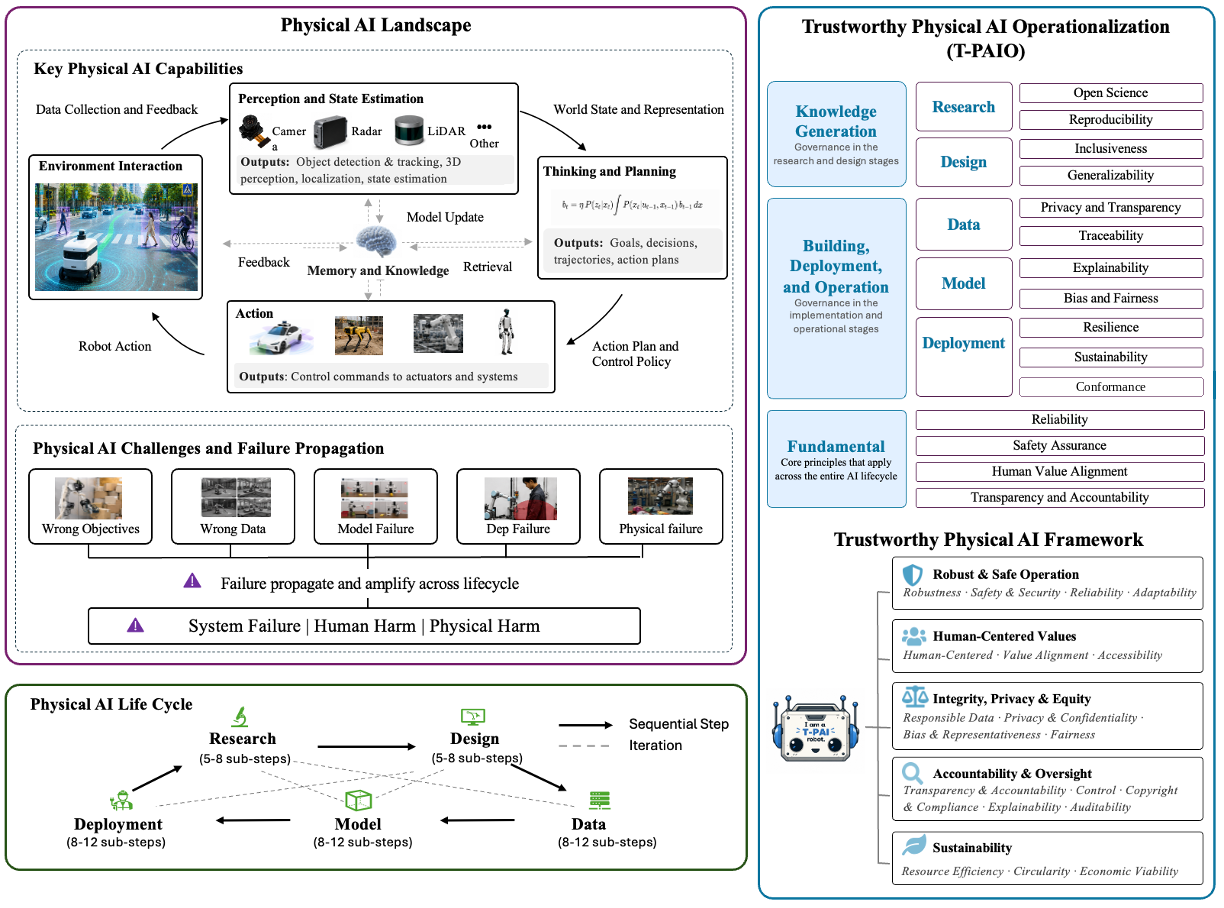}\par

{\small \noindent \raggedright \textbf{Keywords:} Trustworthy Physical Intelligence; Physical Intelligence Governance; Physical Intelligence Life Cycle; Trustworthy Embodied Intelligence. \par} \vspace{0.04in} {\small \noindent \raggedright \textbf{Next Version Update:} Dec 1, 2026 \par}


\endgroup

\clearpage


\thispagestyle{empty}

\begin{center}

\vspace*{0.45in}

{\fontsize{26}{30}\selectfont\bfseries
Contents
}

\vspace{0.12in}

{\large\textit{Towards Trustworthy Physical Intelligence}}

\end{center}

\vspace{0.25in}

\setcounter{tocdepth}{3}

\makeatletter
\@starttoc{toc}
\makeatother

\clearpage
\setcounter{page}{1}

\section*{Abstract}
\addcontentsline{toc}{section}{Abstract}

Physical intelligence refers to intelligence systems that understand, reason about and act in accordance with the physical world and its underlying laws, dynamics, and constraints. Unlike conventional AI systems, physical intelligence interacts continuously with uncertain physical environments, and its actions produce consequences that are physically irreversible. As existing trustworthy AI frameworks have been developed primarily for digital AI systems, they do not fully capture the distinctive challenges of Physical Intelligence, such as physical safety, cyber-physical security, and physical manufacturing process. To address this gap, our paper develop a foundation for understanding, operationalizing, and governing trustworthy physical intelligence. First, we characterize the fundamental capabilities of physical intelligence. Second, we investigate the role of physical laws and principles in AI and analyze the associated challenges of physical intelligence. Third, we provide a unified structure of the end-to-end physical intelligence life cycle across five core stages and introduce the Trustworthy Physical Intelligence Operationalization (T-PAIO) framework. Fourth, we formulate the Trustworthy Physical Intelligence (T-PAI) Principles as a theoretical framework that synthesizes key trustworthiness dimensions and establishes a conceptual foundation for the governance of physical intelligence systems.

\section{Introduction} 

Physical Intelligence refers to physics-aware intelligent systems that perceive, learn, reason, plan, act and adapt to enable or augment actions in the physical world~\citep{verwey2026physical, wu2026physicalai, nature2026physicalai}. The concept that intelligence needs a ``body'' and learns through real-world interaction traces back to Norbert Wiener's cybernetics in the 1940s~\citep{wiener1948cybernetics} and the Shakey robot developed at the Stanford Research Institute in the late 1960s~\citep{nilsson1984shakey}. The modern concept of ``Physical Intelligence'' gained mainstream recognition since 2024 driven by advances from related technologies such as embodied intelligence, world models, and physics-grounded reasoning~\citep{xiang2026aligning}. 

There are three major sub-domains of Physical Intelligence: physics-informed AI, generative physical intelligence, and embodied AI. Physics-informed AI incorporates physical laws, constraints, and scientific knowledge into AI models to improve their ability to reason about physical systems and solve scientific and engineering problems. Generative Physical Intelligence focuses on generating physics-plausible states, trajectories, environments, and scenarios that capture the dynamics and constraints of the real world. Embodied AI emphasizes intelligent agents that perceive and act through physical bodies, enabling robots, autonomous vehicles, and other embodied systems to interact with and adapt to their environments ~\citep{wu2026physicalai}. World models, as an example of generative Physical Intelligence, generate data to enable Physical Intelligence. They learn internal representations of how an environment evolves, enabling AI systems to predict future states, simulate possible outcomes, and anticipate the consequences of actions before executing them. These paradigms connect internal physical understanding with external physical interaction. Physical Intelligence provides a broader framework that integrates these capabilities \citep{xiang2026aligning}.

Physical Intelligence represents an emerging paradigm in the evolution of both artificial intelligence and robotics. First, Physical Intelligence extends artificial intelligence beyond digital environments into the physical world. It enables human-made agents to operate in complex and unstructured real-world environments with an intelligent brain and body \citep{sitti2021physical}. Second, physical intelligence represents a pivotal development in the trajectory of robotics: the first generation, from the 1950s to the 1970s, defined by playback and programmable robots that followed fixed, repeated motions with no sensory feedback; the second generation, from the 1970s to the 2000s, which introduced sensory and adaptive capabilities through vision, force, and tactile sensors, allowing robots to react to their environment within fixed, if-then, rule-based systems; and the third generation, from the 2010s to the present, characterized by intelligent autonomous robots that leverage AI to plan, learn from experience, and generalize beyond their original programming \citep{ma2026breaks}. Physical intelligence marks the transformation of robots from deterministic, pre-programmed machines into intelligent systems capable of perceiving, reasoning and acting in the real world \citep{deloitte2025physicalai}.

However, documented incidents reveal critical challenges and limitations in current physical intelligence systems. For example, in July 2025, a Unitree H1 humanoid robot~\citep{oecd2025unitree_h1_incident} lost control and damaged equipment during testing, with reports attributing the event to a malfunction during operation. A lawsuit filed by former Figure AI safety engineer Robert Gruendel alleged that the company's humanoid robots were capable of generating impact forces more than double those required to fracture an adult human skull \citep{kolodny2025figure}. In a separate case, an autonomous taxi operator paid a multi-million-dollar settlement after one of its self-driving vehicles dragged a pedestrian across a San Francisco street: the perception stack misclassified the situation, and the planner initiated a pull-over maneuver instead of remaining stationary \citep{lin2024cruise}. Two important lessons emerge from these incidents. First, failures in physical intelligence systems can translate directly into physical consequences. Second, such failures are not necessarily confined to a single point of breakdown. Because physical intelligence systems consist of interdependent stages, an error originating upstream can propagate to downstream modules by altering the inputs, internal states, or decisions on which subsequent stages depend. Recent research on embodied AI similarly identifies semantic misgrounding, subtask-level error propagation, execution drift, and contact-related risks as interconnected failure mechanisms that can accumulate over long-horizon robotic tasks. These characteristics motivate an end-to-end life cycle perspective on trustworthy physical intelligence systems, rather than an isolated, modular view that evaluates individual components\citep{Li2026EmbodiedAIAction}.

Physical Intelligence governance is not a wholly distinct discipline from traditional AI governance; rather, it extends established governance principles to address failure modes specific to embodied intelligence operating under physical dynamics and real-time constraints (see Table~\ref{tab:gov_comparison} for illustrative examples). Governance elements fall into three categories: elements retained without modification from traditional digital AI, elements that persist in principle but require reformulation under physical embodiment, and elements with no digital-AI precedent. An autonomous vehicle illustrates all three. Data minimization in telemetry and logging remains governed largely as in traditional digital AI. Transparency, privacy, and accountability persist as requirements but are reformulated: transparency must account for real-time, sensor-conditioned decisions rather than static model outputs; accountability must resolve liability for physical harm to persons and property rather than informational harm alone \citep{li2023trustworthy}. Beyond these, physical intelligence governance imposes requirements absent from digital-AI frameworks entirely: functional safety certification, real-time closed-loop control for mobility, physical human--machine interaction, compliance with traffic regulations, and safe operation in shared road environments \citep{li2023trustworthy, IEC61508-1-2010}. As shown in Table~\ref{tab:survey_comparison}, traditional trustworthy AI framework or machine learning surveys are insufficient. Existing research on the physical intelligence life cycle also remains fragmented, and trustworthy AI frameworks have struggled to keep pace with this fragmentation, remaining siloed and inconsistently applied across organizations. 

\begin{table}[t]
\centering
\caption{Physical Intelligence governance as an extension of traditional AI governance (illustrative examples). \textbf{Retained}: the governance element carries over from traditional digital AI without modification. \textbf{Reformulated}: the governance element originates in traditional digital AI or in established engineering practice but is redefined to account for physical embodiment. \textbf{Physical Extension}: the governance element has no counterpart in traditional digital AI and arises specifically from operation as an embodied system in a dynamic physical environment. The dimensions shown here provide an illustrative comparison with  traditional AI governance. Section~\ref{sec:t-pai} organizes the broader set of Physical Intelligence governance principles into the five T-PAI categories.}
\label{tab:gov_comparison}

\renewcommand{\arraystretch}{1.35}
\resizebox{0.9\textwidth}{!}{%
\begin{tabular}{@{} 
>{\raggedright}p{3.1cm} 
p{4.4cm} 
>{\centering\arraybackslash}p{2.1cm} 
>{\centering\arraybackslash}p{2.1cm} 
>{\centering\arraybackslash}p{2.1cm} 
>{\centering\arraybackslash}p{2.1cm} @{}}
\toprule
\multicolumn{1}{l}{\textbf{Governance Dimension}} & 
\multicolumn{1}{l}{\textbf{Categories}} & 
\multicolumn{1}{c}{\textbf{\makecell{Retained}}} & 
\multicolumn{1}{c}{\textbf{\makecell{Reformulated}}} & 
\multicolumn{1}{c}{\textbf{\makecell{Physical\\Extension}}} & 
\multicolumn{1}{c}{\textbf{\makecell{References}}} \\
\midrule

\multirow{2}{*}{\textbf{Safety}}
 & Model Safety            
 & -- & \checkmark & -- 
 & \cite{jindal2025danger} \\
 
 & Physical Safety         
 & -- & -- & \checkmark 
 & \cite{Kojima2025PhysicalRiskControl} \\
\midrule

\multirow{3}{*}{\textbf{Security}}
 & Dataset Security             
 & \checkmark & -- & -- 
 & \cite{goldblum2022dataset} \\
 
 & Digital Cybersecurity        
 & -- & \checkmark & -- 
 & \cite{giaretta2024cybersecurity} \\
 
 & Cyber-physical Security      
 & -- & -- & \checkmark 
 & \cite{neupane2024security} \\
\midrule

\multirow{2}{*}{\textbf{Adaptability}}
 & Continuous Learning           
 & -- & \checkmark & -- 
 & \cite{liu2024robotonthejob} \\
 
 & Distribution Shift            
 & -- & \checkmark & -- 
 & \cite{Josifovski2025SCDA} \\
\midrule

\multirow{3}{*}{\textbf{Privacy}}
 & Data Privacy               
 & -- & \checkmark & -- 
 & \cite{designCenteredHRIGovernance2022} \\
 
 & Contextual Privacy         
 & -- & \checkmark & -- 
 & \cite{gdprRobotsHRI2026} \\
 
 & Informed Consent           
 & -- & \checkmark & -- 
 & \cite{lintvedt2024consent} \\
\midrule

\multirow{2}{*}{\textbf{Fairness}}
 & Bias \& Discrimination        
 & -- & \checkmark & -- 
 & \cite{londono2024RobotBias} \\
 
 & Accessibility \& Inclusion    
 & -- & \checkmark & -- 
 & \cite{coDesigningAccessibleRobots2024} \\
\midrule

\multirow{3}{*}{\textbf{Transparency}}
 & Explainability                
 & -- & \checkmark & -- 
 & \cite{fernandezbecerra2026b} \\
 
 & Traceability                  
 & -- & \checkmark & -- 
 & \cite{roboLineage2026} \\
 
 & Hardware State Disclosure     
 & -- & -- & \checkmark 
 & \cite{yang2026trustworthy} \\
\midrule

\multirow{3}{*}{\textbf{Accountability}}
 & Responsibility \& Oversight   
 & -- & \checkmark & -- 
 & \cite{faccrt2026roboticAudit} \\
 
 & Software Auditability         
 & \checkmark & -- & -- 
 & \cite{falco2021governingAISafety} \\
 
 & Hardware Auditability         
 & -- & -- & \checkmark 
 & \cite{winfield2017ethical} \\
\midrule

\multirow{3}{*}{\textbf{Control}}
 & Real-time Monitoring          
 & -- & \checkmark & -- 
 & \cite{khalastchi2015} \\
 
 & Reliability \& Latency        
 & -- & \checkmark & -- 
 & \cite{song2025corki,yates2021age} \\
 
 & Fail-safe \& Recovery         
 & -- & \checkmark & -- 
 & \cite{ramaswamy2026managed,Avizienis2004} \\
\midrule

\multirow{3}{*}{\textbf{Sustainability}}
 & Energy Efficiency         
 & -- & \checkmark & -- 
 & \cite{sustainabilityManifesto2026} \\
 
 & Economic Viability        
 & -- & \checkmark & -- 
 & \cite{seong2025costnav} \\
 
 & Material Circularity      
 & -- & -- & \checkmark 
 & \cite{Turner2022Industry5CircularEconomy} \\
\bottomrule
\end{tabular}}
\end{table}

Therefore, our survey addresses the following four questions:

\begin{itemize}
\item Why physics enhance artificial intelligence?

\item What are the key challenges and critical points in physical intelligence, and how can failures emerge, propagate, and cascade across the system?

\item How can trustworthy principles be operationalized into practices across the physical intelligence life cycle?

\item How can governance principles and engineering practices be integrated into a unified framework for trustworthy physical intelligence?
\end{itemize}

\begin{table}[t]
\centering
\caption{Comparison of representative surveys related to physical intelligence. \textbf{Phys.}: focus on physically embodied AI systems; \textbf{E2E LC}: end-to-end life cycle coverage from research and design through deployment, operation, and continuous improvement; \textbf{Gov.}: governance coverage; \textbf{Gov. Ops.}: governance operationalization throughout the life cycle. Symbols: \checkmark~= comprehensive coverage; $\circ$~= partial coverage; $\times$~= not covered.}
\label{tab:survey_comparison}
\renewcommand{\arraystretch}{1.2}
\small
\setlength{\tabcolsep}{5pt}
\begin{tabularx}{\textwidth}{>{\raggedright\arraybackslash}p{3.6cm}Xcccc}
\toprule
\textbf{Survey Reference} &
\textbf{Primary Focus} &
\textbf{Phys.} &
\textbf{E2E LC} &
\textbf{Gov.} &
\textbf{Gov. Ops.} \\
\midrule
\citet{yang2026trustworthy} & Towards Trustworthy Embodied Intelligence: A Systems Framework and Graded Trustworthiness Levels & $\checkmark$ & $\times$ & \checkmark & $\circ$ \\
\citet{tan2025safetrustworthyeai} & Towards Safe and Trustworthy Embodied AI: Foundations,Status,and Prospects & $\checkmark$ & $\times$ & \checkmark & $\circ$ \\
\citet{xing2025robust} & Towards Robust and Secure Embodied AI: A Survey on Vulnerabilities and Attacks & $\checkmark$ & $\times$ & \checkmark & $\circ$ \\
\citet{li2023trustworthy} & Trustworthy AI principles and practices for conventional digital AI & $\times$ & $\checkmark$ & \checkmark & $\circ$ \\
\citet{neupane2024security} & Security of AI--robotics systems & \checkmark & $\times$ & \checkmark & $\circ$ \\
\citet{duan2024} & Embodied AI aligning cyber and physical space & \checkmark & $\times$ & $\times$ & $\times$ \\
\citet{firoozi2025foundation} & Foundation models in robotics & \checkmark & $\times$ & $\circ$ & $\times$ \\
\citet{long2025embodiedworldmodels} & Embodied intelligence via physical simulators and world models & \checkmark & $\times$ & $\times$ & $\times$ \\
\citet{mccarthy2025internetvideo} & Generalist robot learning from internet video & \checkmark & $\times$ & $\times$ & $\times$ \\
\citet{xiang2026aligning} & Aligning Perception, Reasoning, Modeling and Interaction: A Survey on Physical AI & \checkmark & $\times$ & $\circ$ & $\times$ \\
\citet{chen2025emlms} & Embodied multimodal large models: development and datasets & \checkmark & $\times$ & $\times$ & $\times$ \\
\citet{ma2026vla} & Vision--language--action models for embodied AI & \checkmark & $\times$ & $\circ$ & $\times$ \\
\citet{gaba2026physical} & A Comprehensive Review of Generative Physical Artificial Intelligence & \checkmark & $\times$ & $\circ$ & $\times$ \\
\citet{wu2026physicalai} & Evolution, progress, and challenges of physical intelligence & \checkmark & $\times$ & $\circ$ & $\times$ \\
\citet{li2026safety} & Safety risks, attacks, and defenses in embodied AI & \checkmark & $\times$ & \checkmark & $\circ$ \\
\citet{mccormack2026taimm} & Trustworthy-AI maturity model across the AI life cycle & $\times$ & $\checkmark$ & \checkmark & \checkmark \\
\citet{Torras2024SocialRobotics} & Ethics of social robotics and human--robot interaction & \checkmark & $\circ$ & \checkmark & $\times$ \\
\citet{Ambhore2024RoboEthics} & Comprehensive survey on robo-ethics & \checkmark & $\times$ & \checkmark & $\times$ \\
\citet{Torresen2018EthicalRobotics} & Future and ethical perspectives of robotics and AI & $\circ$ & $\times$ & \checkmark & $\times$ \\
\citet{Kurkchi2025EthicsCPS} & Ethics-based requirements for engineering cyber--physical systems & $\circ$ & $\circ$ & \checkmark & $\circ$ \\
\citet{Flick2022EthicsCreativeAI} & Ethics of creative and generative AI & $\circ$ & $\times$ & \checkmark & $\times$ \\
\midrule
\textbf{This Paper} & End-to-end Trustworthy Physical Intelligence Operationalization (T-PAIO) framework with life cycle-aware governance (T-PAI) & \checkmark & \checkmark & \checkmark & \checkmark \\
\bottomrule
\end{tabularx}
\end{table}

To address these four questions, we proceed in four steps. 
Firstly, we identify key Physical Intelligence capabilities and the critical points across the physical intelligence life cycle at which failures can emerge, propagate, and cascade, potentially resulting in system, human, or physical harm (\Cref{sec:failure-propagation}). 
Secondly, we examine how physical laws, dynamics, and constraints can support learning, ground generative models, and inform safety enforcement (\Cref{sec:physics-foundations}).
Thirdly, we introduce Trustworthy Physical Intelligence Operationalization (T-PAIO) framework as an end-to-end physical intelligence life-cycle framework that establishes the principles and practical guidance for data, models, evaluation, and operational controls needed to operationalize trustworthy physical intelligence (\Cref{sec:epalgo}). 
Finally, building on this life-cycle foundation, we develop Trustworthy Physical Intelligence (T-PAI) as a theoretical framework that systematically organizes the key trustworthiness dimensions and provides a foundation for developing trustworthy Physical Intelligence strategies and practices (\Cref{sec:t-pai}).

\begin{itemize}
    \item 
    \textbf{Capabilities and Physics Foundation for AI}: 
    We identify the key capabilities of physical intelligence. We also examine how physical principles can enhance AI by incorporating knowledge of physical laws, constraints, dynamics, and causal relationships into AI models and decision-making.
    
     \item 
    \textbf{Failure Propagation}: 
    We characterize how failures originating at upstream stages can propagate through interdependent components, and ultimately result in system, human, or physical harm.

    \item 
    \textbf{T-PAIO: Trustworthy Physical Intelligence Operationalization}: 
    We first identify and characterize the physical intelligence life cycle. For each stage of the life cycle, we identify the key trustworthiness requirements, governance principles, and engineering practices needed to incorporate trustworthy AI components into the development and operation process.
   
    \item 
    \textbf{T-PAI: Trustworthy Physical Intelligence Framework}: 
   We develop T-PAI (Trustworthy Physical Intelligence) as a theoretical framework for strategic 
Physical Intelligence 
operationalization, translating high-level trustworthiness principles into a systematic structure to guide engineering practices, safeguards, evaluation mechanisms, and governance decisions.
\end{itemize}

\section{Capabilities and Physics Foundations}
\label{sec:failure-propagation}

The ability to interact with the physical world requires more than the capacity to process data, recognize patterns, or generate predictions. Unlike conventional AI applications operating primarily in digital environments, physical intelligence systems must operate within environments governed by physical laws, where actions have material consequences and errors can propagate through sensing, reasoning, and actuation. Purely data-driven approaches do not necessarily guarantee physical consistency, causal understanding, or reliable behavior under changing environmental conditions \citep{karniadakis2021physics,jiao2024aimeetsphysics,ray2026physical}. This creates a fundamental distinction between intelligence that learns statistical regularities from observations and intelligence. Accordingly, understanding the role of physics in AI is essential to advancing Physical Intelligence, particularly in improving model generalization, grounding predictions in physical dynamics, and establishing enforceable constraints for safe interaction with the physical world.

\subsection{Capabilities}

Physical Intelligence systems progress through a life cycle spanning research, design, data collection, model development, system integration and evaluation, deployment, monitoring and adaptation, and eventual decommissioning. Across these stages, their core capabilities include perception and state estimation, cognition and world modeling, planning and decision-making, action and learning, and memory and safety (Table~\ref{tab:physical_ai_primitives}). These capabilities vary in their degree of autonomy, task complexity, environmental adaptability, and social or cognitive interaction ability~\citep{long2025embodiedworldmodels}. They also depend on different combinations of hardware and software. Perception combines physical sensors with software for interpreting observations and estimating state; cognition and planning depend heavily on software; and action combines software-generated commands with physical actuators~\citep{neupane2024security}.

Building on this capability structure, the life cycle of Physical Intelligence systems introduces distinct technical, operational, and governance considerations at each stage. During research and design, system objectives, operational boundaries, embodiment requirements, and safety constraints must be established to align intended capabilities with the physical environments in which systems will operate \citep{driess2023,neupane2024security}. Data collection and model development subsequently determine the quality, diversity, and representativeness of training data, as well as the system's ability to learn physical dynamics, generalize across tasks, and respond to previously unseen conditions \citep{bousmalis2023,brohan2022,long2025embodiedworldmodels}. System integration and evaluation require the coordinated validation of hardware, software, control policies, and human--robot interactions under both simulated and real-world conditions \citep{kim2024,bjorck2025groot,neupane2024security}. Following deployment, continuous monitoring and adaptation become essential for detecting performance degradation, distribution shifts, unexpected behaviors, and emerging safety risks, while enabling model updates, policy refinement, and operational recovery \citep{yates2021age,van2026tap,lei2025,li2026safety}. Finally, decommissioning requires the controlled retirement of physical components, models, data, and associated infrastructure, including the management of residual risks and system dependencies \citep{neupane2024security,li2026safety}. Importantly, these stages are not entirely linear or isolated; feedback from deployment, monitoring, and evaluation can inform subsequent data collection, model retraining, system redesign, and safety validation \citep{bousmalis2023,long2025embodiedworldmodels,lei2025}. Consequently, trustworthy Physical Intelligence requires lifecycle-wide coordination of technical capabilities, hardware--software dependencies, human oversight, and risk management to ensure that system performance, reliability, and safety are maintained throughout its operational lifetime.

\begin{table}[t]
\centering
\caption{Core Capabilities of Physical Intelligence systems.}
\label{tab:physical_ai_primitives}
\renewcommand{\arraystretch}{1.2}
\small
\begin{tabularx}{\textwidth}{
  >{\raggedright\arraybackslash}p{2cm}
  >{\raggedright\arraybackslash}X
  >{\raggedright\arraybackslash}p{2.5cm}
  >{\raggedright\arraybackslash}p{3cm}
  >{\raggedright\arraybackslash}p{2.5cm}
}
\toprule
\textbf{Capability} &
\textbf{Definition} &
\textbf{Inputs} &
\textbf{Outputs} &
\textbf{References} \\
\midrule

\textbf{Perception and State Estimation}
&
Acquire and interpret multimodal sensory signals (vision, depth, touch, force, audio, proprioception, inertial), then align and fuse observations while tracking sensor health, uncertainty, and observation age to estimate robot pose, object state, maps, and spatial relations.
&
Raw sensor streams, acquisition timestamps, sensor health, prior state
&
Objects, poses, geometry, affordances, scene state, belief state, localization, map, tracked object state, uncertainty, observation age
&
\citep{driess2023,duan2024,xu2024,neupane2024security,long2025embodiedworldmodels,yates2021age,van2026tap}
\\
\midrule

\textbf{Cognition and World Modeling}
&
Ground language and goals in perception to infer semantics, relations, and task-relevant context, while modeling environment dynamics to forecast future states under candidate actions, including counterfactuals.
&
Estimated state, images, language, memory, current/history states, candidate actions
&
Semantic scene understanding, grounded reasoning, future states, rollouts under uncertainty, imagined trajectories
&
\citep{bruce2024,driess2023,huang2023,long2025embodiedworldmodels}
\\
\midrule

\textbf{Planning and Decision-Making}
&
Select goals, priorities, and next actions under constraints, rewards, risk, and incomplete information, then decompose chosen goals into feasible skills, subgoals, or action sequences respecting embodiment and environmental constraints.
&
Grounded state, user intent, rewards, constraints, goal, world state, skill library
&
Chosen goal, subgoal or skill, task plan, waypoints, motion/action sequence
&
\citep{ahn2022,cohen2024,liang2025,neupane2024security}
\\
\midrule

\textbf{Action and Learning}
&
Generate and execute closed-loop motor commands for manipulators, mobile bases, humanoids, or other embodied systems, while acquiring or updating policies from demonstrations, interaction, simulation, and feedback to adapt across tasks and embodiments.
&
Planned actions, observations, robot state, demonstrations, rollouts, rewards, synthetic data
&
Joint, end-effector, gripper, or locomotion commands, updated policy/model, new skills, transferred capabilities
&
\citep{bousmalis2023,brohan2022,kim2024,bjorck2025groot}
\\
\midrule

\textbf{Memory and Safety}
&
Store, consolidate, and retrieve spatial, temporal, episodic, semantic, and procedural experience for long-horizon behavior, while detecting hazards or failures (including stale observations, inconsistent sensing, and degraded coverage), enforcing constraints, recovering safely, or requesting human intervention when needed.
&
Past observations, actions, outcomes, knowledge, perception, plans, system health, observation age, risk signals
&
Retrieved experience, persistent world knowledge, retained skills, safe action, fallback, recovery, alert, or shutdown
&
\citep{kwon2025,lei2025,li2026safety,neupane2024security,jindal2025danger}
\\
\bottomrule
\end{tabularx}
\end{table}

\subsection{Intelligence without Physics}

\textbf{Physics Applications in AI.}
The increasing deployment of AI in physical environments has exposed limitations of purely data-driven learning, particularly when models must generalize beyond observed data, operate under physical constraints, or make reliable predictions in regimes where training data are limited. Physics-informed and physics-based AI has therefore emerged as an approach to integrate established knowledge of physical laws, symmetries, constraints, dynamics, and mechanistic relationships into the learning process \citep{karniadakis2021physics,jiao2024aimeetsphysics}. By providing structured inductive biases, physical knowledge can constrain the space of admissible solutions and reduce the extent to which AI systems must infer fundamental relationships solely from empirical correlations. This integration can improve data efficiency, generalization, robustness, physical consistency, interpretability, and computational efficiency, while also enabling AI to address forward and inverse problems, simulation, optimization, and control \citep{karniadakis2021physics,thuerey2021physicsbased}. In particular, physical symmetries and invariances provide transferable structural information that can improve model representations across different configurations and operating conditions \citep{jiao2024aimeetsphysics}.
Importantly, physics-based AI does not require complete knowledge of the underlying physical system. When governing equations are known but constitutive relationships or other system components remain difficult to model, hybrid approaches can preserve the known physical structure while using neural networks to learn the unknown components. For example, Neural Constitutive Laws explicitly enforce governing partial differential equations while learning material behavior from motion observations, enabling generalization across geometry, initial and boundary conditions, temporal ranges, and multiphysics settings \citep{ma2023learning}. Conversely, when physical simulation is computationally expensive, AI can learn surrogate representations that accelerate simulation and enable real-time prediction, optimization, and control \citep{thuerey2021physicsbased}. This complementary relationship positions physics not merely as a constraint on AI, but as a source of inductive bias, mechanistic knowledge, and computational structure. For Physical Intelligence, such integration is particularly important because intelligent systems must ultimately predict, reason about, and act upon physical processes, where statistical plausibility alone does not guarantee physically valid behavior \citep{ray2026physical}. Thus, a central research direction is to determine how and under what conditions physical knowledge can be systematically incorporated into AI to improve its generalization, efficiency, robustness, and trustworthiness in real-world physical environments.

\subsection{Intelligence with Physics}
\textbf{Physics as a learning catalyst} Physical structure can be injected into a model as an observational bias shaping the training data, an inductive bias shaping the architecture, or a learning bias shaping the loss function~\citep{karniadakis2021physics,hao2022physicsinformed}. The signature of this role is a data-efficiency and generalization gap rather than a binary success/failure split. Consider a hypothetical illustration: a model predicting a manipulator's future joint trajectory from a short history of states. Trained purely on observed trajectories with no physical prior, the model must independently rediscover that joint velocities are continuous and that torque is bounded, and it will typically require an order of magnitude more training trajectories to avoid predicting an abrupt, physically impossible velocity jump on an unseen motion. The same model trained with a physics-informed loss term penalizing violations of the manipulator's known dynamics equations reaches comparable accuracy with substantially less data and, more importantly, continues to respect velocity continuity on motions outside its training distribution, because the constraint was never learned empirically in the first place~\citep{karniadakis2021physics}. 

\textbf{Physics as a generative grounding mechanism.} World models and video-based simulators trained to maximize visual likelihood rather than dynamical consistency exhibit a well-documented before/after signature: without an explicit physical grounding, a video-based world model can produce a future frame in which an object grasped by a robot arm deforms in a non-causal way, changes color without cause, or appears in the gripper without a preceding contact event, all while the model reports the prediction with high visual confidence~\citep{mei2025worldmodels}. Physical hallucination like this occurs when embodied AI systems generate actions or plans that are inconsistent with physical constraints, environmental conditions, or feasible real-world interactions \citep{chakraborty2025heal}. This failure mode is consequential precisely because the output is fluent rather than obviously wrong, so a downstream planner has no signal to distrust it.  World models that instead condition generation on an explicit dynamics representation, rather than pixel statistics alone, are reported to suppress this class of error because object motion is constrained by a simulated or inferred physical state rather than sampled from a purely learned prior~\citep{hu2026videofoundations}. The same before/after pattern recurs at smaller scale throughout Generative Physical Intelligence: a simulator or synthetic-data generator without a grounded contact or dynamics model will produce trajectories that are visually plausible but strategically useless for training a downstream policy, since a policy trained on them inherits the same physically inconsistent assumptions.

\textbf{Physics as an enforceable safety constraint.} The first two roles both operate statistically: they shift the average behavior of a model but do not guarantee that any single output obeys a given physical law. Embodied AI closes this gap with an enforcement mechanism external to the learned model itself. Consider a hypothetical collaborative manipulation policy trained with a reward that penalizes proximity to a human collaborator. Without an independent enforcement layer, the policy can still, under distribution shift or an unusual approach angle, propose a motion that closes to an unsafe distance, since the penalty only discourages such actions on average rather than forbidding them outright. Wrapping the same policy with a control barrier function that certifies a safe set $\mathcal{C} = \{x : h(x) \geq 0\}$ and projects any unsafe proposal onto its boundary before actuation changes this outcome categorically rather than statistically: the resulting command is guaranteed to respect the encoded protective distance regardless of what the underlying policy proposed, and regardless of whether the situation that triggered the correction ever appeared during training~\citep{ames2017control}. Differentiable variants of this filter extend the same guarantee to end-to-end learned controllers rather than only hand-designed ones, so the enforcement layer can be trained jointly with the policy without weakening the guarantee it provides~\citep{xiao2023barriernet}.

\section{Failure Propagation}
\label{sec:physics-foundations}

We distinguish two forms of failure propagation. During operation, an error in one capability can alter the inputs or decisions of another: perception errors can produce incorrect state estimates, mislead planning, and generate unsafe actuator commands, so that a single error can cascade across the autonomy stack~\citep{schwarting2018planning,koopman2017autonomous}. The resulting action changes the physical environment and the observations available in subsequent control cycles. Across the life cycle, an incorrect assumption or unresolved defect can persist through development, evaluation, and deployment, becoming consequential only under particular operating conditions~\citep{leveson2011,sculley2015hidden}. These two forms interact when defects introduced during development become the starting points for failures during operation.

\subsection{Failure Driving Factors}
Physical Intelligence failures can arise from fragmented interfaces and shared dependencies, latency, uncertainty and changing conditions, hardware faults and degradation, human and organizational factors, and deliberate interference~\citep{fonseca2026interoperability,zhang2023impacts,gawlikowski2023survey,guiochet2017safety}. Some of these conditions, such as acting on stale observations or relying on a shared faulty dependency, have established counterparts in concurrent, distributed, and real-time systems~\citep{yates2021age,Avizienis2004}. Physical embodiment makes their consequences depend on environmental change, on how components interact, and on whether downstream safeguards detect or contain the initial error\cite{ma2026breaks}.

\emph{Fragmentation and shared dependencies.} Fragmentation refers to the lack of compatible interfaces, data formats, communication protocols, and update practices across components and providers. It becomes consequential when heterogeneous systems must cooperate, such as drones from different vendors or robots connected to hospital infrastructure. Mismatched life cycles add further difficulty: models and software interfaces may change frequently, while physical equipment remains deployed for years. Uneven updates can therefore create compatibility drift and unsafe interactions~\citep{morabito2026internet,quigley2009ros,kramer2007development,koopman2017autonomous}. Shared dependencies introduce a related failure pathway: where fragmentation prevents components from working together, a shared dependency couples them so that one defect reaches all of them at once. For example, several robots receiving the same defective update could develop the same fault, while a planner and safety monitor using the same incorrect state estimate could both accept an unsafe action. Redundant sensing is subject to the same limitation, since duplicated channels remain vulnerable to shared power, timing, calibration, or processing faults and are not independent merely because their hardware differs~\citep{Avizienis2004}. Evaluating these components separately would not establish that their failures are independent.

\emph{Latency and information freshness.} Latency is the delay between observing a physical state and responding to it. Delays in sensing, inference, communication, planning, or actuation can make an observation outdated before the corresponding action is executed. In multi-robot systems, communication delays can also destabilize formation control or produce inconsistent decisions during coordinated tasks~\citep{jiang2012stable,bharadwaj2022concurrent,martorelltorres2024coordination}. Information freshness, or observation age, describes how old the underlying observations are when they are used. It depends on sampling frequency and queueing as well as processing and transmission delays~\citep{yates2021age}. A recently produced estimate may still depend on old sensor data, so freshness must be assessed from acquisition timestamps on a common clock. Acting on an aged observation creates a physical analog of the time-of-check/time-of-use (TOCTOU) race condition, in which a condition changes after it has been checked but before the operation that relies on it~\citep{mitreCWE367}. For example, a robot may verify that a corridor is clear and then execute the accepted trajectory after a person has entered the corridor. The original observation need not have been wrong. Checking again before execution can reduce the window, but the environment continues to change during motion. Safe operation therefore depends on the age and uncertainty of the observations at actuation, the motion possible before the next response, and the time and distance required to stop~\citep{marvel2017speed}. Section~\ref{sec:lifecycle_components} discusses the corresponding action-age, expiration, and fallback requirements.

\emph{Uncertainty and changing conditions.} Physical environments are partially observable and difficult to model completely. Noisy sensors, occlusion, unpredictable human behavior, and unknown dynamics can all introduce uncertainty. Changes in friction, payload, contact conditions, or component wear can further invalidate assumptions made during development~\citep{on2026physics,aljalbout2026reality}. A household robot may encounter an unfamiliar object or human behavior, while an autonomous vehicle may face road or weather conditions absent from its evaluation data~\citep{Kojima2025PhysicalRiskControl,aljalbout2026reality}. Errors can propagate when downstream components act on an estimate without accounting for its uncertainty or the conditions under which it remains valid.

\emph{Hardware faults and degradation.} Physical components can initiate failures as well as experience their consequences~\citep{guiochet2017safety}. For example, sensor calibration drift can corrupt state estimation, and reduced actuator capability can prevent a controller from executing a planned motion. A plan may satisfy the model's assumed limits while exceeding the hardware's actual capability. Hardware-originating errors therefore need to be considered throughout sensing, planning, and control, including during continued operation as component condition changes.

\emph{Human and organizational factors.} Human supervision and organizational processes can also influence whether an error is contained~\citep{leveson2011}. An operator may misunderstand the active control mode, receive an alert too late to intervene, or lack clear authority to stop operation. Maintenance or configuration changes may alter sensing coverage or stopping performance without triggering a corresponding review. In these cases, a technical fault can become harmful because the expected intervention or change-management process does not function as assumed.

\emph{Deliberate interference.} Security threats can arise throughout the life cycle. Attackers may exploit sensors, communication channels, software dependencies, or control interfaces to manipulate observations, decisions, or actions. Because Physical Intelligence systems control physical processes, compromised software or communication can lead to unsafe movement and equipment damage as well as information loss~\citep{dieber2017security}. Security is therefore a source of failure across stages, including development, deployment, and continued operation.

\emph{An illustrative propagation sequence.} Consider a hypothetical mobile robot whose camera mount shifts after an impact. The resulting calibration error places an obstacle farther from the planned path than it actually is. The planner selects a route with insufficient clearance, and the controller begins executing it. As the robot turns, the obstacle leaves the camera's field of view, so subsequent planning depends on the earlier incorrect estimate. The action has therefore changed the observations available to correct the initial error. If the collision monitor uses the same estimate, it may also accept the trajectory. A second camera viewing the region from another position~\citep{arnold2022cooperative} or a separate proximity sensor that does not share the mount fault could interrupt the sequence by triggering reduced speed or a controlled stop. Whether that intervention prevents a collision depends on its coverage, the age of its observations, its detection and response time, and the robot's actual stopping capability. The example distinguishes an initiating fault, its propagation through the system, and the conditions under which a safeguard can prevent harm.

\subsection{Failure Across the Life Cycle}

The propagation of failures is not restricted to the stage in which a defect originates; rather, it can span multiple stages as assumptions, data, models, and dependencies of components are carried forward \cite{xiang2026aligning}. Failures can be introduced, overlooked, or amplified at different stages of the life cycle. The following discussion organizes these failures by the activities that introduce or expose them. Security threats and human or organizational factors can affect several stages, while physical harm is a possible consequence of failures across them.

During research and design, assumptions about the task, environment, hardware, and human supervision shape the system's objectives and operating constraints. Objective misspecification occurs when the formal objective or reward function does not adequately capture the intended outcome. An agent may then optimize the specified metric while causing unintended physical effects. For example, a robot rewarded for moving a box may complete the task while knocking over a vase if collateral effects are absent from the objective~\citep{amodei2016concrete}. Such omissions can persist into data collection, model training, and evaluation when all three use the same incomplete definition of success.

During data collection and preparation, failures can arise when the data do not represent the conditions in which the system will operate. Differences between training and deployment distributions can make learned relationships unreliable. A perception or control model trained under particular lighting, sensing, or environmental conditions may fail when those conditions change~\citep{subbaswamy2019preventing}. If evaluation uses similarly restricted data, the limitation may remain hidden until deployment.

During model development, learned policies may produce incorrect, inconsistent, or unstable actions. Mobile-manipulation policies, for example, can generate temporally inconsistent actions or repeatedly attempt actions that do not complete tasks such as closing a box or covering an object outside their training conditions~\citep{agia2025unpacking}. These failures become consequential through the control interface that converts policy outputs into physical commands, making the behavior of the complete system relevant to evaluation.

During system integration, individually functional components may interact unsafely because of incompatible interfaces, timing mismatches, or inconsistent assumptions about state, units, or operating limits. Such hazards cannot be identified solely by evaluating each component in isolation~\citep{leveson2011}. Evaluation and validation can then fail to expose these interactions when testing omits rare events, environmental variation, or combinations of degraded conditions such as delayed observations, sensor outages, and occlusion. Strong aggregate performance may conceal weaknesses in particular operating conditions and allow them to persist into deployment~\citep{koopman2017autonomous}.

During deployment, differences between simulated or controlled environments and the target setting can invalidate assumptions about sensing, dynamics, and actuation. A policy may depend on idealized sensors, simplified lighting, or actuator behavior that the physical robot cannot reproduce. This reality gap can prevent successful simulated behavior from transferring to real operation~\citep{aljalbout2026reality}. Monitoring and adaptation are important part of a successful deployment. Failures can arise when changes in behavior, data distributions, environmental conditions, or component dependencies go undetected. Updates intended to improve one component may introduce incompatibilities or degrade other parts of the system~\citep{sculley2015hidden,NISTAIRMF2023}. Monitoring, testing, and rollback can help contain these problems, but their effectiveness depends on detecting the affected behavior and retaining a configuration suitable for the current hardware and operating conditions. Operational findings must also reach the teams responsible for design, data, and models so that the initiating defect is addressed. During decommissioning or transfer, failures can arise when physical equipment, software dependencies, data, models, credentials, and downstream users are not managed together. Unsupported hardware may remain in use, retired models may remain accessible through old integrations, and replacement systems may inherit assumptions that no longer hold. Poorly coordinated handoffs can also remove safety controls or create gaps in monitoring and responsibility~\citep{NISTAIRMF2023}.

Across these stages, an error does not necessarily result in physical harm: detection, constraint enforcement, human intervention, or recovery may interrupt its propagation. Harm becomes possible when unsafe behavior reaches the physical system and the available safeguards do not prevent exposure to the hazard. Robot-related accidents illustrate the consequences of machine failures, inadequate risk assessment, insufficient safeguards, and failures in human--robot interaction~\citep{adesiji2026safety}. The life cycle perspective therefore requires identifying both where failures originate and where their progression can be interrupted. Section~\ref{sec:epalgo} develops the corresponding engineering and governance practices for the research, design, data, model, and deployment stages.

Physics contributes to Physical Intelligence systems through three functionally distinct roles. As a \emph{learning catalyst}, physical structure improves data efficiency and generalization in Physics-Informed AI. As a \emph{generative grounding mechanism}, it constrains predicted futures to remain dynamically consistent in Generative Physical Intelligence. As an \emph{enforceable safety constraint}, it bounds executable actions independently of the learned policy in Embodied AI. The first two roles are statistical, altering a model's average behavior; the third is categorical, certifying or rejecting individual outputs irrespective of training. We characterize each role below through its before/after signature, i.e., the failure mode observed when the corresponding physical grounding is absent.

\section{\texorpdfstring{T-PAIO: \underline{T}rustworthy \underline{P}hysical \underline{A}I \underline{O}perationalization}{T-PAIO: Trustworthy Physical Intelligence Operationalization}}
\label{sec:epalgo}

In this section, we distill the life cycle of Physical Intelligence and survey existing research approaches to operationalizing governance throughout each stage of the life cycle. 

Physical Intelligence requires life-cycle-wide trustworthiness assurance because its risks are dynamic and assurance evidence is not automatically transferable across development and deployment stages. Unlike conventional AI models, whose risks are often assessed before release, physical intelligence systems operate in changing environments, experience hardware degradation, and may undergo policy updates, causing their risk profiles to evolve throughout their service lives. Consequently, risk must be continuously assessed as a function of the operating environment, hardware condition, and deployed policy, rather than treated as a fixed property established at release \citep{continuousAssurance2025,assuranceLEAS2023,Avizienis2004}. Moreover, evidence established through simulation, laboratory testing, or pilot deployment cannot be assumed to remain valid in subsequent operational settings, as each transition introduces different environmental conditions and data distributions \citep{zhao2020sim,s2rbench2025,Josifovski2025SCDA}. A one-time assessment or model-level certification is therefore insufficient to establish sustained trustworthiness. Instead, assurance must extend across the entire life cycle through continuous risk monitoring, stage-specific validation, and explicit reassessment at each transition to ensure that safety and performance claims remain valid under evolving operational conditions.

Given the early stage of Physical Intelligence and the broad range of capabilities, we distill the life cycle into two overarching layers: the \emph{knowledge layer} and the \emph{build and operation layer}. The knowledge layer encompasses research and design that generate, validate, and disseminate foundational knowledge, algorithms, and best practices for Physical Intelligence. The build and operation layer comprises data collection, model development and deployment, where these capabilities are implemented and refined in real-world environments. Some trustworthiness considerations span the entire life cycle, while others are specific to particular stages. Moreover, the life cycle is inherently iterative, with continuous feedback among research, development, deployment, and operational experience driving ongoing improvements in both system capabilities and trustworthiness practices.

\begin{table}[t]
\centering
\caption{Structural differences between the traditional AI and Physical Intelligence life cycles, and the trustworthiness obligation each creates. These differences recur, in stage-specific form, in every component of Section~\ref{sec:lifecycle_components}.}
\label{tab:lifecycle_delta}
\small
\renewcommand{\arraystretch}{1.2}
\begin{tabularx}{\textwidth}{>{\raggedright\arraybackslash}p{2.9cm}XX}
\toprule
\textbf{Structural difference} & \textbf{How Physical AI intensifies or reformulates the issue} & \textbf{Governance obligation it creates} \\
\midrule
\textbf{Embodiment coupling} &
A digital model is hardware-independent and copies exactly; a robot result is conditioned on a particular platform, calibration, and sensor suite. &
Artifact and hardware disclosure become safety-relevant, not merely scientific hygiene~\citep{mirrer2024,Bonsignorio2025ReproducibleRobotics}. \\
\addlinespace
\textbf{Distribution gap between stages} &
Digital evaluation sets are drawn from the deployment distribution; simulators, however, model the physics of the real-world with 
\emph{uncertainties} due to simplifying assumptions, computational approximations, and unknown parameters or constraints. &
Measure and report at each handoff instead of reusing upstream evidence~\citep{s2rbench2025,re3sim2025}. \\
\addlinespace
\textbf{Irreversibility under a real-time bound} &
A harmful token can be filtered before it is displayed; a harmful force or torque, on the other hand, can be applied before it is reviewed. &
Safety must be enforced \emph{online}, within the control period, by a filter or fallback controller rather than post-hoc moderation~\citep{Haddadin2014SafeRobots,ISO15066,Chen2021SimplexDrive}. \\
\addlinespace
\textbf{Physical degradation} &
Software does not wear out; sensors, actuators, and mountings do, and their drift is silent. &
Certification validity decays with time; scheduled re-validation and degradation monitoring are required~\citep{okeeffe2023,continuousAssurance2025}. \\
\addlinespace
\textbf{Split accountability chain} &
A single provider typically controls model, serving stack, and interface. &
Responsibility across hardware OEM, model developer, integrator, and site operator must be allocated contractually at design time, and the allocation
must specify the evidence and access each party requires to discharge its
own obligations, not only the liabilities it accepts
~\citep{wef2026infrastructure,pwc2026physical,faccrt2026roboticAudit}. \\
\bottomrule
\end{tabularx}
\end{table}

\textbf{How to read this section?} For each life cycle component, we address three questions in turn: (i) What is involved in the component and how does it differ from its traditional-AI analog? (ii) What concrete research artifacts, benchmarks, and standards should be consulted by a practitioner working at that stage? and (iii) What quantities can actually be measured, budgeted, or bounded? We address these questions in Section~\ref{sec:lifecycle_components}, while 
Section~\ref{sec:operationalizing} maps the governance elements of T-PAI onto each of them.

\subsection{Key Physical Intelligence Life Cycle Components}
\label{sec:lifecycle_components}
\begin{figure}[!t]
    \centering
    \includegraphics[width=\textwidth]{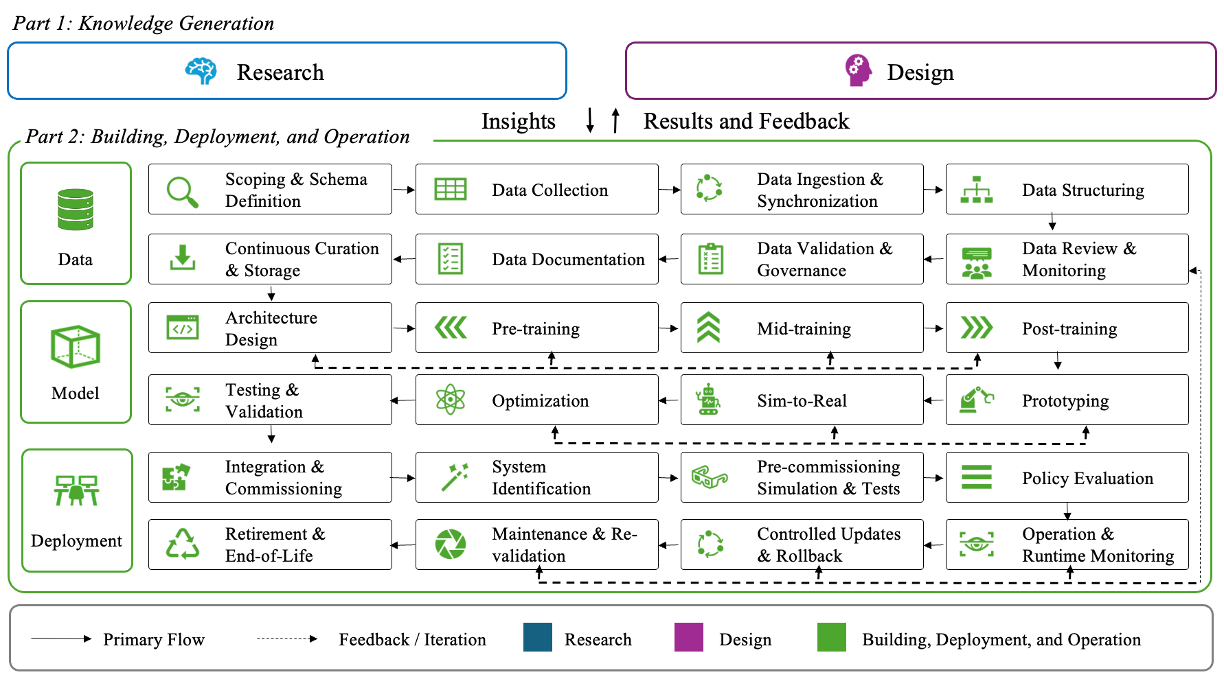}
    \caption{
    Overview of a potential Physical Intelligence life cycle as an example. The framework is organized into two complementary parts. \textbf{Part 1: Knowledge Generation} comprises Research and Design, where scientific knowledge, engineering principles, and system requirements are established through iterative feedback. \textbf{Part 2: Building, Deployment, and Operation} operationalizes these insights through the Data, Model, and Deployment stages, covering data engineering, model development, system integration, and continuous operation. Feedback from deployment informs subsequent research and design, forming a closed-loop life cycle that supports the continuous evolution of Physical Intelligence systems.
    }
    \label{fig:lifecycle}
\end{figure}

As shown in Figure~\ref{fig:lifecycle}, the Physical Intelligence life cycle covers the same core stages as traditional AI but places significantly greater emphasis on iterative development, continuous feedback, and cross-stage interactions among research, design, data, models and deployment.\footnote{Unless a study is explicitly cited, all numerical scenarios below are hypothetical illustrations rather than reported benchmark results.}

\textbf{Research:}
Research in Physical Intelligence follows the canonical scientific process of reviewing prior work, formulating hypotheses, conducting experiments, and validating results~\citep{cooper2017research}. It differs from conventional AI research in two ways that matter for governance. First, the experiments are embodied~\citep{Kober2013RLRobotics}, due to which the results depend on specific hardware, environments, and physical interactions, making reproducibility itself a governance concern. Second, research outputs extend beyond publications to code, datasets, simulation environments, and hardware designs, raising questions of provenance, licensing, and quality assurance~\citep{uhn_research_lifecycle, borgman2019lives}.

\emph{Evidence standards for embodied claims.} The embodiment of AI systems broadens the evidence needed to substantiate experimental claims relative to conventional AI, particularly because performance estimates may depend on repeated trials conducted under shared physical conditions. For evaluations with binary outcomes, a basic distinction is whether the trials can reasonably be treated as independent. When they are sufficiently independent, a policy evaluated over \(n\) trials with \(k\) successes has an estimated success rate \(\hat{p}=k/n\). The observed success rate alone, however, does not capture the uncertainty in this estimate. For a chosen confidence level \(1-\alpha\), this uncertainty can be summarized using the Wilson score interval~\citep{Wilson1927}:

\begin{equation}
\frac{\hat{p} + \frac{z^2}{2n}}{1 + \frac{z^2}{n}} \;\pm\; \frac{z}{1 + \frac{z^2}{n}}\sqrt{\frac{\hat{p}(1-\hat{p})}{n} + \frac{z^2}{4n^2}}, \qquad z = z_{1-\alpha/2},
\label{eq:wilson}
\end{equation}
The Wilson interval generally provides better coverage than the simple normal approximation for small sample sizes and success probabilities near 0 or 1~\citep{brown2001interval}. For a one-sided bound, $z_{1-\alpha/2}$ is replaced by $z_{1-\alpha}$.
Physical experiments, however, may not produce independent trials. Repeated trials within the same experimental session can share lighting, calibration, temperature, object placement, and operator effects, creating dependence between outcomes. In such settings, treating all \(n\) trials as independent can overstate the amount of evidence provided by the experiment. Uncertainty should therefore be estimated in a way that accounts for the session structure, for example by analyzing independent sessions or using a clustered bootstrap. For a simple setting with \(J\) sessions and \(m\) trials per session (\(n=Jm\)), the effect of within-session correlation can also be summarized using the cluster-sampling design effect~\citep{kish1965survey}. Accounting for intra-session correlation, the effective sample size under this cluster-sampling approximation can be expressed as 
\begin{equation}
n_{\mathrm{eff}} \;=\; \frac{n}{1 + (m-1)\,\rho},
\label{eq:neff}
\end{equation}
where $\rho$ is the intra-session correlation and can be estimated from session-level outcomes using a one-way ANOVA estimator~\citep{Ridout1999ICCBinary}. 
\(n_{\mathrm{eff}}\) summarizes the reduction in information caused by within-session correlation. As \(\rho \to 1\), \(n_{\mathrm{eff}} \to J\), meaning that additional trials within the same highly correlated session contribute little new evidence, motivating the collection of additional independent sessions. This distinction matters for uncertainty estimates such as Eq.~\ref{eq:wilson}, which are derived under an independent-trial model; dependence among clustered trials must therefore be accounted for when estimating uncertainty.

Beyond accounting for dependence in collected trials, evaluation design should also justify whether the number of trials is sufficient to support the comparison being made. For an equal-sized comparison between two policies, a useful normal approximation to the number of trials per arm is given by~\citep{Kim2016SampleSizeProportions}:
\begin{equation}
n \;\geq\; \frac{\big(z_{1-\alpha/2} + z_{1-\beta}\big)^{2}\,\big[\,p_1(1-p_1) + p_2(1-p_2)\,\big]}{(p_1 - p_2)^{2}}.
\label{eq:samplesize}
\end{equation}
where $\alpha$ is the significance level and $1-\beta$ the desired statistical power. Substituting $p_1 = 0.60$, $p_2 = 0.70$, $\alpha = 0.05$, and $1-\beta = 0.80$ yields $n \approx 354$ trials per arm. This illustrates that a ten-percentage-point difference observed over only tens of trials may provide weak comparative evidence.

Safety claims pose a different statistical question: how small a failure probability is supported when no failures are observed? For $n$ independent trials with zero observed failures, the rule of three gives the approximate one-sided $95\%$ upper bound~\citep{Hanley1983RuleOfThree}

\begin{equation}
p_{\text{fail}} \;\lesssim\; 3/n.
\label{eq:ruleofthree}
\end{equation}
Thus, an incident-free pilot of $300$ independent runs supports an upper bound of approximately $10^{-2}$, rather than evidence for a substantially smaller failure probability. Dependence between trials further reduces the amount of independent evidence available. For illustration, if the same $300$ runs were collected over $J=15$ sessions with $m=20$ trials per session and an assumed $\rho=0.1$, Eq.~\ref{eq:neff} gives an effective sample size of approximately $103$. Such empirical evidence may still fall far short of the failure rates relevant to high-integrity safety functions~\citep{IEC61508-1-2010}. Physical testing alone may therefore be insufficient to substantiate very rare failure rates and should be complemented by engineering analysis and other assurance evidence. Experimental reports should accordingly include uncertainty estimates, trial counts, and information about session-level dependence alongside headline success rates, making the strength of the evidence easier to assess by reviewers, adopters, and regulators~\citep{pineau2021improving,mirrer2024}.

\emph{The simulation boundary.}
For claims intended to transfer to physical deployment, the same logic applies across the simulation boundary. We define the sim-to-real gap as
$\Delta_{\text{s2r}} = |\hat{p}_{\text{sim}} - \hat{p}_{\text{real}}|$,
corresponding to the instantiation of Eq.~\ref{eq:dep_stagegap} at the research stage. Simulated and real evaluations can differ substantially in some settings~\citep{s2rbench2025}, yet can agree closely when the simulator is aligned to the real system~\citep{realm2025}, and such gaps are a central challenge in sim-to-real transfer~\citep{zhao2020sim}. We therefore recommend reporting $\Delta_{\text{s2r}}$, with its uncertainty, alongside simulation results that are meant to inform physical deployment. Making the gap explicit provides a quantitative measure of sim-to-real transfer that can be tracked across experiments. Furthermore, documenting the simulated and real experimental setups in sufficient detail for other laboratories to repeat the experiment supports reproducibility and replication~\citep{mirrer2024,reproducibleRoboticsArticles2017}. Section~\ref{sec:operationalizing} treats the corresponding cross-laboratory criterion as a governance factor in its own right.

\emph{Where to look and what to release.} Researchers entering the field can operationalize the above through existing infrastructure rather than building their own. As an example, for cross-embodiment generalization in robot manipulation, Open X-Embodiment provides pooled manipulation data across many robot platforms and accompanying RT-X policies~\citep{openXEmbodiment2024}. Real-to-sim pipelines can reconstruct real scenes in simulation and make repeated controlled evaluation more tractable~\citep{robotarena2025,re3sim2025,realToSimToReal2025}. S2R-Bench compares simulated and real corruption robustness for autonomous-driving perception, whereas REALM validates simulation against real-world results for manipulation policies~\citep{s2rbench2025,realm2025}. Neither benchmark is sufficient to establish safety for a different application on its own. On the release side, robotics reproducibility initiatives increasingly encourage the release of software, datasets, and hardware specifications alongside publications~\citep{Bonsignorio2025ReproducibleRobotics,reproducibleRoboticsArticles2017}. Section~\ref{sec:operationalizing} specifies the corresponding disclosure set and identifies the items that have no close analogs in traditional AI reproducibility practice.

\emph{A hypothetical worked research example.} Consider a team that claims a vision-language-action (VLA) policy trained on one manipulator can generalize to a second robot under changes in object classes, lighting, and clutter. The team should first define physical task success and safety violations separately, 
fix the evaluation conditions in advance, and justify that the evaluation includes enough independent evidence to support the intended performance and safety claims (e.g., using Eq.~\ref{eq:samplesize} and Eq.~\ref{eq:ruleofthree}).
The team should then run matched scenarios in simulation and on the target robot (e.g., using Open X-Embodiment data or RT-X models as training or baseline resources~\citep{openXEmbodiment2024}). It should report $\hat{p}_{\text{sim}}$, $\hat{p}_{\text{real}}$, their Wilson intervals, and $\Delta_{\text{s2r}}$, both overall and for the weakest condition, following REALM's validation methods~\citep{realm2025}.
Finally, the release package should contain the policy checkpoint, software environment, robot and end-effector specifications, sensor and firmware versions, calibration results, task definitions, randomization ranges, run-level outcomes, intervention logs, and representative failure videos. The result should advance only when the lower confidence bound on real-world task success exceeds the application threshold and the upper confidence bound on the separately defined safety-violation rate remains within its risk budget. If either condition fails, the team can collect more data for the weak condition, revise the simulator where $\Delta_{\text{s2r}}$ is the largest, or redesign the policy and/or hardware before deployment.

\emph{From research to production.} There is no sharp boundary between research and production, and the transition is iterative. Prior work has proposed a structured framework that bridges scientific research and production systems through staged maturity levels and key transition points, including software readiness, research-to-product transfer, and system deployment readiness~\citep{lavin2022technology}. From a governance perspective, each transition point is a gate with an admissible exit artifact rather than a milestone label. For example, software readiness requires a reproducible build and the disclosure set above; research-to-product transfer requires an estimate of $\Delta_{k \rightarrow k+1}$ measured on the target embodiment rather than the development one; and deployment readiness requires a trial count sufficient to support the claimed failure bound under Eq.~\ref{eq:ruleofthree}. 

\textbf{Design:}
Design in Physical Intelligence is not a stand-alone activity, but an iterative process that spans the entire system life cycle, with design decisions continuously refined through feedback during development, deployment, and operation. The design process typically comprises four interconnected phases: design criteria and specification, conceptual design, detailed design, and road mapping and construction, with iterative feedback loops enabling continuous refinement of system requirements and implementation throughout development \citep{chen2025embodiedsurvey}. What distinguishes design as a governance stage is that design is the highest-leverage or lowest-cost point for allocating enforceable safety controls. Once the physical embodiment, namely, mechanical structure, actuator authority, and sensor placement are fixed, the downstream stages can only mitigate the residual risk that these choices leave behind. Therefore, each phase should produce a concrete exit artifact. The criteria phase produces an operational design domain (ODD) specification, conceptual design produces a hazard analysis and risk assessment, detailed design produces a safety architecture, and road mapping produces a verification plan that maps every requirement to an executable test.

\emph{Quantifying risk before implementation.} Hazard analysis at the conceptual phase can begin with the expected loss associated with each identified hazard $h$,
\begin{equation}
R(h) \;=\; \Pr(h \mid \mathrm{ODD})\,L(h), \qquad
\Pr(h \mid \mathrm{ODD}) \;\approx\; f_{e}(h)\,p_{h}(h)\,\big(1-p_{a}(h)\big),
\label{eq:designrisk}
\end{equation}
where $L(h)$ is the loss or severity if the hazard occurs. The optional factorization uses exposure $f_e$, the conditional probability $p_h$ of a hazardous event during exposure, and the probability $p_a$ that the affected person can avoid harm. It is a screening model only when these terms are estimated as conditional probabilities. Multiplying ordinal risk ranks does not produce a calibrated probability and does not replace the risk assessment required by the applicable standard. Workspace layout changes exposure, sensing and control architecture change $p_h$, robot speed and geometry change loss severity, and warnings or escape space change avoidability. Model quality is therefore one design lever rather than the sole source of risk~\citep{ISO10218,IEC61508-1-2010}. Some hazards arise when the intended function operates in unforeseen conditions rather than when a component fails. This is an important failure mode for learned perception. Such hazards fall under the safety of the intended function framing and require scenario based rather than failure rate based analysis~\citep{ISO21448,Tang2025SOTIFReview,Singh2021SimulationDrivenSafety}.

\emph{Latency is a safety parameter, not only a performance one.} The most concrete illustration of why Physical Intelligence design differs from traditional AI design is the speed and separation monitoring criterion for collaborative operation. It requires that the separation between the human and robot never fall below the protective distance~\citep{ISO15066,Haddadin2016PHRI}
\begin{equation}
S_{p}(t_0) \;=\; S_{h} + S_{r} + S_{s} + C + Z_{d} + Z_{r},
\label{eq:ssm}
\end{equation}
where $S_h$ is the distance the human can travel while the system reacts and stops, $S_r$ the robot's travel during its reaction time $T_r$, $S_s$ its braking distance, $C$ the intrusion distance of a body part, and $Z_d, Z_r$ the position uncertainties of the human-detection system and the robot. Under constant human speed $v_h$, robot speed $v_r$, and deceleration $a_{\max}$, this reduces to
\begin{equation}
S_{p} \;=\; v_{h}\big(T_{r} + T_{s}\big) \;+\; v_{r} T_{r} \;+\; \frac{v_{r}^{2}}{2 a_{\max}} \;+\; C + Z_{d} + Z_{r}, \qquad T_{s} = \frac{v_{r}}{a_{\max}},
\label{eq:ssm_simple}
\end{equation}
and the system reaction time decomposes into
\begin{equation}
T_{r} \;=\; T_{\text{sense}} + T_{\text{perc}} + T_{\text{infer}} + T_{\text{comm}} + T_{\text{act}}.
\label{eq:latency}
\end{equation}
The governance consequence is direct: model inference latency $T_{\text{infer}}$ appears inside a safety related design distance. For illustration, let $v_h = 1.6\,\text{m/s}$, $v_r = 1.0\,\text{m/s}$, $a_{\max} = 2.0\,\text{m/s}^2$, $T_r = 0.2\,\text{s}$, and $C+Z_d+Z_r=0.33\,\text{m}$. Equation~\ref{eq:ssm_simple} then gives $S_p = 1.90\,\text{m}$. Every additional $100\,\text{ms}$ of reaction time widens the required separation by $(v_h + v_r)\Delta T = 0.26\,\text{m}$, which is about $14\%$ of the original separation. These values illustrate the calculation only. An application must use measured stopping behavior, validated sensing uncertainty, the relevant intrusion allowance, and speeds selected through its risk assessment. Because Eq.~\ref{eq:ssm_simple} is quadratic in $v_r$, it can also be inverted to yield the maximum admissible robot speed for a given floor plan. Workspace area, actuator deceleration, sensor coverage, and model latency must therefore be traded jointly. This coupling is much less direct in most screen based AI services, where latency is usually a service quality concern rather than part of a physical safety envelope.

\emph{Contact limits bound commanded velocity.} Where separation cannot be maintained, power and force limiting applies instead. For a transient contact between a human body region of effective stiffness $k$ and mass $m_H$ and a robot of effective mass $m_R$, the reduced mass and the resulting relation between force and velocity are
\begin{equation}
\mu \;=\; \left(\frac{1}{m_{H}} + \frac{1}{m_{R}}\right)^{-1}, \qquad F \;=\; v_{\text{rel}} \sqrt{k\,\mu} \;\; \Longrightarrow \;\; v_{\text{rel}}^{\max} \;=\; \frac{F_{\max}}{\sqrt{k\,\mu}}, \qquad p \;=\; \frac{F}{A} \;\leq\; p_{\max},
\label{eq:pfl}
\end{equation}
with $F_{\max}$ and $p_{\max}$ specified per body region and $A$ the contact area determined by end-effector geometry~\citep{ISO15066,Haddadin2014SafeRobots}. This idealized energy balance provides a preliminary velocity bound and shows why padding, compliant actuation, and rounded end-effector geometry can reduce contact severity. It does not replace application specific contact testing, biomechanical limits, or the risk assessment required by the relevant robot safety standard.

\emph{Design the enforcement layer, not only the policy.} Because a learned policy cannot generally be shown to avoid every unsafe proposal, the design phase should introduce an explicit enforcement layer between policy output and actuation. For control affine dynamics and a safe set $\mathcal{C} = \{x : h(x) \geq 0\}$ encoded by a barrier function $h$, the nominal policy action $u_{\pi}(x)$ can be projected onto the admissible set by solving, at every control step,
\begin{equation}
u^{\star} \;=\; \operatorname*{arg\,min}_{u \in \mathcal{U}} \; \big\| u - u_{\pi}(x) \big\|^{2} \quad \text{s.t.} \quad L_{f}h(x) + L_{g}h(x)\,u \;\geq\; -\,\alpha\big(h(x)\big),
\label{eq:safetyfilter}
\end{equation}
so that the filter, rather than the learned model, enforces the encoded safety property. A simplex architecture that switches to a verified fallback controller when the state approaches the boundary of the recoverable set serves a similar role at the architecture level~\citep{Chen2021SimplexDrive,Bak2021LearningEnabledCPS,Liu2022PhysicsAwarePlanner}. A conformal score can provide a separate trigger for escalation. Given $n$ exchangeable calibration examples with nonconformity scores $s_1,\dots,s_n$, the $\lceil (n+1)(1-\alpha) \rceil$ order statistic defines a threshold $\hat{q}$. Under exchangeability, the next score exceeds this threshold with probability at most approximately $\alpha$~\citep{Luo2024ConformalSafety}. This is a marginal coverage statement, not an automatic bound on unsafe behavior. A safety interpretation additionally requires a score that has been validated against the relevant hazard, and the guarantee may degrade under temporal dependence or distribution shift. The design review should therefore state which hazards are handled by the safety filter, which trigger human escalation, and which remain outside the validated envelope.

\textit{Control interface as a governance boundary.}
For Physical Intelligence systems, the learned model is only one part of the safety-relevant artifact.
The deployable artifact is the closed-loop stack that converts model outputs into physical commands.
In a hierarchical architecture, a vision-language-action model may propose an action chunk, while a tracking or impedance controller converts the selected references into actuator commands~\cite{black2025rtc,frankaControlInterface}.
Building on the enforcement layer above, these commands should be checked against the relevant constraints using time-aligned state estimates before actuation~\cite{wabersich2021predictive}.
We therefore treat this interface as a governance boundary: it makes explicit the conditions under which policy proposals may be applied and the responsibility for enforcing those conditions.

Consider a policy proposal $a_k = \pi_\theta(o_k,g)$ generated at nominal policy rate $f_\pi$, where $o_k$ is the policy observation and $g$ the task goal, while the low-level controller executes at rate $f_c$.
In the hierarchical setting considered here, $f_c$ can substantially exceed $f_\pi$.
The applied command can be represented schematically as
\begin{equation}
    u_t =
    \mathcal{F}\!\left(
        \kappa(\xi_t, a_{\eta(t)}, \hat{p}_t),
        \xi_t,
        \mathcal{E}
    \right),
    \label{eq:controller}
\end{equation}
where $\eta(t)$ identifies the proposal selected for execution at control time $t$, and $\kappa$ includes reference selection or interpolation and the tracking or impedance controller.
The augmented state $\xi_t$ contains the estimated robot and contact state, the previous applied command, timestamps, and any controller or supervisor memory needed for execution.
The parameters $\hat{p}_t$ describe estimated physical properties such as mass, friction, and calibration; $\mathcal{F}$ denotes the safety filter or fallback arbitration layer; and $\mathcal{E}$ specifies the validated operating envelope and associated model and uncertainty assumptions.
Including command history and timestamps makes torque-rate and action-age constraints representable.
This decomposition is not itself a safety guarantee: constraint preservation depends on the filter's assumptions, feasibility, and implementation~\cite{ames2017control,wabersich2021predictive}.

A nominal command should be admitted only when the applicable input, rate, state, contact, and timing conditions are satisfied under those assumptions.
State and contact constraints require an argument about the resulting motion, beyond checking command magnitudes alone~\cite{ames2017control}.
When a proposal expires or a constraint cannot be enforced, the supervisor should invoke a fallback within its validated recovery conditions.
Task success should consequently be reported together with actuator saturation and safety-filter activity.
Frequent intervention can be consistent with the intended architecture; it should be interpreted against the filter's specified role and the operating conditions, rather than treated as evidence of unsafety by itself.

The interface can introduce failures that task-success scores alone do not characterize.
Inference delay and chunk transitions can reduce responsiveness or introduce discontinuities~\cite{black2025rtc}.
Changes in impedance parameters require accounting for their effect on energy exchange and interaction stability; tank-based control provides one established mechanism~\cite{ferraguti2013tank}.
Even interface-level smoothing and rate limiting can alter closed-loop behavior, so their configuration must be included in the evaluated stack~\cite{frankaControlInterface}.
These mechanisms motivate collecting evidence at the boundary between inference and physical execution.

The design review should therefore specify a control-interface contract.
At minimum, we propose recording policy and controller rates, the endpoints of each latency measurement, latency quantiles and observed maxima ( distinguished from justified worst-case bounds), action prediction and execution horizons, observation timestamps and expiration rules, actuator and torque-rate limits, and fallback triggers and recovery conditions.
The contract should also record intervention frequency, duration, and command modification magnitude, with the measurement units and denominators fixed in advance.
For contact-rich systems, it should identify the commanded quantities and controller structure, including any impedance behavior, and the stability or passivity conditions during parameter changes and mode transitions.
Contact-force limits require their own justification; passivity alone does not establish a force bound.

Evidence should include measurements on the target robot and the versioned firmware, controller configuration, calibration, payload, communication setup, and contact conditions used during testing.
Simulation can support this evidence, but the residual transfer error should be assessed for the intended configuration and operating domain.
For example, actuator modeling and latency modeling have been shown to affect the transfer of learned locomotion controllers~\cite{tan2018simtoreal}.

A release rule should require the control-interface criteria to pass alongside task-success criteria.
For each safety-critical scenario cell, the team should report latency quantiles with their estimation uncertainty, deadline misses, action-age violations, saturation, interventions, and fallback outcomes against predeclared criteria.
A 99th-percentile latency below a deadline is a useful performance indicator, but does not establish a hard real-time guarantee.
Where safety depends on a hard deadline, the safety argument must establish the relevant timing bound or demonstrate that a timely independent fallback maintains the specified invariant despite a missed nominal-policy deadline~\cite{buttazzo2011realtime,wabersich2021predictive}.
Saturation and intervention thresholds should reflect the intended architecture; low rates alone do not demonstrate safety.
These measurements connect the design-stage safety argument to the model release gate and deployment monitoring score later in this section.
Together with versioned execution logs, they support tracing failures across the learned model, controller, safety filter, hardware interface, and operating-domain assumptions, including failures caused by interactions among them.

\emph{Co-design under joint budgets.} Physical Intelligence design requires the co-design of AI models, software, and hardware rather than optimizing each component independently. AI design determines the learning algorithms, perception, reasoning, planning, and control capabilities; software design provides the system architecture, middleware, communication, and real-time execution; while hardware design encompasses the mechanical structure, sensors, actuators, embedded computing, and power systems that enable embodied intelligence \citep{roy2021embodied,firoozi2025foundation,coDesigningAccessibleRobots2024}. Three budgets couple these layers quantitatively. The control loop must sample well above the closed-loop bandwidth $f_{\text{bw}}$, conventionally $f_{s} \gtrsim 10\,f_{\text{bw}}$. Many current large multimodal policies cannot run directly at that rate, which motivates a hierarchical split between a slower policy with frequency $f_{\pi}$, often around $5$ to $10\,\text{Hz}$, and a faster low level controller, often around $10^{2}$ to $10^{3}\,\text{Hz}$. Bridging that split with action chunking imposes a hard real-time feasibility condition on the chunk horizon $H$ executed at rate $f_{\text{exec}}$,
\begin{equation}
\frac{H}{f_{\text{exec}}} \;\geq\; T_{\text{infer}} \quad \Longleftrightarrow \quad H \;\geq\; T_{\text{infer}}\, f_{\text{exec}},
\label{eq:chunk}
\end{equation}
so that the next chunk is available before the current one is exhausted; violating Eq.~\ref{eq:chunk} produces control gaps that manifest as jerk or stalls rather than as a degraded benchmark score~\citep{song2025corki}. Finally, the onboard energy budget over a task of duration $T$,
\begin{equation}
E \;=\; \int_{0}^{T} \big(P_{\text{compute}}(t) + P_{\text{actuation}}(t)\big)\,dt, \qquad E_{\text{decision}} \;\approx\; \frac{\overline{P}_{\text{compute}}}{f_{\pi}},
\label{eq:energy}
\end{equation}
couples policy frequency to endurance and thermal headroom, which is why dynamic-inference, pruning, and compression-recovery methods for embodied policies are design-stage rather than deployment-stage decisions: they change $T_{\text{infer}}$, and through Eq.~\ref{eq:ssm_simple} they change the safety envelope~\citep{deerVLA2024,specPruneVLA2025,rlrc2025}.

\emph{Human-centered and governance-by-design requirements.} Human-centered design remains a fundamental principle throughout the design process. Rather than replacing humans during design, Physical Intelligence systems should keep humans in the loop by incorporating stakeholder input, expert evaluation, and iterative human feedback to guide system development and improve safety, usability, and performance \citep{Nilles2018RobotDesign,ISO9241-2102019,doncieux2022human}. As companion and assistive robots become increasingly prevalent, design must extend beyond functional performance to include social interaction, user experience, trust, accessibility, and long-term acceptance. These considerations influence robots' physical appearance, behavior, and interaction strategies, drawing increasingly on insights from psychology, interaction between humans and robots, and artificial intelligence to enable natural and effective collaboration with people \citep{Goodrich2007HRI,Fink2012Anthropomorphism,Cross2026SocialRobotics}. In practice these commitments are discharged through named methods with reportable outputs, which Section~\ref{sec:operationalizing} treats as a governance factor in its own right~\citep{designCenteredHRIGovernance2022,Hornbaek2025DesignEngineering}. Because such commitments are easily stated and rarely verified, we suggest treating governance requirements as first-class design requirements and tracking the fraction of them that carry an executable acceptance test,
\begin{equation}
Cov_{\text{gov}} \;=\; \frac{\big|\{\,g \in \mathcal{G} \;:\; \exists\ \text{an executable test } t(g)\,\}\big|}{|\mathcal{G}|},
\label{eq:govcoverage}
\end{equation}
where $\mathcal{G}$ is the set of governance requirements in the specification. A requirement such as ``the robot shall yield to pedestrians'' scores zero under Eq.~\ref{eq:govcoverage} until it is restated as a measurable condition on $S_p$ from Eq.~\ref{eq:ssm}; the metric thus makes the difference between a governance aspiration and a governance requirement observable at review time~\citep{designCenteredHRIGovernance2022,continuousAssurance2025}.

\begin{sidewaystable}[p]
\centering
\caption{Physical Intelligence Data Taxonomy}
\label{tab:data_taxonomy}
\footnotesize
\renewcommand{\arraystretch}{1.15}
\setlength{\tabcolsep}{5pt}

\newcolumntype{Y}[1]{>{\raggedright\arraybackslash\hsize=#1\hsize}X}

\begin{tabularx}{\textheight}{
  Y{0.60}   
  Y{1.65}   
  Y{1.20}   
  Y{1.00}   
  Y{0.60}   
  Y{0.90}   
  Y{0.70}   
}

\toprule

\textbf{Data Category} &
\textbf{Definition} &
\textbf{Capabilities Enabled} &
\textbf{Applicable Models} &
\textbf{Stage} &
\textbf{Representative Research} &
\textbf{References}
\\

\midrule

\textbf{Foundation \& Human Knowledge Data}
&
Internet-scale vision--language datasets, egocentric videos, language instructions, task annotations, and human demonstrations that provide semantic knowledge and behavioral priors for embodied intelligence.
&
Semantic grounding, visual representation learning, instruction following, affordance learning, commonsense reasoning, and long-horizon planning.
&
Vision--Language Models (VLMs), Vision--Language--Action (VLA) models, LLM planners, video foundation models.
&
Mainly pretraining.
&
PaLM-E, RT-1, OpenVLA, GR00T N1.
&
\citep{driess2023,brohan2022,kim2024,bjorck2025groot}
\\

\midrule

\textbf{Embodied Robot Experience}
&
Robot observation--state--action trajectories collected through teleoperation, tactile sensing, proprioception, autonomous rollouts, and fleet experience.
&
Closed-loop control, dexterous manipulation, continual learning, morphology generalization, and policy adaptation.
&
Robot foundation models, diffusion policies, Vision--Language--Action models.
&
Pretraining, post-training, and continual learning.
&
RT-X, Open X-Embodiment, RoboCat.
&
\citep{openXEmbodiment2024,bousmalis2023,bjorck2025groot,lin2026cotraining}
\\

\midrule

\textbf{Synthetic \& Simulation Data}
&
Physics-engine simulations, digital twins, synthetic demonstrations, world-model rollouts, neural-generated trajectories, and simulated environments.
&
Counterfactual reasoning, scalable policy learning, sim-to-real transfer, rare-event generation, and planning.
&
World models, reinforcement learning, diffusion models, inverse dynamics models.
&
Pretraining, simulation, and post-training.
&
Genie, Isaac Sim, MuJoCo.
&
\citep{bruce2024,long2025embodiedworldmodels,zhao2020sim,aljalbout2026reality}
\\

\midrule

\textbf{Spatial, Semantic \& Reasoning Data}
&
Scene graphs, spatial maps, semantic memory, task graphs, object relations, language grounding, and structured world knowledge.
&
Localization, navigation, spatial reasoning, instruction grounding, hierarchical planning, and task decomposition.
&
Spatial world models, multimodal VLAs, LLM planners.
&
Pretraining and post-training.
&
SayCan, PaLM-E, embodied world models.
&
\citep{ahn2022,driess2023,chen2025emlms,ma2026vla}
\\

\midrule

\textbf{Evaluation, Safety \& Reliability Data}
&
Failure cases, intervention records, recovery trajectories, robustness benchmarks, uncertainty estimation, and safety annotations.
&
Capability evaluation, robustness assessment, uncertainty estimation, safety alignment, and deployment monitoring.
&
Safety critics, runtime monitors, evaluation frameworks.
&
Alignment and evaluation.
&
LIBERO-Safety, OopsieVerse, Safety-CHORES, and safety-instrumented LIBERO or RoboCasa evaluations.
&
\citep{cui2026liberosafety,balaji2026oopsieverse,zhang25safevla,li2026safety}
\\

\midrule

\textbf{Governance \& Ecosystem Data}
&
Policies, regulations, standards, deployment records, patents, investment trends, supply-chain information, incident databases, and ecosystem indicators.
&
Life cycle governance, AI assurance, regulatory compliance, deployment analysis, and ecosystem intelligence.
&
Model-agnostic governance systems.
&
Usually outside model training.
&
UNESCO AI Recommendation, OECD AI Incidents Monitor, CSET, Deloitte.
&
\citep{unesco2021ethics,verwey2026physical,trustmarque2025governance,deloitte2025physicalai}
\\

\bottomrule
\end{tabularx}
\end{sidewaystable}

\emph{A hypothetical worked design review.} Consider a collaborative industrial manipulator that passes components to workers at a station where the maximum available separation along the approach direction is $1.5\,\text{m}$. The illustrative inputs above give a protective distance of $1.9\,\text{m}$, so this proposed layout cannot satisfy the separation requirement. The team can lower robot speed, improve braking, reposition the person detection sensors, restrict the shared workspace, or add a physical barrier. Each alternative changes a term in Eq.~\ref{eq:ssm_simple} and can be compared before hardware is ordered. If contact remains possible, Eq.~\ref{eq:pfl} provides a preliminary bound for reviewing speed and end-effector geometry, followed by the application-specific testing required by the relevant standard. The review should also include operators and maintenance staff, since visibility, emergency-stop access, and the handover procedure affect whether the design is usable in practice. Each selected control becomes an acceptance test. The design leaves this stage only when the operational design domain is explicit, high severity hazards have an enforced control, mandatory governance requirements have full test coverage under Eq.~\ref{eq:govcoverage}, and the responsible owner has accepted the remaining risk~\citep{ISO10218,ISO15066,coDesigningAccessibleRobots2024}.

\textbf{Data:} Data used in Physical Intelligence can be organized into six overlapping categories: foundation and human knowledge data; embodied robot experience; synthetic and simulation data; spatial, semantic, and reasoning data; evaluation, safety, and reliability data; and governance and ecosystem data (Table~\ref{tab:data_taxonomy}). Different modalities enter the life cycle at different points and have distinct limitations. Web-scale video, for example, provides visual and semantic knowledge but usually lacks explicit action labels and measurements of physical interaction. A model trained mainly on passive video may therefore struggle to infer contact dynamics or the consequences of an action~\citep{mccarthy2025internetvideo}. Combining complementary sources can improve both training efficiency and model performance. A data-pyramid strategy, for instance, can co-train on action-labeled robot trajectories and action-free observations~\citep{bjorck2025groot,lin2026cotraining}. Continued collection across tasks, environments, and robot embodiments can then target gaps revealed during evaluation rather than simply increasing the total number of trajectories~\citep{chen2025emlms}.

\emph{From data volume to risk-weighted coverage.} A trajectory count alone is a poor governance measure because robot data are temporally correlated and common operating conditions can overwhelm rare but safety-critical cases. Before collection, a team can define a scenario set $\mathcal{S}$ across the dimensions that delimit its operational design domain. These dimensions may include task, environment, embodiment, user group, and hazard condition. The team then assigns each scenario cell $s$ a risk weight $w_s$ and a minimum evidence requirement $n_{\min,s}$. Dataset coverage can be summarized with a weighted score and a separate gate for safety-critical cells:
\begin{equation}
\mathrm{Cov}_{\mathrm{data}} \;=\; \frac{\sum_{s\in\mathcal{S}} w_s\,\mathbb{I}[n_s \geq n_{\min,s}]}{\sum_{s\in\mathcal{S}} w_s},
\qquad
G_{\mathrm{crit}} \;=\; \min_{s\in\mathcal{S}_{\mathrm{crit}}}\mathbb{I}[n_s \geq n_{\min,s}],
\label{eq:data_coverage}
\end{equation}
where $n_s$ is the number of independent episodes in cell $s$ and $\mathcal{S}_{\mathrm{crit}}$ contains the safety-critical cells. The weighted score shows overall progress, while $G_{\mathrm{crit}}=1$ is a hard release condition. This separation prevents well-covered nominal cells from compensating for an uncovered high-risk cell. The weights should be fixed before evaluation from expected exposure and harm severity. The minimum $n_{\min,s}$ should be justified by a target confidence interval or rare-event bound, such as Eqs.~\ref{eq:wilson} and~\ref{eq:ruleofthree}, rather than chosen after observing the results. Repeated trajectories from one run should not be counted as independent evidence. Teams should use independently initialized sessions where possible and cluster-aware uncertainty estimates when episodes within a session remain correlated. Data quality is therefore assessed by the fitness for the intended use, not by the size of the dataset alone~\citep{Mehak2024DataQuality}.

\emph{Data curation under distribution shift.}
Changes in task mix, operating conditions, or robot configuration can leave previously collected data insufficient for current safety-relevant scenarios.
Coverage-analysis methods provide a starting point for identifying underrepresented combinations of scenario attributes.
Data interventions include systematically varying environmental factors during collection~\cite{gao2024compositional} and selecting demonstrations based on rollout feedback~\cite{agia2025cupid}.
Practitioners should report coverage by collection period and operating condition, relate gaps to the critical cells in Eq.~\eqref{eq:data_coverage}, and compare current exposure with the release baseline using Eq.~\eqref{eq:dep_workload}.
These measurements guide targeted collection and data selection; adequate counts alone do not establish the reliability or safety of the resulting policy.

In a hypothetical hospital-robot example, one critical cell combines nighttime illumination, a wet floor, a crowded corridor, and a wheelchair user; $100{,}000$ empty daytime runs do not cover it. A practical workflow is to (i) publish the scenario matrix, $S_{\mathrm{data}}$, and $G_{\mathrm{crit}}$; (ii) split by site, period, and robot rather than adjacent frames; (iii) target uncovered critical cells while separating real and synthetic counts; and (iv) record configuration, calibration, consent, licenses, annotations, and interventions. Versioned lineage must link each policy to its exact training data and transformations~\citep{privacySensitiveRobotDesign2025,roboLineage2026}. General task suites such as LIBERO, CALVIN, and RoboCasa are not safety evidence without collision, damage, intervention, and constraint-violation labels. LIBERO-Safety and OopsieVerse expose these outcomes but cannot replace application-specific testing~\citep{cui2026liberosafety,balaji2026oopsieverse}. The dataset advances only when $G_{\mathrm{crit}}=1$, provenance and privacy checks pass, and no intolerable residual risk remains; otherwise Eq.~\ref{eq:data_coverage} identifies the next collection target.

\textbf{Model:} The model stage converts curated data into learned components for perception, reasoning, planning, action generation, and control (Table~\ref{tab:model_taxonomy})~\citep{gaba2026physical,firoozi2025foundation}. The categories in the table describe complementary, overlapping roles and interfaces rather than a fixed sequence of modules. Vision--language foundation models provide semantic representations that align visual observations with language, while visual and spatial representation models emphasize object, geometric and spatial structure. Embodied reasoning models provide task decomposition, skill selection, and feedback-based replanning.

Robot foundation models are distinguished primarily by their broad training and transfer across robotic tasks, environments and embodiments, whereas vision--language--action (VLA) models are defined by their interface: they map visual observations, language instructions and optionally robot state to actions or action-oriented representations. Their outputs can include waypoints or intermediate spatial and control representations for downstream execution. Across model families, action generation combines output representations, such as continuous trajectories, discrete action tokens (including frequency-domain encodings), and latent actions, with generation mechanisms such as autoregressive decoding, diffusion, or flow matching~\citep{pertsch2025fast,ye2024lapa,chi2023diffusionpolicy,black2024pi0}. Actions may be generated in chunks or organized hierarchically into subtasks and lower-level commands~\citep{zhao2023act,black2025pi05}. For example, the same model may be both a robot foundation model and a vision--language--action model, while using flow matching to generate action chunks. In humanoids and mobile manipulation systems, learned low-level or whole-body controllers can operate beneath VLA or reasoning models, translating motion targets and sensory feedback into motor commands for locomotion, balance, contact handling, and coordinated manipulation. Examples include Helix 02's System 0, which tracks whole-body targets, and WholeBodyVLA's loco-manipulation-oriented reinforcement learning policy, which executes locomotion commands~\citep{figure2026helix02,jiang2025wholebodyvla}.

Within world and dynamics models, action-free predictive representations capture temporal structure without conditioning on actions and can support downstream control, while action-conditioned models explicitly predict future states or observations under candidate actions for planning and policy learning~\citep{hafner2023dreamerv3}. For example, V-JEPA 2 learns action-free predictive representations from video, whereas V-JEPA 2-AC adds action-conditioned post-training on robot trajectories to predict latent outcomes of candidate actions for robotic planning~\citep{assran2025vjepa2}. Generative world models synthesize visual environments or possible futures for simulation and synthetic experience generation, and may also be action-conditioned~\citep{deepmind2025genie3,nvidia2025cosmos}. Newer world--action models, such as DreamZero, jointly predict future observations and executable actions, linking world modeling with action generation~\citep{ye2026dreamzero}.

Physical-system modeling and simulation models approximate physical processes for prediction, simulation, and system identification, using physics-informed neural networks, neural operators, or hybrid analytical--learned dynamics~\citep{raissi2017pinn,li2020fno,lu2019deeponet,heiden2020neuralsim}. Physical knowledge can enter through equation-based training constraints, structured mechanics priors, or analytical dynamics augmented by learned residuals, while differentiable simulators enable gradient-based optimization of model parameters and policies~\citep{raissi2017pinn,greydanus2019hamiltonian,heiden2020neuralsim,pan2025learningfly}. This category overlaps with world and dynamics models but also covers scientific and engineering simulation beyond embodied planning; neural operators need not explicitly enforce physical laws.

Architecture and model development establish the network structure, interfaces, learning objective, and action representation. Pretraining can build general representations from broad data, while mid-training specializes those representations to a target task, environment, or embodiment~\citep{kawaharazuka2025vla,kim2024,tu2025midtraining}. Post-training can further adapt behavior through task-specific fine-tuning, preference optimization~\citep{rafailov2023direct, mahmud2025maple}, or safety-focused training~\citep{kawaharazuka2025vla}. Prototyping and sim-to-real transfer expose discrepancies between simulated and physical execution~\citep{zhao2020sim,bousmalis2023}. Optimization then addresses latency, memory, energy, and hardware constraints, and testing determines whether the complete system meets the release criteria~\citep{aljalbout2026reality,yi2026dnntesting}. These activities are iterative rather than a universal linear sequence. A failed physical test may send the team back to data collection, architecture changes, or safety-controller design.

\emph{Release gates for the complete system.} Average task success is insufficient for selecting a Physical Intelligence model because a gain on nominal tasks can hide unsafe actions in a rare scenario or a control-loop deadline miss. Using the safety-critical cells $\mathcal{S}_{\mathrm{crit}}\subseteq\mathcal{S}$ defined during data curation, a candidate model should be admitted only if
\begin{equation}
\begin{aligned}
\min_{s\in\mathcal{S}_{\mathrm{crit}}}\mathrm{LCB}_{1-\alpha}\!\left[p_{\mathrm{succ}}(s)\right] &\geq \tau_{\mathrm{succ}}, \\
\max_{s\in\mathcal{S}_{\mathrm{crit}}}\mathrm{UCB}_{1-\alpha}\!\left[p_{\mathrm{unsafe}}(s)\right] &\leq \tau_{\mathrm{unsafe}}, \\
\max_{s\in\mathcal{S}_{\mathrm{crit}}} Q_{0.99}\!\left[T_{\mathrm{loop}}(s)\right] &\leq T_{\mathrm{deadline}}
\end{aligned}
\label{eq:model_gate}
\end{equation}
where $T_{\mathrm{loop}}$ includes sensing, inference, communication, and actuation, all thresholds are fixed from the application risk analysis before model comparison, and $p_{\mathrm{succ}}(s)$ denotes the task-success probability for critical cell $s$. Equation~\ref{eq:wilson} can be used for binary outcomes when trials are sufficiently independent. Correlated robot trials require session-level or cluster-aware bounds, as discussed in the Research component. The 99th-percentile latency estimate also needs enough observations in each critical cell and should be accompanied by uncertainty when the sample is small. All three conditions must pass: performance in an easy cell cannot compensate for a safety or timing violation in a critical one. Evaluation should additionally record human interventions, recovery success, energy per episode, and the sim-to-real gap. Task success and safe execution should be reported separately, using safety-specific benchmarks where applicable, but these diagnostic measures do not replace the release conditions~\citep{li2026safety,yi2026dnntesting,Kojima2025PhysicalRiskControl,cui2026liberosafety,balaji2026oopsieverse}.

For a hypothetical warehouse-picking decision, suppose that each model is evaluated in $1{,}000$ sufficiently independent human-present trials. Model A reaches $94\%$ average task success, produces $12$ unsafe actions, and has $180\,\mathrm{ms}$ 99th-percentile loop latency. Model B reaches $91\%$ average success, produces $2$ unsafe actions, and has $60\,\mathrm{ms}$ latency. The observed unsafe-action rates are $1.2\%$ and $0.2\%$, but the release gate uses uncertainty rather than these point estimates. A one-sided $95\%$ Wilson calculation gives upper bounds of approximately $1.91\%$ for A and $0.60\%$ for B. With $\tau_{\mathrm{unsafe}}=1\%$ and a $100\,\mathrm{ms}$ deadline, A fails both the safety and timing gates. B passes those two gates but remains ineligible until its minimum per-cell success bound also passes. 

\begin{table}[H]
\centering
\caption{Physical Intelligence Model Taxonomy.}
\label{tab:model_taxonomy}

\small
\renewcommand{\arraystretch}{1.35}
\setlength{\tabcolsep}{5pt}

\begin{tabularx}{\textwidth}{
>{\raggedright\arraybackslash}p{3cm}
>{\raggedright\arraybackslash}p{5.3cm}
>{\raggedright\arraybackslash}X
>{\raggedright\arraybackslash}p{2.0cm}
}

\toprule

\textbf{Category}
&
\textbf{Representative Models}
&
\textbf{Primary Role}
&
\textbf{References}
\\

\midrule

\textbf{Vision--Language Foundation Models}
&
CLIP, SigLIP, LLaVA, PaLI-X, PaliGemma 2, Qwen3-VL
&
Provide aligned visual--language representations and semantic knowledge to support embodied perception, instruction interpretation, and reasoning.
&
\citep{radford2021clip,zhai2023siglip,liu2023llava,steiner2024paligemma2,bai2025qwen3vl,driess2023}
\\

\midrule

\textbf{Visual \& Spatial Representation Models}
&
DINOv3, SAM~3.1, Grounding DINO, VGGT, Depth Anything 3, R3M, MVP
&
Extract visual features, object information, and scene geometry to support recognition, segmentation, grounding, localization, and manipulation.
&
\citep{simeoni2025dinov3,kirillov2023sam,liu2023groundingdino,wang2025vggt,lin2025depthanything3,firoozi2025foundation,xu2024,duan2024}
\\

\midrule

\textbf{Robot Foundation Policies}
&
Octo, RoboCat, RT-1, RT-X, RT-H, GR00T N1, GR00T~1.7, $\pi_{0}$, $\pi_{0.5}$, Gemini Robotics, MolmoAct~2
&
Learn reusable manipulation and locomotion skills from broad, multi-embodiment robot datasets to support cross-task generalization and adaptation across tasks, environments, and embodiments.
&
\citep{ghosh2024octo,bousmalis2023,brohan2022,lin2026cotraining,bjorck2025groot,nvidia2026groot17,black2024pi0,black2025pi05,geminiRobotics2025,fang2026molmoact2,firoozi2025foundation}
\\

\midrule

\textbf{Vision--Language--Action Models}
&
RT-2, OpenVLA, OpenVLA-OFT, $\pi_{0}$, $\pi_{0.5}$, CogACT, GR00T N1, GR00T~1.7, MolmoAct~2, Gemini Robotics, Gemini Robotics~2, Helix
&
Map visual observations, language instructions, and optional robot states to actions or intermediate action representations for execution through downstream decoders or controllers.
&
\citep{zitkovich2023rt2,kim2024,kim2025openvlaoft,li2024cogact,bjorck2025groot,nvidia2026groot17,black2024pi0,black2025pi05,fang2026molmoact2,geminiRobotics2025,deepmind2026geminirobotics2,figure2025helix,ye2024lapa,figure2026helix02,jiang2025wholebodyvla}
\\

\midrule

\textbf{World \& Dynamics Models}
&
DreamerV3, V-JEPA~2, V-JEPA~2-AC, Genie~3, Cosmos-Predict2, Cosmos Predict, DreamZero, Video World Models, Robot Dynamics Models
&
Learn action-free predictive representations, predict action-conditioned dynamics, generate possible environments and futures, or jointly predict observations and actions.
&
\citep{hafner2023dreamerv3,bruce2024,assran2025vjepa2,deepmind2025genie3,nvidia2025cosmos,ye2026dreamzero,kim2026cosmospolicy,long2025embodiedworldmodels,gaba2026physical}
\\

\midrule

\textbf{Learning/ Adaptation Method} &
$\pi^{*}_{0.6}$ (RL post-training); RoboTTT (test-time training); Skild S1, Generalist GEN-1.5 (in-context/physical prompting) & Adapt policy behavior after pretraining through reward-based reinforcement learning, partial online updates at inference time, or in-context demonstrations, to support post-deployment correction and rapid task acquisition without full retraining. &
\citep{pi_star_0.6_2025, jiang2026robottt, skild2026s1, generalist2026gen15} \\

\midrule

\textbf{Deployment \& Efficiency} &
Shallow-$\pi$; SpecPrune-VLA, RLRC; DeeR-VLA (dynamic depth/early exiting); LightDP; SmolVLA&
Reduces a model's latency, memory, and compute cost for real-time control on resource-limited robotic hardware, via pruning, quantization, dynamic-depth inference, asynchronous pipelines, or on-device deployment. &
\citep{shallowpi2026, specPruneVLA2025, rlrc2025, yue2024deervla, shukor2025smolvla, deepmind2026geminirobotics2} \\

\midrule

\textbf{Physical-System Modeling \& Simulation Models}
&
PINNs, Fourier Neural Operator, DeepONet, Hamiltonian Neural Networks, hybrid residual dynamics models
&
Approximate physical processes for prediction, simulation, and system identification using learned surrogates, physical constraints, or combined analytical and learned dynamics.
&
\citep{raissi2017pinn,li2020fno,lu2019deeponet,greydanus2019hamiltonian,heiden2020neuralsim,pan2025learningfly}
\\

\midrule

\textbf{Embodied Reasoning \& Planning Models}
&
SayCan, Inner Monologue, PaLM-E, RoboBrain~2.0, Gemini Robotics ER~2, MolmoAct~2, Cosmos Reason~2, ThinkAct, STEER, Hume
&
Support goal interpretation, spatial reasoning, task decomposition, skill selection, and planning or replanning from execution feedback.
&
\citep{ahn2022,huang2022inner,driess2023,cao2025robobrain2,deepmind2026geminiroboticser2,geminiRobotics2025,fang2026molmoact2,nvidia2025cosmosreason2,liang2025}
\\

\midrule

\textbf{Physics-Informed \& Hybrid Dynamics Models}
&
PINNs, Hamiltonian Neural Networks, Neural ODEs, Fourier Neural Operator (FNO), DeepONet, DiffTaichi, Brax, MuJoCo MJX, AC-HGN
&
Incorporate physical laws, governing equations, conservation principles, differentiable physics, or learned dynamics into model learning.
&
\citep{raissi2019physics,lutter2019deep,greydanus2019hamiltonian,
cranmer2020lagrangian,hu2020difftaichi,troch2025action}
\\

\midrule

\textbf{Low-Level Physical Control Models}
&
ANYmal Neural Control Policy, DeepMimic, Neural Motor Policies, Neural Inverse-Dynamics Models, Physics-Based Control Policies & Translate physical states, desired motion, or high-level commands into executable motor actions, joint trajectories, forces, or torques while accounting for system dynamics, feedback, and real-time physical constraints.
&
\citep{hwangbo2019learning,peng2018deepmimic}
\\

\bottomrule

\end{tabularx}

\end{table}

\begin{samepage}
\definecolor{revblue}{RGB}{20,50,180}
\definecolor{revred}{RGB}{150,35,35}
\providecommand{\prov}[1]{{\footnotesize\textsf{\textcolor{revred}{[#1]}}}\ }

Teams can implement the gate in three steps. First, evaluate the incumbent and candidate on the same versioned scenario matrix. Second, stress-test sensor dropout, delayed observations, ambiguous instructions, unseen objects, and actuator saturation. Third, attach the per-cell results, model checkpoint, training-data version, hardware target, and threshold decisions to a release record. Model validation then becomes an auditable deployment decision rather than a leaderboard comparison.

\textbf{Deployment:} The deployment stage integrates a validated model into a complete physical system and manages that system throughout operation. System integration and end-to-end testing connect the model to sensing, communication, safety logic, and actuation~\citep{roy2021embodied,li2026safety}. Hardware commissioning calibrates sensors and actuators, while system identification estimates properties such as dynamics, friction, and mass distribution for the actual robot~\citep{aljalbout2026reality}. Precommissioning simulation and controller tests can expose integration failures before physical trials~\citep{long2025embodiedworldmodels,brohan2022}. Policy evaluation then repeats the safety-critical scenarios on the target hardware and in the intended operating environment. After release, operational monitoring, maintenance, and controlled updates are required because wear, calibration drift, firmware changes, and environmental changes can invalidate earlier evidence~\citep{thakur2025physical,neupane2024security}. A change to any safety-relevant component creates a new deployable unit that must be assessed against the approved configuration and operating domain.

\textbf{The role of deployment in the life cycle}

Everything Research, Design, Data and Model establish is established under conditions: a particular configuration, a particular operating environment, a particular way of sampling. Deployment is where
those conditions can finally be checked against what the system meets. Episodes turn out to be dependent, or their outcomes unresolved; design parameters move with the hardware and the site; the workload in the field is not the workload that was evaluated; and the distance to a release criterion shrinks. This is the reasoning behind treating risk as time-dependent. It does not mean that every earlier claim lapses on the day of deployment --- only that none of them is self-renewing.
\begingroup
\setlength{\intextsep}{10pt}
\begin{table}[H]
\centering
\caption{The six deployment concerns: the assumption each one revisits, and
the measurement deployment reports.}
\label{tab:dep_concerns}
\small
\setlength{\tabcolsep}{5pt}
\renewcommand{\arraystretch}{1.2}
\begin{tabularx}{\textwidth}{@{}>{\raggedright\arraybackslash}p{3.7cm}
                               >{\raggedright\arraybackslash}p{1.5cm}
                               >{\raggedright\arraybackslash}X
                               >{\raggedright\arraybackslash}X@{}}
\toprule
\textbf{Concern} & \textbf{Stage} & \textbf{Assumption} &
\textbf{Deployment measurement} \\
\midrule
Episodes ended by intervention & Research &
Every episode outcome is observed &
Completion range over resolved and unresolved outcomes;
Eq.~\ref{eq:dep_completion} \\
\addlinespace[3pt]
Evidence transfer to the fleet & Research &
Pilot evidence describes fleet operation &
Stage and site gaps for the deployed configuration;
Eq.~\ref{eq:dep_stagegap} \\
\addlinespace[3pt]
Protective separation in operation & Design &
Protection parameters hold at their design values &
Required separation recomputed on measured parameters, and the separation
margin remaining; Eq.~\ref{eq:dep_separation} \\
\addlinespace[3pt]
Changing site conditions & Design &
Committed site conditions continue to hold &
Condition checks against the declared acceptable range \\
\addlinespace[3pt]
Change in the deployed workload & Data &
The evaluated exposure mix describes operation &
Workload divergence and unassessed cells; Eq.~\ref{eq:dep_workload} \\
\addlinespace[3pt]
Release criteria in operation & Model &
Meeting the release criteria settles the question &
Release margin to each criterion, reported separately;
Eqs.~\ref{eq:model_gate}, \ref{eq:dep_margins} \\
\bottomrule
\end{tabularx}
\end{table}
\endgroup
Six concerns recur often enough to be worth treating individually, and
Table~\ref{tab:dep_concerns} collects them. Each takes its name from what deployment observes and
sits with the stage whose claim it informs. They overlap, and they are
certainly not a complete account of how deployments fail. Each is set out
below in the same way: the assumption being revisited, the measurement that
replaces it, and an example.

\textbf{Episodes ended by intervention}

\emph{Assumption.} A supervised system and the policy running inside it are
not the same object, and operational evaluation has to keep them apart. When
an operator takes over, or a fallback controller fires, the intervention may
well have been the right call; what it leaves behind is an episode whose
outcome nobody observed. Episodes from the same robot, the same shift, or the
same site may also be dependent. Both facts bear on Eqs.~\ref{eq:wilson}--\ref{eq:ruleofthree}, which hold
only under the sampling assumptions they state. Reported experience points
the same way: disengagement records vary with reporting practice and
operating conditions~\citep{favaro2018disengagement}, and a rare-event safety
claim resting on observed mileage alone can require more exposure than a
programme will ever accumulate~\citep{kalra2016driving}.

\emph{Deployment measurement.} Begin by fixing what counts as an adverse
outcome: a binary event over a fixed episode horizon, for a named nominal
configuration, with no protective intervention. Of $n$ evaluated episodes,
let $k$ end in a known adverse outcome and $m$ end unresolved after an
intervention. The two counts are disjoint, and every remaining episode is
known to be non-adverse for this target. The fraction of adverse outcomes in
the sample is then compatible with any value in the completion range
\begin{equation}
  \mathcal{I}_n = \left[\, \frac{k}{n},\; \frac{k+m}{n} \,\right].
\label{eq:dep_completion}
\end{equation}
What the range expresses is uncertainty about outcomes that were never seen~\citep{manski2003partial}, and three things about it are worth saying wherever
it is reported. Its width is $m/n$, which coincides with an intervention rate
only when the intervention definition, the episode denominator and the
unresolved count all refer to the same thing. It is not a confidence interval
for the population failure probability. And either endpoint may turn out to
equal the fraction a complete sample would have given.

\emph{Example.} With $n = 500$, $k = 0$ and $m = 12$, Eq.~\ref{eq:dep_completion} gives
$[0,\,0.024]$. Treating every unresolved episode as adverse yields a
conservative population bound, provided the episodes can be modelled as
independent and identically distributed and the outcome classification is
sound: the one-sided exact $95\%$ binomial upper bound for $12$ adverse
outcomes in $500$ episodes is about $0.0386$. Correlated or non-stationary
operation calls for a dependence-aware or time-indexed analysis instead.
Adjudication can move episodes out of the unresolved count, but only where
the assumptions that identify the counterfactual are themselves justified;
calling an intervention precautionary does not establish that the outcome
would have been safe. The counts, the basis for any adjudication, and the
uncertainty analysis go back to Research, where they inform the evidence
statement and the design of the next evaluation.

\textbf{Evidence transfer to the fleet} 

\emph{Assumption.} Evidence is carried from one stage to the next, and the deployed fleet is simply the last of those stages. Pilot evidence therefore has to be reassessed once the fleet exists, with the configuration, the metric and the operating scope all named.
Sim-to-real research supplies the motivation for examining residual transfer
discrepancies~\citep{zhao2020sim}; a fleet assessment carries that examination into
routine operation and into the variation between sites. Where those
differences bear on the claim, a fleet average should travel with the
site-specific estimates and the exposure weights behind it.

\emph{Deployment measurement.} For a fixed policy and metric, write $\mu_j$ for
the expected metric at stage $j$. The absolute endpoint difference satisfies
\begin{equation}
  |\mu_0 - \mu_K| \;\leq\; \sum_{j=0}^{K-1} |\mu_j - \mu_{j+1}|.
\label{eq:dep_stagegap}
\end{equation}
The sum of the stage gaps bounds the endpoint difference; it does not in
general equal it, since a gain at one handoff can offset a loss at another.
Sampling uncertainty belongs in the report as its own quantity, and where the
policy or the embodiment changed between stages, that should be said rather
than folded into the distribution gap.

\emph{Example.} Success estimates of $0.94$, $0.91$, $0.89$ and $0.82$ give
successive gaps of $0.03$, $0.02$ and $0.07$. Here the sum happens to equal
the endpoint difference of $0.12$, but only because the estimates fall
monotonically. Site estimates of $0.88$, $0.86$ and $0.72$ average to the
fleet figure of $0.82$ under equal exposure weights. The third site is the
one to look at --- its workload, its infrastructure, its hardware, and how
the policy meets that environment --- although the number alone does not say
which of these is responsible. Stage and site comparisons return to Research,
where they guide evaluation design and set how far upstream evidence can be
carried forward.

\textbf{Protective separation in operation}

\emph{Assumption.} The protective-separation calculation of Eqs.~\ref{eq:ssm}--\ref{eq:latency}
rests on reaction time, stopping behavior, approach speeds, and sensing
uncertainty. None of these stays fixed in operation: thermal load, payload,
surface condition, wear and sensor contamination all move them. Perception
evaluations under sensor anomalies and environmental variation show how far
nominal sensing performance can sit from what a given condition delivers~\citep{s2rbench2025}. Operational measurements therefore belong against validated
parameter envelopes, uncertainty included, rather than being substituted into
the calculation as exact physical values.

\emph{Deployment measurement.} In the constant-deceleration approximation, let
$\overline{T}_r(t)$ bound reaction time from above, let
$\underline{a}_b(t) > 0$ be a validated lower bound on available braking
deceleration, and let $v_h(t)$ and $v_r(t)$ bound the relevant approach
speeds over the response horizon. Define
\begin{equation}
\begin{aligned}
  \overline{T}_s(t) &= \frac{v_r(t)}{\underline{a}_b(t)}, \\[2pt]
  S_p^{\mathrm{cons}}(t) &= v_h(t)\big(\overline{T}_r(t)
     + \overline{T}_s(t)\big) + v_r(t)\overline{T}_r(t)
     + \frac{v_r(t)^2}{2\underline{a}_b(t)} + C
     + \overline{Z}_d(t) + \overline{Z}_r(t), \\[2pt]
  M_{\mathrm{cons}}(t) &= \hat{d}(t) - S_p^{\mathrm{cons}}(t).
\end{aligned}
\label{eq:dep_separation}
\end{equation}
Here $\hat{d}(t)$ is the measured separation between human and robot, and
$M_{\mathrm{cons}}(t)$ is what remains of it once the required separation has
been subtracted: the clearance that the deployment actually has, in meters. The
$Z$ terms carry the corresponding position uncertainties, counted once, and
the superscript marks the calculation as conservative within the stated model
and the validated bounds. Where constant deceleration is not a good enough
approximation, application-specific stopping measurements are needed. Keeping
this quantity non-negative is a control objective, and sampling it at
intervals and finding it non-negative is not the same as showing that it
never fell below zero.

\emph{Example.} With $v_h = 1.6\,\mathrm{m/s}$,
$v_r = 1.0\,\mathrm{m/s}$, $\underline{a}_b = 2.0\,\mathrm{m/s^2}$,
$\overline{T}_r = 0.2\,\mathrm{s}$, and
$C + \overline{Z}_d + \overline{Z}_r = 0.33\,\mathrm{m}$, the required
separation is $1.90\,\mathrm{m}$. Increasing reaction time by
$60\,\mathrm{ms}$, reducing braking deceleration to $1.2\,\mathrm{m/s^2}$,
and increasing the uncertainty allowance by $0.10\,\mathrm{m}$ gives
approximately $2.86\,\mathrm{m}$. The increase of $0.956\,\mathrm{m}$
comprises $0.70\,\mathrm{m}$ from the reduced braking deceleration,
$0.156\,\mathrm{m}$ from the longer reaction time, and $0.10\,\mathrm{m}$
from the sensing-uncertainty allowance. At a measured separation of
$2.0\,\mathrm{m}$ the clearance is about $-0.86\,\mathrm{m}$: the robot is
closer than the conditions of the day allow. Compared against the original
requirement of $1.90\,\mathrm{m}$, the same measurement looks acceptable.
Validated parameter envelopes and observed degradation return to Design,
where they inform protection, maintenance intervals and operating
restrictions.

\textbf{Changing site conditions}

\emph{Assumption.} Some operating-domain conditions are not the deploying
organization's to hold. A facility operator may change a lift interface, a
floor treatment or an access arrangement; lighting, surface condition and
wireless performance can also drift with no single event to mark the change.
Facility and asset management supply the change-management processes for the
first kind~\citep{iso41001},~\citep{iso55000}, while wireless coexistence and
lighting maintenance practice illustrate the second, where the dependency has
to be assessed on a schedule rather than at a moment~\citep{iec62657},~\citep{cie97}. Change that is uncoordinated across an organizational boundary
can affect system-level safety while every component continues to behave as
its own owner intends~\citep{leveson2011}.
Infrastructure-support classifications in automated driving~\citep{carreras2018isad}, work on public-area mobile robots~\citep{isotr4448}, and
operational-design-domain taxonomies~\citep{iso34503},~\citep{pas1883} provide ways
to describe relevant conditions.

Where a safety claim depends on such a condition, the description has to
become a verifiable requirement. For each dependency the deployment record
should carry the acceptable range, the responsible party, the verification
method, the review interval and the notification arrangement for planned
changes. None of this establishes that the physical condition holds. It
establishes who owes what, and how a breach would come to light.

\emph{Deployment measurement.} Record the result of each condition check and
the claim it bears on. A planned change calls for an impact assessment and
for communication; gradual deterioration calls for measurement, periodic or
continuous as feasibility allows. The two overlap more often than the
distinction suggests, since a change of cleaning product is a discrete event
whose effect on friction then develops slowly. Where a condition cannot be
measured directly, the record should say what evidence stands in its place
and what operating restriction accompanies it. The threshold rule of
Eq.~\ref{eq:dep_threshold} applies only where the condition can be written as a degradation
variable with a justified rate bound and a response model.

\emph{Example.} A fleet is commissioned in a building whose facilities
contractor has committed to a lift interface, a floor finish, and wireless
coverage above a stated level. Over two quarters, wireless performance
declines as further devices are added to the shared space, the cleaning
contractor changes product, and a lift firmware update alters a handshake
timeout. Only the third is a discrete event that a notification arrangement
would have caught; the first two show up only if someone measures them.
Comparing performance across sites stratified by condition can point the
investigation in the right direction, but it does not isolate an
environmental cause unless configuration, workload and operator differences
are controlled for. Condition-check results and the performance evidence
beside them return to Design and to the site operator, who between them
restore the condition, extend the evaluation, or revise the operating
domain.

\textbf{Change in the deployed workload}

\emph{Assumption.} Equation~\ref{eq:reliability_gate} keeps two things apart: an
exposure-weighted aggregate success score, and the minimum critical-cell
reliability gate. Routes get extended, schedules change, task allocation
shifts and seasons turn, and any of these moves the exposure distribution
while performance within each cell stays exactly where it was. A monitor
watching sensor statistics alone will not necessarily register it~\citep{subbaswamy2019preventing}.
What deployment has to retain is explicit information about which tasks and
which operating conditions actually occurred.

\emph{Deployment measurement.} Let $\mathcal{S}$ be a common, mutually exclusive
and exhaustive partition of episodes into scenario cells, including an
unassessed category where needed. For the release exposure weights
$\omega_s^{\mathrm{rel}}$ and the observed weights $\omega_s(W)$ over a
declared window $W$, define
\begin{equation}
  \Delta_{\exp}(W) = \frac{1}{2} \sum_{s \in \mathcal{S}}
  \big| \omega_s(W) - \omega_s^{\mathrm{rel}} \big| \in [0,1].
\label{eq:dep_workload}
\end{equation}
Both weight vectors sum to one and use the same exposure unit, and the
superscript marks the release baseline without implying a formal
certification. Divergence says that exposure has moved. It is not itself a
measure of risk, and where the affected cells perform equally well the
aggregate score need not move at all.

\emph{Example.} Take four disjoint scenario cells with success rates
$(0.98,\,0.95,\,0.86,\,0.94)$. Move the exposure weights from
$(0.60,\,0.20,\,0.05,\,0.15)$ to $(0.40,\,0.25,\,0.20,\,0.15)$ and
$\Delta_{\exp}$ is $0.20$, while the aggregate score $\bar p(v)$ of Eq.~\ref{eq:reliability_gate}
falls from $0.962$ to $0.9425$. Against a separate service-performance
criterion of $0.95$ the fleet is now below the line, and nothing about the
policy or the hardware has changed. What has not happened is a failure of the
critical-cell gate $G_{\mathrm{rel}}(v)$ in the same equation; the cell at
$0.86$ could never have met a $0.95$ critical-cell requirement in the first
place, so the gate was never what that number was speaking to.

A cell being operated with $n_s = 0$ is a different matter, and calls for a
review of both the evidence and the operating domain. Where that cell belongs
to the assessed critical set and carries a positive minimum evidence
requirement, $G_{\mathrm{crit}}$ goes to zero under Eq.~\ref{eq:data_coverage}. Where the cell
is newly identified, the critical set and the operating scope have to be
reassessed before anything else. Either way the question is whether operation
continues, not whether the historical release result was wrong. Workload
divergence and the coverage gaps beside it return to Data, which targets the
next collection, and to Model, which reassesses the claims affected.

\textbf{Release criteria in operation}

\emph{Assumption.} Equation~\ref{eq:model_gate} asks that success, unsafe-action and
timing criteria hold together across the safety-critical cells. What it
returns is a decision, and a decision says nothing about how close the system
came. Keeping the underlying estimates, their uncertainty and the distance to
each threshold makes the next reassessment far more informative, and is what
allows a safety case to be maintained rather than re-derived~\citep{denney2015dynamic}. That distance is evidence about the present
assessment only; it carries no information about how long the criterion will
go on being satisfied.

\emph{Deployment measurement.} For each critical cell $s$ and assessment
window $W$, retain the distance to each of the three criteria of Eq.~\ref{eq:model_gate}
separately,
\begin{equation}
\begin{aligned}
  m_{\mathrm{succ},s}(W) &= \mathrm{LCB}_{1-\alpha}\!\left[p_{\mathrm{succ}}(s;W)\right]
    - \tau_{\mathrm{succ}}, \\
  m_{\mathrm{unsafe},s}(W) &= \tau_{\mathrm{unsafe}}
    - \mathrm{UCB}_{1-\alpha}\!\left[p_{\mathrm{unsafe}}(s;W)\right], \\
  m_{\mathrm{T},s}(W) &= T_{\mathrm{deadline}}
    - Q_{0.99}\!\left[T_{\mathrm{loop}}(s);W\right].
\end{aligned}
\label{eq:dep_margins}
\end{equation}
Each quantity is the distance from a current estimate to the criterion the
configuration was released against, which is why the two probability margins
are in percentage points and the timing margin in milliseconds. They should
be computed on comparable outcome definitions and with justified uncertainty
procedures, and they should be reported one by one: a raw minimum taken
across quantities in different units means nothing. A margin of zero still
satisfies the weak inequality, though no headroom is left. A negative margin,
or evidence too thin to support one, sends the configuration back through the
release procedure.

\emph{Example.} Consider one safety-critical cell released against
$\tau_{\mathrm{unsafe}} = 1\%$ and $T_{\mathrm{deadline}} = 100\,\mathrm{ms}$.
The release evidence for that cell was $4$ unsafe actions in $1500$ episodes,
giving a one-sided exact $95\%$ upper bound of $0.61\%$ and hence
$m_{\mathrm{unsafe}} = 0.39$ percentage points; the 99th-percentile loop latency was
$60\,\mathrm{ms}$, giving $m_{\mathrm{T}} = 40\,\mathrm{ms}$. Reassessment over a later
window records $10$ unsafe actions in $2000$ episodes, an upper bound of
$0.85\%$ and $m_{\mathrm{unsafe}} = 0.15$ percentage points, with a latency quantile of
$88\,\mathrm{ms}$ and $m_{\mathrm{T}} = 12\,\mathrm{ms}$. Both criteria are still
satisfied and no gate was re-run, yet on both criteria the system now sits
closer to the line than it did at release. The two figures are in different
units and are read separately. What the reduction identifies is where the
current assessment has least support; it is a finding to take through the
release procedure, not a prediction of when, or whether, either criterion
will cease to hold. Establishing that this is a real change rather than
sampling variation calls for a declared analysis with its own uncertainty
treatment, since the two windows
differ in exposure and may not be independent. The margins and affected cells
return to Model and the release authority to guide targeted evaluation,
revision, or operating restrictions.

\emph{Deployment as a monitored safety case.} Commissioning should freeze and identify the complete deployable unit. This unit includes the model checkpoint, preprocessing and post-processing code, runtime container, robot and sensor firmware, calibration, safety controller, and operational design domain. The team should then establish baseline measurements on the same scenario matrix used at the data and model stages. Runtime monitoring must combine hard physical invariants with leading indicators of degradation. One compact alarm score over a rolling window $W$ is
\begin{equation}
A_t = \max\!\left\{
\frac{[S_p(t)-d(t)]_+}{S_p(t)},\;
\frac{[Q_{0.99}(T_{\mathrm{loop}};W)-T_{\mathrm{deadline}}]_+}{T_{\mathrm{deadline}}},\;
\frac{[\mathrm{UCB}_{1-\alpha}(r_{\mathrm{int}};W)-\tau_{\mathrm{int}}]_+}{\tau_{\mathrm{int}}},\;
\frac{[e_{\mathrm{cal}}(t)-\tau_{\mathrm{cal}}]_+}{\tau_{\mathrm{cal}}}
\right\},
\label{eq:deployment_alarm}
\end{equation}
where $[x]_+=\max(x,0)$, $d(t)$ is the measured separation between the human and robot, $S_p(t)$ is the protective distance from Eq.~\ref{eq:ssm}, $r_{\mathrm{int}}$ is the rate of human interventions or emergency stops, and $e_{\mathrm{cal}}$ is calibration error. The denominators and the confidence level for the intervention-rate bound must be positive and fixed during commissioning. The window $W$ must also contain enough independent operational units to support the chosen quantile and confidence bound. The score $A_t$ is a dimensionless alarm index, not an estimate of accident probability or proof of safety. A value of zero only means that these four monitored signals have not crossed their thresholds; unmonitored hazards may still be present. A separation violation should immediately invoke the verified safety response or stop the system. A timing violation should prevent the learned controller from issuing further actuation commands unless an independent safety channel continues to enforce the physical invariant. Rising intervention or calibration signals can first trigger a degraded mode, such as reduced speed, a restricted workspace, or mandatory human confirmation~\citep{Chen2021SimplexDrive,khalastchi2015,continuousAssurance2025}.
\end{samepage}

Every term of Eq.~\ref{eq:deployment_alarm} is a positive part, so it stays at zero
until its threshold has already been crossed. The score records an
exceedance; the advance warning has to come from elsewhere, and the
protective control that acts before the physical limit is reached is not
replaced by it. Where the parameter envelopes above are in use, the
separation term is better evaluated against the conservative distance
$S_p^{\mathrm{cons}}(t)$ of Eq.~\ref{eq:dep_separation}: measured against the design value, the
term stays quiet when the clearance has been eaten by degradation rather than
by proximity. In the underlying record, human takeovers, fallback
activations and emergency stops need to remain distinguishable from one
another.

For a hypothetical deployment example, suppose a camera firmware update on a mobile manipulator leaves task success at $96\%$ but raises the 99th-percentile loop latency from $72\,\mathrm{ms}$ to $128\,\mathrm{ms}$ against a $100\,\mathrm{ms}$ deadline. The timing term in Eq.~\ref{eq:deployment_alarm} becomes $0.28$, so the system should not wait for a collision or a fall in average success. It should restrict motion or enter the safety controller, freeze the rollout, capture the sensor and timing logs, and restore the last validated bundle of firmware, model, and calibration settings. Model and software updates should follow four steps. First, run a shadow evaluation without actuation. Second, make a canary release to a small fleet or restricted operating domain. Third, compare the update with the incumbent on success, unsafe actions, latency, interventions, and energy use in every scenario cell. Fourth, promote or automatically roll back the update according to predeclared thresholds. Every transition should leave a tamper-evident, versioned record linking the event to the deployed components and the evidence that authorized the change~\citep{assuranceLEAS2023,continuousAssurance2025,roboLineage2026}. Incidents should be added to the scenario matrix and routed back to the Data and Model stages instead of being handled as undocumented maintenance patches.

\emph{Formal Verification}. This step is suggested and provides a mathematical approach to assessing whether a Physical AI system satisfies predefined safety properties and operational requirements\cite{Luckcuck2019}. In robotics and safety-critical autonomous systems, model checking systematically explores modeled system states and transitions to identify violations of safety constraints, timing requirements, and expected behaviors before deployment. These techniques can be applied to verify robotic behaviors, motion planning, collision avoidance, and controller logic. However, verification results depend on the accuracy and completeness of the underlying models and specifications, while state-space explosion can limit computational scalability. Therefore, formal verification complement rather than replace simulation-based testing and runtime monitoring, combining design-time verification with operational monitoring to strengthen end-to-end assurance \cite{Araujo2023}.

\subsection{Operationalizing Governance Elements into Each Life Cycle Component}
\label{sec:operationalizing}

Operationalization means converting a governance value into a testable control at a named life cycle decision. Prior life cycle-oriented trustworthy-AI work, the NIST AI Risk Management Framework, and ISO/IEC 42001 provide process-level scaffolding for governing, mapping, measuring, and managing risk, but they do not by themselves specify the physical evidence needed for a particular robot, operating environment, or human--robot interaction~\citep{li2023trustworthy,NISTAIRMF2023,ISO42001}. We therefore represent the operationalization of each governance element $g$ by a control contract
\begin{equation}
\Gamma_g = \big(\mathrm{owner}_g,\,\mathrm{artifact}_g,\,m_g,\,\tau_g,\,\mathrm{evidence}_g,\,\mathrm{trigger}_g\big),
\label{eq:governance_contract}
\end{equation}
where the owner has decision authority, the artifact is a versioned deliverable, $m_g$ is a measurable quantity, $\tau_g$ is a predeclared acceptance threshold, the evidence records the test result, and the trigger specifies when the evidence expires. Typical triggers for Physical Intelligence include a change to the model, middleware, sensor, firmware, calibration, end effector, human-oversight procedure, or operational design domain (ODD), as well as an incident or statistically significant drift in field performance. At every life cycle handoff, the owner should record a go, conditional-go, or no-go decision and link it to the exact system configuration and evidence. This makes governance continuous without reducing it to continuous paperwork: review is repeated when risk-relevant facts change, not on an arbitrary schedule alone~\citep{NISTAIRMF2023,continuousAssurance2025}.

Table~\ref{tab:gov_framework} organizes the resulting elements into three tiers: Fundamental factors that apply across all stages; Knowledge Generation (Research and Design); and Building, Deployment, and Operation (Data, Model, and Deployment). The subsections below instantiate Eq.~\ref{eq:governance_contract} with artifacts, metrics, benchmarks, and revalidation triggers. Benchmarks are treated as diagnostic evidence rather than universal certificates: results from a simulator, dataset, or reference robot must still be revalidated for the target embodiment and ODD.

\subsubsection{Fundamental Trustworthiness Factors}
The four fundamental factors apply at every life cycle stage and cannot be delegated to a final model audit. We consider them as \emph{fundamental} because the failure of any one can invalidate evidence or responsibility at every subsequent handoff. Moreover, the remaining elements in Table~\ref{tab:gov_framework} have a primary life cycle owner. In summary, the fundamental factors aim to answer four review questions (RQs):
\begin{enumerate}
    \item [RQ1:] Does the evidence remain valid even under realistic variations (Reliability)? 
    \item [RQ2:] Can the identified hazards be prevented, detected, and brought to a safe state (Safety Assurance)?
    \item [RQ3:] Do system requirements and evaluations reflect the people who use, supervise, and are affected by the system (Human Value Alignment)?
    \item [RQ4:] Can a safety-relevant action be reconstructed and assigned to a responsible decision-maker (Transparency and Accountability)?
\end{enumerate}
 Following Eq.~\ref{eq:governance_contract}, each factor should be operationalized through a versioned artifact, a pre-declared acceptance criterion, recorded evidence, and an explicit revalidation trigger.

\definecolor{purpleCat}{RGB}{90,60,150}
\definecolor{blueCat}{RGB}{30,70,170}
\definecolor{greenCat}{RGB}{40,110,60}
\newcommand{\catblock}[2]{{\bfseries\textcolor{#1}{#2}}}
\begin{table*}[t]
\caption{Governance elements organized by category, sub-category, and references across the Physical Intelligence life cycle.}
\label{tab:gov_framework}
\centering
\renewcommand{\arraystretch}{1.5}
\resizebox{\textwidth}{!}{%
\begin{tabular}{@{} >{\raggedright}p{3.0cm} >{\raggedright\arraybackslash}p{2.0cm} >{\raggedright}p{6.5cm} >{\raggedright\arraybackslash}p{6.5cm} @{}}
\toprule
\multicolumn{2}{c}{\textbf{Category}} & \textbf{Governance Element} & \textbf{References} \\
\midrule
\multicolumn{2}{c}{\multirow{4}{*}{\catblock{purpleCat}{Fundamental}}}
 & \textbf{Reliability} & \citep{NISTAIRMF2023, robotarena2025, s2rbench2025} \\
\multicolumn{2}{c}{} & \textbf{Safety Assurance} & \citep{ISO12100, ISO10218, ISO15066, Kojima2025PhysicalRiskControl} \\
\multicolumn{2}{c}{} & \textbf{Human Value Alignment} & \citep{ISO9241-2102019, coDesigningAccessibleRobots2024, Abbo2026HRIValues} \\
\multicolumn{2}{c}{} & \textbf{Transparency and Accountability} & \citep{ieee7001, NISTAIRMF2023, euaiact2024, unece2021r157} \\
\midrule
\multirow{4}{*}{\catblock{blueCat}{\parbox{2.6cm}{\centering Knowledge\\Generation}}}
 & \multirow{2}{2.0cm}{\raggedright\textcolor{blueCat}{\textbf{Research}}}
 & \textbf{Open Science} & \citep{Bonsignorio2025ReproducibleRobotics, cervera2023runtothesource} \\
& & \textbf{Reproducibility} & \citep{mirrer2024} \\
\arrayrulecolor{gray!45}\cmidrule(l){2-4}\arrayrulecolor{black}
& \multirow{2}{2.0cm}{\raggedright\textcolor{blueCat}{\textbf{Design}}}
 & \textbf{Inclusiveness} & \citep{valueElicitationSAR2025, whichValuesSAR2025} \\
& & \textbf{Generalizability} & \citep{mirrer2024, openXEmbodiment2024} \\
\midrule
\multirow{7}{*}{\catblock{greenCat}{\parbox{2.3cm}{\centering Building,\\Deployment,\\and Operation}}}
 & \multirow{2}{2.0cm}{\raggedright\textcolor{greenCat}{\textbf{Data}}}
 & \textbf{Privacy and Transparency} & \citep{gdprRobotsHRI2026, llmPrivacyHRI2025} \\
& & \textbf{Traceability} & \citep{roboLineage2026, ieee7001} \\
\arrayrulecolor{gray!45}\cmidrule(l){2-4}\arrayrulecolor{black}
& \multirow{2}{2.0cm}{\raggedright\textcolor{greenCat}{\textbf{Model}}}
 & \textbf{Explainability} & \citep{darpaRobotExplains2019, codeAsMonitor2025} \\
& & \textbf{Bias and Fairness} & \citep{robotsNotImmune2020, londono2024RobotBias} \\
\arrayrulecolor{gray!45}\cmidrule(l){2-4}\arrayrulecolor{black}
& \multirow{3}{2.0cm}{\raggedright\textcolor{greenCat}{\textbf{Deployment}}}
 & \textbf{Resilience} & \citep{assuranceLEAS2023, doremi2023} \\
& & \textbf{Sustainability} & \citep{sustainabilityManifesto2026, climateRoboticsRoadmap2025} \\
& & \textbf{Conformance} & \citep{cc_assurance_continuity, unece_r156, eu376_2014} \\
\bottomrule
\end{tabular}}
\end{table*}

\textbf{Reliability} asks whether the \emph{complete configured system} continues to perform its required function for a stated duration and under the conditions in its ODD; it is not equivalent to a high average benchmark score. Modeling realistic variations in embodiment makes this distinction operationally important because lighting, friction, payload, calibration, sensor contamination, network delay, and actuator wear can change the same policy's behavior without changing its weights~\citep{staticFrictionSim2Real2025,re3sim2025}. Recent research has shown that minor changes to the environment can lead to rapid fail of existing policies~
\citep{kumar2021rma, pathak2017learning}. RobotArena$\infty$ supports controlled perturbation of manipulation scenes, while S2R-Bench measures autonomous-driving perception under real sensor anomalies, weather, lighting, and road variation~\citep{robotarena2025,s2rbench2025}. Open X-Embodiment captures reliability through cross-task, cross-embodiment, and cross-environment variation, enabling assessment of whether robotic policies remain effective beyond a single training distribution~\cite{openXEmbodiment2024}. These resources are valuable stress-testing templates, but their benchmark-specific perturbations and limited coverage cannot establish reliability for a new robot or ODD, where unmodeled hardware issues, sensing, and environment interactions can produce entirely different failure modes.

Accordingly, we propose a practical reliability workflow with three non-negotiable steps. First, define what must be reliable by constructing a versioned scenario matrix spanning task, environment, embodiment, human presence, and degradation conditions, and identifying safety-critical cells before any testing begins. Second, measure reliability on the actual system by running independently initialized trials in both simulation and target hardware, and reporting run-level failures, Wilson intervals (Eq.~\ref{eq:wilson}), the stage-transfer gap (Eq.~\ref{eq:dep_stagegap}), and tail rather than mean control-loop latency. Third, make deployment conditional on the weakest critical case by applying Eq.~\ref{eq:model_gate} to gate the system on its worst safety-critical cell. Strong nominal performance must never offset failure in a critical condition. 

Writing $p_{\mathrm{succ}}(s\mid v)$ for the per-episode success probability in scenario cell $s\in\mathcal{S}$ under deployed configuration $v$, i.e., the probability that the configuration $v$ (model, software, firmware, calibration, and hardware; the bundle later logged as $v_t$ in Eq.~\ref{eq:physical_event_record}) successfully completes the task in a single episode, the aggregate score and reliability gate are:
\begin{equation}
\bar{p}(v) \;=\; \sum_{s\in\mathcal{S}} \omega_s\, p_{\mathrm{succ}}(s \mid v),
\qquad
G_{\mathrm{rel}}(v) \;=\; \mathbb{I}\!\left[\,\min_{s\in\mathcal{S}_{\mathrm{crit}}} \mathrm{LCB}_{1-\alpha}\!\left[p_{\mathrm{succ}}(s \mid v)\right] \;\geq\; \tau_{\mathrm{succ}}\right],
\label{eq:reliability_gate}
\end{equation}
where $\omega_s$ is the fraction of episodes that fall in cell $s$ (exposure, not the risk weight $w_s$ of Eq.~\ref{eq:data_coverage}) and the right-hand condition is the first condition of Eq.~\ref{eq:model_gate}. The score is a weighted sum, so a rarely visited cell can fail badly without moving it; the gate is a minimum, so no cell can be compensated. 

For example, a warehouse policy with $96\%$ aggregate success but $78\%$ success under simultaneous low light and partial shelf occlusion
(a cell with $\omega_s = 0.05$ and $97\%$ success elsewhere gives exactly this $\bar{p}$)
should be held at the current stage if the predeclared threshold $\tau_{\mathrm{succ}}$ on that cell's lower confidence bound is $90\%$. The team can then collect targeted data, revise sensing or illumination, or restrict the ODD. A passed result remains valid only for the recorded model, firmware, calibration, end effector, and ODD; changing any of these triggers revalidation~\citep{NISTAIRMF2023,re3sim2025,staticFrictionSim2Real2025}.

\textbf{Safety Assurance}
starts from hazards and system boundaries, not from a generic safety benchmark. A complete safety-assurance process should specify the ODD and reasonably foreseeable misuse, identify hazards, estimate and evaluate risk, apply the risk-reduction hierarchy, and record the residual risk and its accepting authority. ISO 12100 supplies this machinery-level workflow; application-specific standards then constrain the implementation, such as ISO 10218 for industrial robots, ISO/TS 15066 for collaborative operation, and ISO 21448 for hazards caused by performance limitations of an intended function in road vehicles~\citep{ISO12100,ISO10218,ISO15066,ISO21448}. This ordering prevents a perception score or language-model refusal rate from being mistaken for evidence that a complete Physical Intelligence system is safe.

Each intolerable hazard should become a row in a living hazard log containing its causal scenario, affected person or asset, severity and exposure assumptions, preventive control, runtime detector, fallback or safe state, verification test, evidence link, owner, and revalidation trigger. The argument connecting these claims to evidence can be organized as a Goal Structuring Notation safety case. Learned policies should be separated from independently enforced constraints wherever practicable: the safety filter in Eq.~\ref{eq:safetyfilter} can reject an unsafe proposal, and a simplex architecture can transfer control to a verified fallback before the recoverable set is left~\citep{Bak2021LearningEnabledCPS,Chen2021SimplexDrive}. To keep this safety argument hazard-indexed rather than model-score-indexed, the extent of that separation can be summarized over the hazard log itself. Writing $\mathcal{H}_{\mathrm{crit}} \subseteq \mathcal{H}$ for the hazards judged intolerable in the risk assessment,
\begin{equation}
\mathrm{Cov}_{\mathrm{enf}} \;=\; \frac{1}{|\mathcal{H}_{\mathrm{crit}}|}
\sum_{h \in \mathcal{H}_{\mathrm{crit}}}
\mathbb{I}\big[\,h\ \text{has a policy-independent safety barrier}\,\big],
\label{eq:enfcoverage}
\end{equation}
where a policy-independent barrier is one whose actuation does not rely on the nominal learned policy, such as a safety-rated stop, a mechanical limit or interlock, an independently enforced speed or force restriction, or an independently evaluated safety filter. This quantity is not a residual-risk metric or a certification threshold, and a barrier counted here is only as dependable as the sensing coverage, timing, and integrity of the channel that carries it. Equation~\ref{eq:enfcoverage} makes explicit which critical hazards still rely on the learned policy as part of their safety argument. For these hazards, a change in $\pi_t$ can invalidate part of the evidence supporting the current safety claim and may therefore require renewed evidence, whereas hazards protected by independent enforcement are primarily revalidated when the enforcing channel, its sensing coverage, timing assumptions, or the ODD changes~\citep{ISO12100,IEC61508-1-2010,Bak2021LearningEnabledCPS}. Conformal monitors can provide calibrated warning-rate guarantees under their stated exchangeability assumptions, but such guarantees apply to the monitor's score and do not by themselves prove system safety under distribution shift~\citep{Luo2024ConformalSafety}.

Benchmarks should be chosen by hazard. LIBERO-Safety separates task success from physical and semantic violations in VLA manipulation, while OopsieVerse measures mechanical, thermal, and fluid damage during household manipulation~\citep{cui2026liberosafety,balaji2026oopsieverse}. They are valuable for finding failure modes, but simulation results must be followed by target-hardware tests of stopping time, separation, contact force or pressure, collision energy, and fallback activation as applicable. For example, the hazard ``unexpected human entry during a manipulator handover'' should be linked to a human-detection requirement, the protective-distance calculation in Eq.~\ref{eq:ssm}, a safety-rated stop, measured worst-case reaction and stopping times, and a named owner. A camera, firmware, layout, or policy update could invalidate the linked evidence and reopen that hazard. The release decision is therefore based on a traceable safety case for the complete configuration, not on an assertion that the learned model is safe~\citep{Kojima2025PhysicalRiskControl,Haddadin2016PHRI}.

\textbf{Human Value Alignment}
Physical Intelligence turns values such as autonomy, accessibility, privacy, dignity, appropriate trust, and meaningful human control into properties of a shared physical interaction\cite{Abbo2026HRIValues}. 
Operationalization of human values into Physical Intelligence therefore must consider all people affected by a physical system: primary users, operators, maintainers, caregivers, bystanders, and people exposed without choosing to interact \cite{nanavati2023design}. Value Sensitive Design, participatory design, and human-centered design processes provide methods for eliciting these auxiliary users' interests and resolving conflicts early \cite{ISO9241-2102019}. Co-design studies with people with mobility disabilities show why accessibility requirements cannot be inferred solely by developers~\citep{khosravy2024human,coDesigningAccessibleRobots2024}. The output of co-design process should be a versioned value-to-requirement matrix in which each material value has a stakeholder, system requirement, foreseeable failure, acceptance test, responsible owner, and method for contesting or revising the requirement. Tools such as the HRI Value Compass can support elicitation, but the resulting values become governable only after they are linked to testable requirements~\citep{Abbo2026HRIValues}.

\begin{table}[t]
\centering
\caption{Behavioral Measures.}
\label{tab:behavioral_measures}
\small
\renewcommand{\arraystretch}{1.4}
\setlength{\tabcolsep}{5pt}
\begin{tabularx}{\textwidth}{
>{\raggedright\arraybackslash}p{3.2cm}
>{\raggedright\arraybackslash}X
>{\raggedright\arraybackslash}p{2.8cm}
}
\toprule
\textbf{Behavioral Measure} & \textbf{Definition} & \textbf{Reference} \\
\midrule

Missed or unwanted interventions
&
An intervention is triggered when the autonomy's inferred intended action diverges from the user's measured command under sufficiently low predictive uncertainty: $\phi_{\mathrm{inferred}}^t \neq \phi_m^t$ and $H(p(a^t \mid \phi_m^t)) < \epsilon$, where $p(a^t \mid \phi_m^t) = \eta\, p(a^t) \sum_{\phi_i^t \in \Phi} p(\phi_m^t \mid \phi_i^t)\, p(\phi_i^t \mid a^t)$. An intervention is unwanted when it is triggered despite the measured command reflecting the user's true intent.
&
\citep{gopinath2021customized}
\\
\midrule

Legibility (fraction of tasks in which intent was inferable and correctable)
&
The time-weighted probability that an observer correctly infers the robot's intended goal $G^*$ from a partial trajectory: $\mathrm{legibility}(\xi) = \dfrac{\int P(G^* \mid \xi_{S \to \xi(t)})\, f(t)\, dt}{\int f(t)\, dt}$, where $f(t)$ weights earlier segments of the trajectory more heavily.
&
\citep{dragan2013legibility}
\\
\midrule

Mode-confusion errors
&
An error that occurs when a user's belief regarding the currently active control mode diverges from the robot's true active mode, resulting in a command appropriate to an unintended mode.
&
\citep{gopinath2017mode}
\\
\midrule

Task and recovery success by user group
&
The proportion of task attempts completed successfully, and the corresponding proportion of failure trials from which the system or user recovers, computed separately per user subgroup (e.g., by impairment level).
&
\citep{park2020active}
\\
\midrule

Override success
&
The proportion of user override or correction attempts that successfully redirect robot behavior toward the user's intended goal.
&
\citep{javdani2018shared}
\\
\midrule

Time to reach a safe state after an override
&
The elapsed time between an override or stop command and the moment at which the robot reaches a verified safe state (e.g., zero velocity or a retracted, safe configuration).
&
\citep{ISO15066}
\\
\bottomrule
\end{tabularx}
\end{table}

Evaluation should combine behavioral and self-reported evidence. Behavioral measures can include task and recovery success by user group \citep{park2020active}, missed or unwanted interventions \citep{gopinath2021customized}, override success \citep{javdani2018shared}, time to reach a safe state after an override \citep{ISO15066}, mode-confusion errors \citep{gopinath2017mode}, and the fraction of tasks for which a person could understand and change the robot's intended action \citep{dragan2013legibility}. Subjective instruments should be selected for the construct rather than treating ``trust'' as a universal score: NASA-TLX measures perceived workload, while the Trust Perception Scale-HRI is designed for perceived trust in human--robot interaction~\citep{Hart1988TLX,Schaefer2016TrustHRI}. Results should be reported with uncertainty and disaggregated across relevant abilities, experience levels, and interaction contexts. High trust is not itself the target; calibrated reliance is, so a participant who refuses an unsafe robot recommendation can represent a successful outcome.

For example, an assisted-living service robot should be reviewed with residents, caregivers, visiting family, maintenance staff, and bystanders \cite{stegner2022designing}. A value such as autonomy can be translated into notice before recording, reversible task choices, an accessible stop or decline mechanism, and escalation to a caregiver only under predeclared conditions \cite{vandepoel2013translating}. Before a pilot, the team sets thresholds for stop-command detection, time to safe state, task completion across mobility groups, and operator workload; records both failures and workarounds; and names the person who can accept a residual value conflict\cite{iso13482}. A change to the user population, autonomy mode, interface, or escalation policy triggers a new Human-Robot Interaction (HRI) evaluation \cite{porter2022principles}. An HRI evaluation refers to a structured and user-centered evaluation of how a change affects the interaction between humans and the robot, including relevant dimensions such as usability, social acceptance, user experience, workload, trust, and reliance \cite{Weiss2011USUS}. It employs quantitative methods, including task-performance metrics, behavioral measures, standardized scales, and statistical analysis; qualitative methods, including interviews, observations, and open-ended responses; or mixed-methods approaches that integrate quantitative and qualitative evidence to assess complementary dimensions of human–robot interaction \cite{Tian2025ExperimentalHRI}. This converts ``human-centered'' from a design aspiration into evidence that can block or condition a release~\citep{Martinetti2021RedefiningSafety,NISTAIRMF2023}.

\textbf{Transparency and Accountability}
solve different but coupled problems. Transparency determines what information each stakeholder can obtain before, during, and after operation; accountability determines who has authority and responsibility to act on that information. IEEE 7001 accordingly treats transparency as stakeholder-specific and measurable, while the NIST AI RMF requires documented roles, escalation paths, go/no-go decisions, and executive responsibility for risk decisions~\citep{ieee7001,NISTAIRMF2023}. For Physical Intelligence, a model card alone is insufficient because a physical action depends on the sensor observation, estimated world state, nominal policy output, safety-layer modification, control mode, and deployed hardware and software configuration.

Transparency should be stakeholder-specific rather than uniforms~\citep{winfield2021ieee}. Operators may 
require real-time system state and safety-mode information, auditors require 
reconstructible logs and configuration provenance, while affected users should 
receive information appropriate to their interaction context. Access to these 
records should therefore be role-based and constrained by privacy, security, 
and hazard relevance~\citep{Callander2024PrivacyConscious}.

For every safety-relevant event at time $t$, a minimal event record can be represented as
\begin{equation}
\ell_t = \big(t,\,o_t,\,\hat{x}_t,\,u_t^{\pi},\,u_t^{\mathrm{safe}},\,m_t,\,v_t\big),
\label{eq:physical_event_record}
\end{equation}
where $o_t$ identifies the sensor observation or an integrity-protected reference to it, $\hat{x}_t$ is the estimated state, $u_t^{\pi}$ is the learned policy's proposed action, $u_t^{\mathrm{safe}}$ is the action after safety filtering, $m_t$ is the autonomy or fallback mode, and $v_t$ identifies the deployed model, software, firmware, calibration, and hardware bundle. Records should use synchronized clocks, integrity protection, access controls, and retention periods appropriate to the hazard and privacy risk. The EU AI Act requires automatic event logging for covered high-risk AI systems, and UN Regulation No.~157 provides a domain-specific example by requiring timestamped records that clarify control transitions, system activation, and overrides in automated lane-keeping systems~\citep{euaiact2024,unece2021r157}. Data sovereignty and minimization remain important, but belong in the privacy control: transparency does not require retaining every raw sensor stream indefinitely.

We propose \emph{trace completeness} as a simple audit gate. If $\mathcal{E}_{\mathrm{crit}}$ is the set of sampled safety-critical events and $\mathcal{F}_e$ the fields required to reconstruct event $e$, then
\begin{equation}
C_{\mathrm{trace}} = \frac{\sum_{e\in\mathcal{E}_{\mathrm{crit}}}\sum_{f\in\mathcal{F}_e}\mathbb{I}[f\ \text{is present and integrity-verified}]}{\sum_{e\in\mathcal{E}_{\mathrm{crit}}}|\mathcal{F}_e|}.
\label{eq:trace_completeness}
\end{equation}
Before release, teams should require $C_{\mathrm{trace}}=1$ for the sampled critical events and complete an incident-replay drill within a predeclared time budget. This metric measures record completeness, not explanation quality or causal certainty.

\textbf{Accountability} attaches decision authority and corrective responsibility to the evidence provided by transparency. A responsibility matrix should span the model provider, hardware manufacturer, system integrator, deployer, and site operator. Every safety-relevant life-cycle gate and incident class should have at least one responsible party for execution and exactly one accountable owner with final decision authority. Let $\mathcal{D}_{\mathrm{crit}}$ denote the set of safety-critical decision points and incident classes, $N_R(d)$ the number of assigned responsible parties, and $N_A(d)$ the number of accountable decision owners for $d$. Responsibility assignment completeness can then be defined as

\begin{equation}
C_{\mathrm{acct}}
=
\frac{1}{\left|\mathcal{D}_{\mathrm{crit}}\right|}
\sum_{d \in \mathcal{D}_{\mathrm{crit}}}
\mathbb{I}
\left[
N_R(d) \geq 1
\land
N_A(d) = 1
\right].
\label{eq:acctcoverage}
\end{equation}

A safety-relevant life-cycle gate should require $C_{\mathrm{acct}} = 1$. This metric does not measure whether the assigned party will make a correct decision; rather, it measures whether authority and responsibility are unambiguously allocated before a safety-relevant event occurs.

For example, after an unexpected manipulator motion, the team should be able to replay the chain from observation through safety-filtered command to actuator, identify whether the failure arose in calibration, inference, integration, or operating procedure, and route the corrective action to its owner. A missing record, ambiguous owner, or failed replay is a no-go result, even if the robot's average task score remains unchanged~\citep{roboLineage2026,faccrt2026roboticAudit}.

\subsubsection{Research Trustworthiness Factors}

Research governance establishes the scientific foundation upon which trustworthy Physical Intelligence systems are built. Unreliable or irreproducible research can spread into downstream engineering decisions and eventually affect real-world safety because Physical Intelligence systems operate in highly dynamic and safety-critical environments.

\textbf{Open Science} covers the practice of making research artifacts, including code, datasets, hardware designs, and experimental protocols---openly accessible to enable independent verification, reuse, and extension by other researchers \citep{pineau2021improving}. In traditional AI governance, open science is largely confined to the release of source code, trained model weights, and digital datasets, since the artifacts under study are themselves purely digital and can be shared at near-zero marginal cost. Physical Intelligence governance extends open science to the additional burden of hardware transparency: reproducing a result in embodied AI often requires not only open code but open mechanical designs, bills of materials, sensor specifications, and assembly documentation \citep{ma2017yaleopenhand,pearce2012openhardware}, since the physical platform itself is frequently a confound that closed-source or proprietary hardware makes impossible for other labs to control for \citep{mirrer2024}. Open hardware in robotics therefore carries a documentation burden with no digital analogue, and community efforts to standardize how mechanical designs, part sourcing, and assembly procedures are published are what make a released design usable by a second lab rather than merely visible to it \citep{bonvoisin2020standardisation,cervera2023runtothesource}. The goal of this governance factor is to lower the barrier to independent replication and cross-lab comparison in Physical Intelligence research, extending open science norms from the digital artifact to the physical system that produced it.

Openness in Physical Intelligence, however, is not simply the digital release debate with hardware attached. Traditional discussions of staged release and structured access concern the diffusion of \emph{informational} capability, and the harms they anticipate are mediated by a human who acts on the model's output. What an embodied release confers is closer to operational capability: a mechanical design, a control stack, and a policy checkpoint together constitute a system that acts on the world directly, and the robotics software ecosystem in which such releases circulate has documented exploitable weaknesses of its own \citep{giaretta2024cybersecurity,neupane2024security,ma2026breaks}. The concern is not hypothetical, since language-conditioned robot policies have been shown to be steerable into unsafe physical actions that their language backbone would refuse to describe \citep{roboPAIR2025}. The resolution is not to release less but to be precise about what each artifact is for. Artifacts that make a claim checkable---mechanical designs, calibration and evaluation protocols, task definitions, benchmark scripts---are largely distinct from artifacts that make a system deployable, such as tuned safety envelopes, site-specific configurations, and operational fleet data; publishing the former is what open science requires, while the latter can be made available to independent auditors under structured access \citep{falco2021governingAISafety} without open publication. Framing the question this way lets a research community satisfy replication and proliferation concerns simultaneously, rather than treating them as a trade-off in which one norm must give way.

\textbf{Reproducibility}, closely related to reliability, ensures that independent researchers can replicate published findings using the same methodology. The robotics community has documented its own version of the reproducibility crisis: closed-source, simplified, or specially tailored hardware and software in laboratory demonstrators renders reproduction of most published robotics experiments difficult, a concern serious enough that a unified conceptual framework (MIRRER) has been proposed specifically to formalize and jointly evaluate reproducibility, replicability, and generalizability in robotics research \citep{mirrer2024}. The problem is not confined to any one subfield---reviews of surgical robotics have likewise found that reproducibility is not yet mainstream practice, with patenting restrictions, safety constraints, and closed hardware compounding the difficulty of replication \citep{surgicalRoboticsRepro2023}, while IEEE Robotics \& Automation Magazine's Reproducible Articles initiative was created specifically to encourage detailed, replicable experimental protocols in intelligent robotics \citep{reproducibleRoboticsArticles2017}. For Physical Intelligence, reproducibility extends well beyond releasing source code. A minimal disclosure set for an embodied result comprises: (i) the hardware bill of materials, including robot model, end effector, and sensor part numbers and firmware versions; (ii) the calibration procedure and its residual error; (iii) the controller type and control frequency; (iv) the simulator version and its physics parameters; (v) datasets, preprocessing pipelines, hyperparameters, random seeds, and released checkpoints; and (vi) the number of trials, the success criterion, and the interval from Eq.~\ref{eq:wilson}. Items (i)--(iii) have no analogue in traditional AI reproducibility checklists, and they are precisely the ones that closed or specially tailored hardware makes impossible to satisfy \citep{ma2017yaleopenhand,mirrer2024}.

Disclosure alone, however, does not say when a replication has succeeded. If the originating laboratory reports a success rate $\hat{p}_{A}$ over $n_A$ trials and an independent group reports $\hat{p}_{B}$ over $n_B$ trials, the reproducibility gap
\begin{equation}
\Delta_{\text{repro}} \;=\; \big| \hat{p}_{A} - \hat{p}_{B} \big|
\label{eq:repro}
\end{equation}
is itself a reportable statistic, and a replication should be judged successful when the Wilson intervals of the two estimates overlap rather than when a demonstration video looks similar. Consider a concrete case: a laboratory publishes a cloth-folding policy fine-tuned from an open-source vision--language--action checkpoint, reporting an $82\%$ success rate on a $7$-DoF Franka arm with a parallel-jaw gripper, and a second group reproduces it at $61\%$ on a UR5e with a different gripper and camera mounting. Without the disclosure set, that $\Delta_{\text{repro}} = 21$ points is uninterpretable: it is consistent with a fragile method, with an $8\,\text{mm}$ hand--eye calibration residual, with a control loop running at $20\,\text{Hz}$ instead of $50\,\text{Hz}$, or with a gripper whose finger compliance differs enough to change the contact dynamics of the fold. With items (i)--(iv) disclosed, each of these can be checked and ruled in or out; without them, the community can only record that the result ``did not reproduce'' and lose the information about why. The second group should therefore report $\Delta_{\text{repro}}$ together with its own trial count and platform deltas rather than a bare success rate, so that a discrepancy is attributable to the method rather than to an undocumented difference in platform. Such practices improve scientific credibility while accelerating cumulative progress across the Physical Intelligence research community.

To achieve reproducibility, a structured literature-search process informed by established systematic review guidelines is essential for enabling independent researchers to trace, assess, and replicate the review process\cite{Page2021}. Inclusion criteria prioritize peer-reviewed journal articles, conference proceedings, and authoritative technical publications that directly address the capabilities, risks, assurance mechanisms, and governance of Physical AI systems. Exclusion criteria include duplicate publications, studies with limited relevance to the review objectives, and publications that do not provide sufficient technical or conceptual contributions \cite{Kitchenham2007}. Standards and regulations are considered as a distinct category of evidence, with their issuing organizations, publication versions, geographical applicability, and normative status examined to distinguish mandatory regulatory requirements from voluntary technical guidance \cite{EuropeanCommission2024}.  The taxonomy and conceptual framework are developed through an iterative synthesis of the selected literature, combining recurring themes and concepts identified through qualitative analysis with established principles of safety, reliability, and governance \cite{ISOIEC17000}.

\subsubsection{Design Trustworthiness Factors}

Governance should be embedded during the system design phase before implementation decisions constrain future development. Early design decisions determine system objectives, operational boundaries, acceptable risks, and stakeholder priorities, thereby influencing downstream data collection, model development, and deployment strategies. 

\textbf{Inclusiveness} seeks to ensure that Physical Intelligence systems reflect the needs, expectations, and values of the diverse individuals and communities that interact with them rather than optimizing exclusively for technical performance. Value-sensitive approaches applied directly to embodied systems illustrate how this can be done in practice: participatory design studies for socially assistive robots have used structured value-elicitation methods with real stakeholder groups to surface conflicting priorities before a system is built, rather than after deployment \citep{valueElicitationSAR2025}, and multi-stakeholder empirical studies of socially assistive robots in elder care have shown that patients, family caregivers, and clinical staff often prioritize different values (autonomy versus safety, companionship versus efficiency), values that only surface through direct, structured elicitation \citep{whichValuesSAR2025}. Operationally, inclusiveness can be incorporated through stakeholder mapping, participatory design workshops, interviews, usability testing, Wizard-of-Oz prototyping to elicit interaction expectations before a policy exists, accessibility review against published interaction guidelines, and iterative human-in-the-loop design reviews involving both technical and non-technical users \citep{coDesigningAccessibleRobots2024,Qbilat2021}. For example, when designing a hospital delivery robot, developers should engage nurses, physicians, patients, maintenance personnel, and hospital administrators during requirement elicitation to identify workflow constraints, accessibility concerns, communication preferences, and safety expectations that may not emerge through purely engineering-driven design.

\textbf{Generalizability} complements inclusiveness by ensuring that system designs remain applicable across diverse deployment contexts instead of being optimized for a single environment. Physical Intelligence systems frequently encounter distribution shifts after deployment due to changing environments, hardware upgrades, or evolving operational requirements. The MIRRER framework treats generalizability as a formally measurable robotics capability---the ability of a robot to perform a task under varied contexts---precisely because the field has historically lacked shared definitions or protocols for evaluating it \citep{mirrer2024}, and empirical real-to-sim evaluations of vision-language-action policies confirm the concern in practice: performance declines markedly once policies are evaluated in environments or object configurations beyond their training distribution, indicating that today's ``generalist'' policies still specialize to their training data \citep{realm2025}. Domain-randomization and domain-adaptation techniques have been proposed specifically to close this gap by exposing manipulation policies to deliberately varied simulated conditions during training \citep{sim2realVLA2026}. Two concrete strands of Physical Intelligence research illustrate what governance-relevant generalizability looks like in practice: cross-embodiment work such as Open X-Embodiment has shown that pooling manipulation data across many distinct robot bodies and training a single policy on the combined corpus improves generalization beyond what any single-embodiment dataset can provide \citep{openXEmbodiment2024}, and follow-on architectures have pursued the same goal by explicitly sharing representations or attention mechanisms across heterogeneous robot morphologies rather than training a separate model per platform \citep{crossformer2024,hpt2024}. A second, complementary strand targets generalization across open-ended language and object categories rather than across hardware: composable value-map and semantic-field approaches allow a manipulation policy to act on objects and instructions never seen during training by grounding pretrained vision-language representations directly in the robot's 3D workspace \citep{voxposer2023,clipFields2023}. Consequently, generalizability should be operationalized at design time through modular architectures, cross-domain validation, and simulation-to-real testing under diverse environmental conditions, rather than through optimization against a single benchmark environment. Because MIRRER treats generalizability as a measurable capability, the quantity worth reporting is a gap rather than a score: the drop in success rate between the design context and each held-out context, which is the construction of Eq.~\ref{eq:dep_stagegap} with the index ranging over deployment contexts rather than life cycle stages. For example, an agricultural harvesting robot designed initially for apple orchards should be evaluated on vineyards and citrus farms, and what a design review needs is the success-rate gap on those held-out contexts, not the apple-orchard score alone. Such evaluations provide confidence that the design remains robust under changing operational contexts and future deployment scenarios.

\subsubsection{Data Trustworthiness Factors}
\label{sec:data_trustworthiness}

\textbf{Privacy and Transparency} Data governance is particularly important for Physical Intelligence because embodied systems continuously collect multimodal information from their surrounding environments through cameras, LiDAR, microphones, inertial sensors, and positioning systems. Unlike traditional AI applications that operate on static datasets, Physical Intelligence systems often collect data throughout their operational lifetime, creating additional privacy and accountability challenges. Robotics-specific studies of household and social robots show these challenges are distinctive to embodiment: because social robots rely on rich multimodal sensing---audio, video, motion, proximity, and affective cues---they can capture information beyond what a user explicitly intends to share, and their personalized behavior can itself leak sensitive information to bystanders who never interacted with the robot directly \citep{gdprRobotsHRI2026}. Empirical work benchmarking privacy-relevant decision-making in household robots found that users' own privacy expectations for what a robot may record, process, or share frequently diverge from what current AI decision-making systems would actually do, underscoring the need for explicit, engineered privacy safeguards rather than reliance on model judgment alone \citep{llmPrivacyHRI2025}. Robot-specific design work has responded with concrete privacy-by-design techniques---on-device processing, real-time user feedback and intervention interfaces, and privacy-sensitive sensor design---built specifically for domestic and social robot deployments \citep{privacySensitiveRobotDesign2025}. Minimization
has two parts, and only the second is a ratio: how much is captured, and how much of it leaves the device in identifiable form. Each needs a ceiling declared before deployment if minimization is to be audited rather than asserted. For each data category $c$, let $b^{\mathrm{capture}}_c$ and $b^{\mathrm{egress}}_c$ be the bytes captured and the bytes exported in identifiable form over a declared accounting window, and let
\begin{equation}
R_c = \frac{b_c^{\mathrm{egress}}}{b_c^{\mathrm{capture}}}.
\label{eq:egress}
\end{equation}
Release is conditioned on two limits: $b^{\mathrm{capture}}_c \le B_c$, which bounds collection against the stated purpose, and $R_c \le \tau_c$, which bounds identifiable disclosure, with $\tau_c = 0$ for categories processed on-device by design. $R_c$ alone would be gameable, since capturing more lowers the ratio without disclosing less; the two conditions are therefore
reported together with the absolute byte counts. Exceeding either is a governance trigger under Eq.~\ref{eq:governance_contract}: the affected category is suspended until the control is restored, or the limit is revised and re-approved. An assisted-living robot performing face recognition locally reports $R_{\mathrm{video}} = 0$ while still exporting derived events; a debugging build that raises $R_{\mathrm{video}}$ above zero is a configuration change that reopens the privacy control.

\textbf{Traceability} complements transparency by ensuring that every dataset, preprocessing step, annotation procedure, and model version can be traced throughout the AI life cycle. This is a distinctly harder problem in Physical Intelligence than in conventional ML pipelines: robot policy iteration generates a messy, embodied data ecosystem in which a rollout is more than a video, a dataset is more than a folder of accepted episodes, and a trained policy is more than a checkpoint---each is an artifact that must be linked across robot configuration, task context, sensor observations, human review, dataset admission, training provenance, and deployment decisions, a governance problem serious enough to motivate purpose-built agent-native lineage-tracking systems for robot learning \citep{roboLineage2026}. IEEE's transparency standard for autonomous and intelligent systems similarly calls for autonomous systems to make their reasoning and data provenance auditable to operators and, where appropriate, to affected users \citep{ieee7001}. Traceability admits the same treatment as the runtime records of Eq.~\ref{eq:trace_completeness}, applied to training data rather than events. For sampled episodes $\mathcal{D}$ with required provenance fields $\mathcal{P}_d$---robot configuration, calibration, operator, task context, admission decision, and consent basis---
\begin{equation}
\mathrm{Cov}_{\mathrm{lineage}} = \frac{\sum_{d\in\mathcal{D}}\sum_{f\in\mathcal{P}_d}\mathbb{I}[f\ \text{resolves}]}{\sum_{d\in\mathcal{D}}|\mathcal{P}_d|},
\label{eq:lineage}
\end{equation}
with release conditioned on $\mathrm{Cov}_{\mathrm{lineage}} = 1$ over the sample and on resolving an incident to its contributing episodes within a declared time budget. A team that cannot name the recordings, annotation version, and preprocessing revision behind a deployed policy after an unexpected navigation failure has a governance failure, independent of whether the policy is later found at fault.

\subsubsection{Model Trustworthiness Factors}

Model governance focuses on ensuring that learned decision-making processes remain understandable, equitable, and aligned with human expectations. 

\textbf{Explainability} is particularly important for Physical Intelligence because it directly influences physical actions that may affect human safety. Work on Explainable Autonomous Robots (XAR) has produced concrete demonstrations of this: a DARPA-funded study had a robot narrate the steps it took while performing a manipulation task (opening a medicine bottle), and found that even simple ``what''-level explanations measurably increased human participants' trust in the robot's behavior, though the researchers cautioned that trust must be appropriately calibrated rather than simply maximized \citep{darpaRobotExplains2019,sanneman2020}. More recent work integrating LLMs into robot explanation pipelines has focused on grounding generated explanations in the robot's actual internal state and sensor logs, rather than generic post-hoc narratives, specifically to preserve the link between what a robot says and what it did \citep{xarRAG2026}. Some work treats explainability as inseparable from failure detection: VLM-based monitors have been developed to watch an executing robot policy, flag the point at which a task-relevant constraint has been violated, and produce a human-readable account of what went wrong and why \citep{codeAsMonitor2025}. The governance value of these mechanisms is clearest where they were absent. The investigation of the 2018 Tempe fatality found that the automated driving system classified the pedestrian in turn as an unknown object, a vehicle, and a bicycle, and that each reclassification discarded the tracking history used to predict the object's path, so no stable motion prediction was ever established. The system's object taxonomy also doesn't contain a class for a pedestrian crossing outside a crosswalk \citep{ntsb2019tempe}. What made the failure reconstructible afterwards was retained perception logs, not any explanation the system produced at the time. This points to two different requirements: a robot must explain itself to the operator in plain language, and it must also leave a complete record for reconstruction later. Natural-language explanation serves the first requirement and logging serves the second. Regulation now mandates the second directly: the EU AI Act requires high-risk systems to log events automatically throughout their lifetime \citep{euaiact2024}, and UNECE Regulation No.\ 157 requires automated lane-keeping systems to carry a data storage system recording specified occurrences such as activation, transition demand, and override \citep{unece2021r157}. The two requirements are governed differently. Logging is a completeness property, already captured by Eq.~\ref{eq:trace_completeness}; explanation is a fidelity property. For a sample $\mathcal{I}$ of interventions,
\begin{equation}
F_{\mathrm{expl}} = \frac{1}{|\mathcal{I}|}\sum_{i\in\mathcal{I}} \mathbb{I}\big[\,\text{stated reason}_i = \text{trigger recorded in } \ell_{t_i}\,\big]
\label{eq:expl_fidelity}
\end{equation}
measures agreement between what the system reported and what the record of Eq.~\ref{eq:physical_event_record} shows. A fluent explanation that disagrees with the record is worse than none, since it misdirects the operator. A collaborative robot halting an assembly task should report whether the trigger was obstacle detection, a force-limit violation, or predicted collision risk, and that report should survive audit against $\ell_t$.

\textbf{Bias and Fairness} remain equally important because Physical Intelligence systems interact directly with diverse human populations and physical environments. Biased training data, unrepresentative simulation environments, or demographic imbalances may produce unequal system performance across user groups, potentially resulting in unsafe or discriminatory behavior. Four sources of such disparity recur in embodied systems.
The first is demographic: commercial facial-recognition systems show measurable demographic differentials \citep{robotsNotImmune2020}, with false positive rates in one government evaluation varying by up to two orders of magnitude across groups \citep{grother2019frvt}. Robotics audits extend this to human-robot interaction, where models estimating engagement and affect lose accuracy for older adults \citep{hriFacialBias2026}. The second is environmental: recognition accuracy for ordinary household objects such as soap and stoves from low-income households drops substantially\citep{deVries2019}, so a domestic service robot can fail due to the material conditions of the home. The third is infrastructural: driving perception is predominantly trained and benchmarked on data from a small number of high-income cities with regularized road networks \citep{geiger2012kitti,caesar2020nuscenes,sun2020waymo}. This results in mixed traffic streams, including electric two-wheelers, three-wheelers, unchannelized intersections, and informal crossing behaviors, being systematically underrepresented. The fourth is morphological: wheelchairs, walkers, crutch gait, strollers, and the smaller stature of children have received comparatively little systematic evaluation in pedestrian detection and trajectory prediction, and errors here concentrate on agents whose motion departs from a walking-adult prior, which is where prediction carries the most safety weight \citep{Nanavati2024}. These four sources define the group set $\mathcal{G}$ over which performance must be disaggregated rather than averaged. Following the treatment of critical scenario cells in Section~\ref{sec:operationalizing}, the release condition binds the weakest group rather than the mean:
\begin{equation}
\min_{g\in\mathcal{G}} \mathrm{LCB}_{1-\alpha}\big(\hat{p}_g\big) \;\geq\; \tau,
\qquad
\Delta_{\mathrm{fair}} = \max_{g\in\mathcal{G}} \big(\hat{p} - \hat{p}_g\big),
\label{eq:fairness_gate}
\end{equation}
where $\mathrm{LCB}$ is the lower Wilson bound of Eq.~\ref{eq:wilson} and $\Delta_{\mathrm{fair}}$ is reported alongside aggregate accuracy. Because the bound tightens with per-group sample size, the gate also forces adequate testing of small groups instead of permitting them to be averaged away. A domestic assistance robot should therefore be evaluated across elderly residents, wheelchair users, children, and homes differing in lighting and clutter, with $\mathcal{G}$ revisited whenever the deployment population changes.


\subsubsection{Deployment Trustworthiness Factors}

Governance responsibilities continue long after initial deployment because Physical Intelligence systems operate in dynamic, unstructured environments where software models, physical hardware, and operational conditions continually drift and change. 

Neither the problem nor the response is new. Asset management and
resilience engineering have long-standing approaches to holding capability
over a service life~\citep{iso55000},~\citep{hollnagel2006resilience}, and AI
management frameworks address continuing monitoring and improvement in the
same spirit~\citep{NISTAIRMF2023},~\citep{ISO42001}. What Physical Intelligence adds is the coupling
between a learned decision, the physical dynamics it acts on, and the time
available to intervene before harm. Behavior here can change through an
update, but equally through a change of input or physical condition, with the
model weights untouched throughout.

The deployment obligations fall to three factors. \emph{Resilience} covers
disturbance detection, safe response and recovery. \emph{Sustainability}
covers environmental, economic and societal impact across the service life.
\emph{Conformance} covers whether the system still satisfies the
configuration, operating-domain and authorization requirements it runs under.
They look at different objects but share the control-contract structure of
Eq.~\ref{eq:governance_contract}, and their feedback obligations are collected at the end of this
subsection.

\textbf{Resilience} refers to the ability of a Physical Intelligence system to anticipate, withstand, recover from, and adapt to unexpected disruptions while maintaining safe operational behavior. A systems-engineering mapping of assurance techniques for learning-enabled autonomous systems identifies resilience engineering as a distinct discipline that accommodates sensor inaccuracies arising from measurement limitations, environmental dust, debris, or adversarial conditions, allowing a system to recognize degraded performance and adapt---for example, by shifting from perception-based to odometry-based navigation \citep{assuranceLEAS2023}. Empirical work on robot swarms reinforces the need for this adaptation to be predictive rather than purely reactive: faults such as dust-related sensor degradation tend to accumulate gradually rather than occur as sudden step changes, meaning anomaly-detection methods that catch early-stage degradation before it escalates into a critical system failure meaningfully improve long-term autonomy \citep{okeeffe2023, khalastchi2015}.

Resilience failures are not merely mechanical or hardware-bound: language-model-driven planners can also silently drift out of alignment with what the physical robot is actually able to execute in real time. 
This has motivated methods that explicitly detect plan-execution misalignment and ground the language model's next decision back in the robot's true physical state \citep{doremi2023}, as well as corrective-planning frameworks that use the detected mismatch to revise the plan online rather than aborting the task outright \citep{copal2024}. These mechanisms support resilience only within their evaluated conditions; changing
localization or replanning does not itself establish that the resulting behavior is safe.

Resilient governance therefore rests on continuous monitoring, fallback controllers, and periodic re-validation rather than on pre-deployment verification alone, and it declares two quantities rather than one. Given the degradation alarm $A_t$ of Eq.~\ref{eq:deployment_alarm}, a deployment fixes an alarm threshold $\tau_A$, a bound on the time to reach a safe state once the alarm fires, and a minimum detection lead time
\begin{equation}
L = t_{\mathrm{threshold}} - t_{\mathrm{detect}} \;\geq\; L_{\min}.
\label{eq:lead_time}
\end{equation}
$L_{\min}$ is what makes the requirement predictive rather than reactive: a monitor firing at $L = 0$ satisfies the alarm while leaving no operational margin. 
A warehouse robot with dust-degraded LiDAR should thus be detected while its localization error is still within tolerance, then transition to alternative localization, reduce speed, and notify operators before the threshold is crossed.

Both equations take their thresholds as given. At commissioning
they are not given, and a deployment cannot wait for a baseline it can only
acquire by operating. What follows sets out where a threshold comes from, the
rule that fixes it, how it is governed once set, and what evidence the
monitor owes about itself.

\emph{Where they come from.} Not every monitored signal gets its threshold
the same way. The separation and timing terms of Eq.~\ref{eq:deployment_alarm} are \emph{derived}:
$S_p$ comes out of Eqs.~\ref{eq:ssm}--\ref{eq:latency} and $T_{\mathrm{deadline}}$ out of the
real-time analysis, so their limits are conclusions of the safety analysis
rather than settings. The intervention rate is \emph{empirical}; no analysis
fixes the rate at which human takeover becomes anomalous, because that
depends on the site, the task mix and the people operating. Calibration drift
sits between the two, since a derived allowable error may exist while the
warning level below it is calibrated from observation. Keeping the
distinction visible is what stops a derived limit being relaxed to quiet a
nuisance alarm. The lead time $L_{\min}$ of Eq.~\ref{eq:lead_time} is derived as well: the
time to reach a safe state, which the design analysis already produces as
$T_s$ in Eq.~\ref{eq:ssm_simple}, plus human response time where the response is not
automatic, plus an allowance for the detector's own latency. The trajectory
$L$ is measured against has to describe how the variable would evolve
\emph{without} the corrective response, since a successful intervention keeps
the crossing from ever being seen; the requirement is therefore checked by
justified prediction or by controlled test.

\emph{The rule that fixes a threshold.} Let $x(t)$ increase toward an
unacceptable limit $x_{\lim}$, let the measurement underestimate $x$ by at
most $\epsilon_x$, let a validated bound $\dot{x} \leq \rho$ hold until the
response completes, and let $H$ bound the sum of sampling, detection,
response, and other timing allowances. For an alarm at measured value $\tau$,
\begin{equation}
\begin{aligned}
  \tau &\leq x_{\lim} - \epsilon_x - \rho H,  x_{\mathrm{nom}}^{U} &< \tau,
\end{aligned}
\label{eq:dep_threshold}
\end{equation}
where $x_{\mathrm{nom}}^{U}$ bounds the nominal measured signal. The first
inequality puts the alarm early enough to act on; the second keeps it quiet
in normal operation. Both can hold only when
$x_{\mathrm{nom}}^{U} < x_{\lim} - \epsilon_x - \rho H$, and where that fails
there is no usable threshold at all. That is a design finding, answered by
reducing $\rho$, shortening $H$, or narrowing the operating domain --- not by
moving $\tau$ until the alarm behaves. Three cautions attach to the rule. A
rate estimated from limited observation is not a worst-case bound. Abrupt
faults need separate treatment. And a variable that falls toward danger needs
the inequality reversed.

\emph{Example.} Ignoring measurement and timing uncertainty for illustration,
let localization error have a limit of $0.50\,\mathrm{m}$, a nominal upper
value of $0.10\,\mathrm{m}$, a growth bound of $0.02\,\mathrm{m}$ per day,
and a seven-day maintenance horizon; the latest allowable threshold is then
$0.36\,\mathrm{m}$. At $0.08\,\mathrm{m}$ per day the same horizon admits no
threshold above the nominal range, while a verified three-day response gives
$0.26\,\mathrm{m}$. More frequent cleaning may reset accumulated degradation
without reducing the rate between cleanings, and immediate protection must
remain adequate while maintenance is pending.

\emph{Governing the thresholds.} A threshold is a governance element in the
sense of Eq.~\ref{eq:governance_contract}, and carries the same fields as any other: an owner, who is
the authority accepting the residual risk; the configuration it applies to; a
versioned artifact; the measured quantity; the acceptance rationale and its
evidence; and a revalidation trigger. At commissioning the level is set
conservatively, either from the nearest validated configuration --- with
evidence that it transfers to this configuration and this site --- or from a
supervised pilot in which the warning runs in shadow mode behind
independently justified protection. Revision then follows a procedure agreed
in advance, so that it does not become a standing negotiation. After that,
relaxing a safety-relevant level takes a decision by the accepting authority,
recorded against the release record; tightening takes an assessment wherever
it changes operational behavior or alarm burden. More evidence sharpens the
estimate without implying that the level should move in either direction. The
asymmetry is deliberate, because relaxation is how monitored systems drift
toward the boundary of safe operation while every indicator continues to read
nominal~\citep{rasmussen1997},~\citep{vaughan1996}.

\emph{Monitor assurance.} The monitor has to be monitored in turn. A safety
performance indicator is a claim, and like any claim it needs evidence~\citep{ul4600}; indicators upon the indicators are what reveal that a signal has
stopped tracking the hazard it was chosen for~\citep{ratiu2024spi}. Evaluation
should combine controlled degradation or fault-injection tests with
operational events reviewed independently, and should report false alarms per
declared exposure, missed detections among established events, warning lead
time and response completion time, each with its uncertainty and its
unresolved cases kept. A standing burden of nuisance alarms is a defect in
the monitor rather than an operational nuisance: an alarm an operator cannot
act on is eventually one the operator stops attending to~\citep{sorkin1988},
and disuse of this kind is a documented failure mode of automation whose
reliability the operator has learned to distrust~\citep{parasuraman1997}. Alarm
management practice in the process industries treats nuisance-alarm rate as a
managed performance quantity for exactly this reason~\citep{iec62682}. An
absence of recorded misses, meanwhile, is not the same as an absence of
missed detections. The same applies to the site conditions
above: coexistence and lighting maintenance data can guide the choice of
tests~\citep{iec62657},~\citep{cie97}, but have to be mapped onto the variable
actually being monitored before either can supply a rate bound for Eq.~\ref{eq:dep_threshold}.
Where observability is inadequate, the rate bound unsupported, or the
response time too short, the finding returns to Design as a limit on what the
assurance can claim.

\textbf{Sustainability} extends deployment governance beyond immediate technical performance by evaluating the environmental, economic, and societal impacts of Physical Intelligence systems across their entire life cycle. A recent manifesto formally proposes Sustainable Robotics as its own research discipline, arguing that sustainable robot design must account for the environmental footprint across the full physical life cycle---material selection, energy consumption, and manufacturing---rather than software efficiency alone, since physical robots bear a hardware and end-of-life burden that purely digital AI systems do not have \citep{sustainabilityManifesto2026}.

This life cycle view has been applied concretely: a case study integrating full life-cycle assessment (LCA) into the design of an urban municipal-waste service robot demonstrated how environmental, social, and economic sustainability criteria can be engineered into system requirements from the earliest design stage rather than retrofitted afterward \citep{urbanServiceRobotLCA2023}. Moreover, a roadmap for climate-relevant robotics research cautions that the cumulative emissions from manufacturing, training, and operating growing fleets of physical robots represent a substantial sustainability cost that must be explicitly weighed against the environmental benefits these systems provide \citep{climateRoboticsRoadmap2025}.

To operationalize sustainability governance, metrics must explicitly separate embodied carbon (e.g., raw material extraction, component manufacturing, supply chain logistics, and battery chemistry) from operational carbon (e.g., compute energy and fleet energy draw). At the computational layer: A growing body of Physical Intelligence research targets the energy cost of inference directly, as a vision-language-action policy running continuously on onboard hardware exhibits a vastly different energy profile than a cloud-hosted model queried intermittently. Dynamic-inference methods skip unnecessary computation on routine control steps to reduce energy consumption and latency \citep{deerVLA2024}, action-aware pruning techniques cut redundant computation from vision-language-action inference without degrading task success \citep{specPruneVLA2025}, and reinforcement-learning-based recovery methods restore performance lost when a policy is aggressively compressed for cheaper onboard deployment \citep{rlrc2025}. At the physical layer, circular-economy practice supplies the corresponding levers: battery state-of-health monitoring, predictive maintenance to extend asset life, modular designs for component-level repair, and structured end-of-life recycling.

Separating the two components is useful only if they are compared on a common functional unit. 
Over a service life yielding $N_{\mathrm{success}}$ completed tasks,
\begin{equation}
e_{\mathrm{task}} = \frac{E_{\mathrm{embodied}} \;+\; \int_{0}^{T_{\mathrm{life}}}\big(P_{\mathrm{compute}}(t) + P_{\mathrm{actuation}}(t)\big)\,\mathrm{CI}(t)\,dt \;+\; E_{\mathrm{EOL}}}{N_{\mathrm{success}}},
\label{eq:carbon_per_task}
\end{equation}
where $\mathrm{CI}(t)$ is grid carbon intensity. Written this way, three levers usually discussed apart become commensurable: a larger policy raises the integral, a longer service life amortizes $E_{\mathrm{embodied}}$, and an unreliable policy raises $e_{\mathrm{task}}$ through the denominator. Reliability is thus a sustainability property, not only a safety one.

A figure of this kind means little without its boundary, so the
boundary should be declared alongside the result: sensing, computation,
actuation, idle operation, charging losses, off-board services, transport and
the allocation of training, each included term counted once and each
exclusion justified. The functional unit needs stating too --- task duration,
load, distance, or whatever specification makes the comparison fair --- and
failed attempts stay in the numerator wherever they fall inside the declared
boundary. Energy use, output, retries and maintenance tend to move together,
so the three levers are measured jointly rather than inferred from model size
or asset age.

Reporting $e_{\mathrm{task}}$ alongside safety and reliability evidence keeps this trade-off visible at each revalidation rather than deferring it to an end-of-program assessment.

Economic assessment uses the same functional unit and boundary as $e_{\textrm{task}}$ but, unlike Eq.~\eqref{eq:carbon_per_task}, which sums emissions without discounting, this carries the time value of money. If we let $\rho$ be the discount rate, $K_0$ he acquisitions and commissioning cost, $K_{\textrm{EOL}}$ the end-of-life cost, and $\dot N_{\textrm{success}}(t)$ the rate of completed tasks, then the levelized cost for each successful task will be 

\begin{equation}
  \kappa_{\text{task}} =
  \frac{K_0 + \int_0^{T_{\text{life}}}\!\big[\dot K_{\text{om}}(t)
        + \dot K_{\text{en}}(t) + \dot K_{\text{rev}}(t)
        + \dot K_{\text{fail}}(t)\big]e^{-\rho t}\,dt
        + K_{\text{EOL}}\,e^{-\rho T_{\text{life}}}}
       {\int_0^{T_{\text{life}}}\dot N_{\text{succ}}(t)\,e^{-\rho t}\,dt}.
  \label{eq:ktask}
\end{equation}

Here, $\dot K_{\textrm{om}}$ is maintenance and repair, $\dot K_{\textrm{en}}(t) =
\big(P_{\textrm{compute}}+P_{\textrm{actuation}}\big)\lambda_{\textrm{el}}(t)$ prices
the same energy draw as Eq.~\eqref{eq:carbon_per_task} at an electricity price defined by $\lambda_{\textrm{el}}(t)$ instead of carbon intensity, $\dot K_{\textrm{rev}}$
is the expected cost rate of monitoring and of change-driven revalidation defined by triggers of the governance contract in Eq.~\eqref{eq:governance_contract}, while $\dot K_{\textrm{fail}} = \kappa_f\,\dot N_{\textrm{fail}}$ sets the price of each failed attempt at $\kappa_f$. The levelized cost construction of energy economics is thus the discounted life-cycle cost divided by the discounted output. Practitioners in the field raise the point that unstated assumptions yield widely varying or even contradictory figures at times ~\cite{branker2011lcoe} -- the same discipline as the boundary rule from before. Life-cycle costing guidance for reliable equipment it similar in treating the costs of dependability as part of the account ~\cite{iec60300_3_3}.

In the case of constant flows, with attempt rate $\nu$, success probability
$p_{\textrm{success}}$, cost rates $k_{\textrm{om}},k_{\textrm{en}},k_{\textrm{rev}}$ and
$\varphi_\rho=(1-e^{-\rho T_{\textrm{life}}})/\rho$, Eq.~\eqref{eq:ktask}
simplifies to:

\begin{equation}
  \kappa_{\textrm{task}} =
  \frac{K_0+K_{\textrm{EOL}}e^{-\rho T_{\textrm{life}}}}{\varphi_\rho\,\nu\,p_{\textrm{success}}}
  + \frac{k_{\textrm{om}}+k_{\textrm{en}}+k_{\textrm{rev}}}{\nu\,p_{\textrm{success}}}
  + \kappa_f\,\frac{1-p_{\textrm{success}}}{p_{\textrm{success}}}.
  \label{eq:ktask-const}
\end{equation}

The first two terms spread fixed cost over successes, while the third term represents the cost of failure per success, vanishing only if failures are free. Since $\kappa_f$ includes the cost of human intervention, the
intervention rate monitored in Eq.~\eqref{eq:deployment_alarm} is indeed an economic quantity, but $k_{\textrm{rev}}$ was the one left unquantified. Safety margins play a role as well: a longer reaction time widens $S_p^{\textrm{cons}}$ in Eq.~\eqref{eq:dep_separation} and, at fixed layout, a wider protective separation may lower $\nu$ or demand more floor area.

As a numerical example, consider $K_0=80{,}000$, $K_{\textrm{EOL}}=2{,}000$,
$k_{\textrm{om}}=8{,}000$, $k_{\textrm{en}}=1{,}500$, $k_{\textrm{rev}}=3{,}000$ per
year, $\kappa_f=2$, $\nu=25{,}000$ per year, $T_{\textrm{life}}=7$ years,
$\rho=0.08$, all in arbitrary currency units. $p_{\textrm{success}}=0.96$ gives
$\kappa_{\textrm{task}}=1.24$ and $p_{\textrm{success}}=0.78$ gives $1.98$, a rise of
about 60\%, well above the 23\% that the $1/p_{\textrm{success}}$ factor alone would
suggest, so the change is significant.

Since $\kappa_{\textrm{task}}$ drops in $p_{\textrm{success}}$ at fixed cost parameters, a conservative result follows from Eq.~\eqref{eq:wilson}, as in the release gates from Eq.~\eqref{eq:fairness_gate}. With
288 successes in 300 independent trials, the one-sided 95\% lower bound is
0.937, raising $\kappa_{\textrm{task}}$ from the previous example from 1.24 to a value of 1.31.
Economic viability would then require
$\kappa^{\textrm{UCB}}_{\textrm{task}} = \kappa_{\textrm{task}}
(\mathrm{LCB}_{1-\alpha}[p_{\textrm{success}}]) \le v_{\textrm{task}}$, where
$v_{\textrm{task}}$ is the value of a completed task. This bound covers only an uncertainty in
$p_{\textrm{success}}$ -- uncertainty in the cost parameters must be accounted for separately.

Worker and user impacts belong under
Human-Centered Values in Section 5.2. What returns from here is concrete: the service life actually achieved, the
realized $N_{\text{success}}$, and how the cost divided between embodied,
operational, failure-related and revalidation components. Design uses them for
repairability, modularity and expected asset life; Model uses them to set the
efficiency targets that dynamic-inference and pruning methods are asked to
meet, always subject to the safety and reliability requirements continuing to
hold.

\textbf{Conformance} asks whether the system still satisfies the
configuration, operating-domain, evidence and authorization requirements it
runs under. It is the factor that ties operational change to the claims and
decisions permitting continued use. Assurance-continuity practice already
draws the line this factor needs, between change absorbed by maintenance and
change requiring renewed evaluation~\citep{cc_assurance_continuity}. For
Physical Intelligence the changes in question are model and software updates, hardware
replacement, recalibration, site modification, and a shift in who is exposed
to the system. Each is assessed against what the affected claim actually
depends on, not against a general category.

\emph{Authorized configuration and dependencies.} The versioned bundle $v_t$
in Eq.~\ref{eq:physical_event_record} should resolve the model, software, firmware, calibration, safety
controller, hardware, and operating domain in force. Its evidence record
should include relevant site commitments, the supplier notification
arrangements in force, and access to whatever records an incident
reconstruction would need. Notification is a support for change control, not
a substitute for it, and works only alongside configuration verification and
condition checks.

\emph{Change impact assessment.} Before a planned change goes in, identify
which claims it affects, which evidence still applies, and what tests or
authorization continued operation now needs. Unplanned change and
deterioration get the same assessment once detected, with interim
restrictions while it runs. Evidence does not have to be discarded whenever
something changes: what matters is whether its assumptions still hold. The
converse is easy to miss --- a barrier that does not depend on the policy
still needs review when its sensing coverage, timing, actuation or
operating-domain assumptions move. The indicator $\mathrm{Cov}_{\mathrm{enf}}$ in
Eq.~\ref{eq:enfcoverage} helps organize that review, though it does not quantify how much
revalidation effort or cost will follow. For a planned change the
corresponding control is the four-step update procedure of Section~\ref{sec:lifecycle_components}, with
promotion or rollback recorded against the complete configuration.

\emph{Notification and occurrence reporting.} Change notification and
incident reporting are distinct obligations. Vehicle software-update
regulation provides a sector-specific example of documented update management
and approval-related assessment~\citep{unece_r156}; aviation
occurrence-reporting arrangements provide an example of mandatory and
voluntary reporting with follow-up~\citep{eu376_2014}. These regimes inform the
proposed process without applying automatically to every Physical Intelligence
deployment. For systems within its scope, the AI Act also provides
post-market and serious-incident obligations~\citep{euaiact2024}; the applicable actor,
system category, jurisdiction, and commencement provisions must be identified
before asserting a specific duty.

Internally, the categories worth capturing are safety-filter activations,
emergency stops, near misses and departures from the operating conditions,
against defined event types and triage criteria. What leaves the
organization is a narrower set, fixed by whatever legal, contractual or
authorization threshold applies; not every logged protective activation is an
external report. The record itself follows Eq.~\ref{eq:physical_event_record}: the configuration in
force, the scenario, the autonomy or fallback mode, the proposed and executed
actions, and whether a human intervened. Who triages, who preserves
evidence, who notifies and who follows up are all settled in advance rather
than after the first occurrence.

Reporting incentives reach further than they appear to. If interventions go
systematically unreported, the unresolved-outcome count in Eq.~\ref{eq:dep_completion} and the
exposure record beside it are both biased, and the evidence base for the
whole deployment weakens with them. Reporter protection, clear event
definitions and, where appropriate, independent logging are what hold that
base in place. Conformance needs no scalar score of its own; what each
control does need is an auditable acceptance criterion under Eq.~\ref{eq:governance_contract} --- a
change assessment completed before release, say, or a report filed within its
applicable deadline.

\textbf{Closing the loop from deployment}

The three factors are not independent of Section~\ref{sec:lifecycle_components}. Each is the
standing mechanism through which one or more of the six concerns is detected
while the system operates, decided on when it matters, and returned to the
stage that can act on it. Resilience carries the concerns that present as
physical degradation inside a control period: protective separation is
watched through the separation term of Eq.~\ref{eq:deployment_alarm}, evaluated against the
conservative distance of Eq.~\ref{eq:dep_separation}, and a site condition that decays rather
than changes is watched under the threshold rule of Eq.~\ref{eq:dep_threshold}. Conformance
carries the concerns that bear on whether this is still the system that was
authorized --- whether pilot evidence applies to the fleet, whether the
workload being operated is the one assessed, whether the release margins that
supported the original decision still hold, and whether a site commitment
held by an outside party is still in force. Intervention outcomes belong to
both: Resilience supplies the intervention rate $r_{\mathrm{int}}$ that
Eq.~\ref{eq:deployment_alarm} monitors, while Conformance decides, through the reporting regime,
whether that count is complete enough for Eq.~\ref{eq:dep_completion} to mean anything.

Sustainability carries none of the six, and that is a property of how the
concerns were partitioned rather than a gap. All six ask whether an
evidential claim still holds. Sustainability asks something else --- whether
the operation can be afforded across the service life --- and so contributes
a return of its own, carrying the realized service life and the split between
embodied and operational cost back to Design and Model.

Table~\ref{tab:dep_returns} gives the resulting pathways. Feedback from deployment counts for
something only once it has produced a documented decision and an action
someone has verified, and each row instantiates Eq.~\ref{eq:governance_contract} in the ordinary way:
an owner, a versioned artifact, an acceptance criterion, the evidence, and
the trigger, with a response deadline and any interim operating restriction
added for the application. Where several parties have to act, the one
accepting the residual risk is named.

A parameter update closes a loop just as a release decision does, provided
its effect is implemented and then checked. A measurement that goes back with
no response attached leaves the claim exactly where it was. Closure means a
record of the finding, the decision taken and by whom, the action completed,
the result of verifying it, and the operating status that follows. Where the
instrumentation available cannot support the claim at all, the response is
additional evidence, alternative protection, or a restriction that can be
justified. The closure record has to show why the resulting operating status
is warranted; recording that an issue remains open is not the same as
resolving it.

These pathways are what make the life cycle of Figure~1 a loop rather than a
sequence. Section~\ref{sec:lifecycle_components} establishes what deployment can observe that no earlier
stage could; this subsection assigns each observation to the factor that owns
it and the stage that receives it. Where a return crosses the boundary of the
operating organization --- a site commitment renegotiated with its owner, an
occurrence reported to an authority --- the instruments involved are
developed in Section 6. The three factors themselves supply the
deployment-stage evidence for Robust and Safe Operation, Accountability and
Oversight, and Sustainability in Section 5, which is where the obligations
that begin as measurements here are finally expressed as values.
\begingroup
\setlength{\intextsep}{10pt}
\begin{table}[t]
\centering
\caption{Deployment feedback pathways. Each concern of
Section~\ref{sec:lifecycle_components} is carried by the factor that detects
and governs it, and returns to a named recipient. The application-specific
control contract of Eq.~\ref{eq:governance_contract} adds the acceptance
criterion, the response deadline, and the verification evidence.}
\label{tab:dep_returns}
\footnotesize
\setlength{\tabcolsep}{5pt}
\renewcommand{\arraystretch}{1.2}
\begin{tabularx}{\textwidth}{@{}>{\raggedright\arraybackslash}p{2.6cm}
                               >{\raggedright\arraybackslash}p{2.2cm}
                               >{\raggedright\arraybackslash}p{2.4cm}
                               >{\raggedright\arraybackslash}X
                               >{\raggedright\arraybackslash}X@{}}
\toprule
\textbf{Concern} & \textbf{Carried by} & \textbf{Accountable role} &
\textbf{Trigger and required record} &
\textbf{Response and closure evidence} \\
\midrule
Episodes ended by intervention & Conformance, with Resilience &
Evaluation lead &
Unresolved outcomes or unjustified sampling assumptions; updated evidence
statement &
Resolve classifications where justified or revise evaluation; document the
resulting uncertainty and claim scope \\
\addlinespace[3pt]
Evidence transfer to the fleet & Conformance &
Evaluation lead, supported by simulation lead &
Material transfer discrepancy; site- and configuration-specific comparison &
Investigate applicability; verify and record the revised use of evidence \\
\addlinespace[3pt]
Protective separation in operation & Resilience &
Safety engineer &
Unsupported parameter bounds, warning-level margin erosion, or a violation;
updated protection analysis &
Restore capability or restrict operation; verify the required separation
margin \\
\addlinespace[3pt]
Changing site conditions & Resilience and Conformance &
Operator, coordinating with site owner &
Departure from a declared condition; condition and change record &
Restore the condition or revise the operating domain; verify the affected
claims \\
\addlinespace[3pt]
Change in the deployed workload & Conformance &
Data lead, with Model &
Material exposure shift or unassessed work; updated scenario coverage &
Collect evidence and reassess scope; verify critical-cell coverage before
expansion \\
\addlinespace[3pt]
Release criteria in operation & Conformance &
Release authority &
Predeclared margin warning, negative margin, or insufficient evidence;
per-cell assessment &
Revalidate, revise, or restrict; record the evidence supporting the resulting
operating status \\
\addlinespace[3pt]
Service-life cost & Sustainability &
Asset owner, with Model &
$e_{\textrm{task}}$ (Eq.~\eqref{eq:carbon_per_task}) or $\kappa_{\textrm{task}}$
(Eq.~\eqref{eq:ktask}) outside the declared boundary, or
$\kappa^{\textrm{UCB}}_{\textrm{task}}>v_{\textrm{task}}$;
updated life-cycle account &
Revise hardware life, maintenance or efficiency targets; verify that safety
and reliability requirements still hold \\
\bottomrule
\end{tabularx}
\end{table}
\endgroup

\section{Trustworthy Physical Intelligence Framework (T-PAI)}
\label{sec:t-pai}

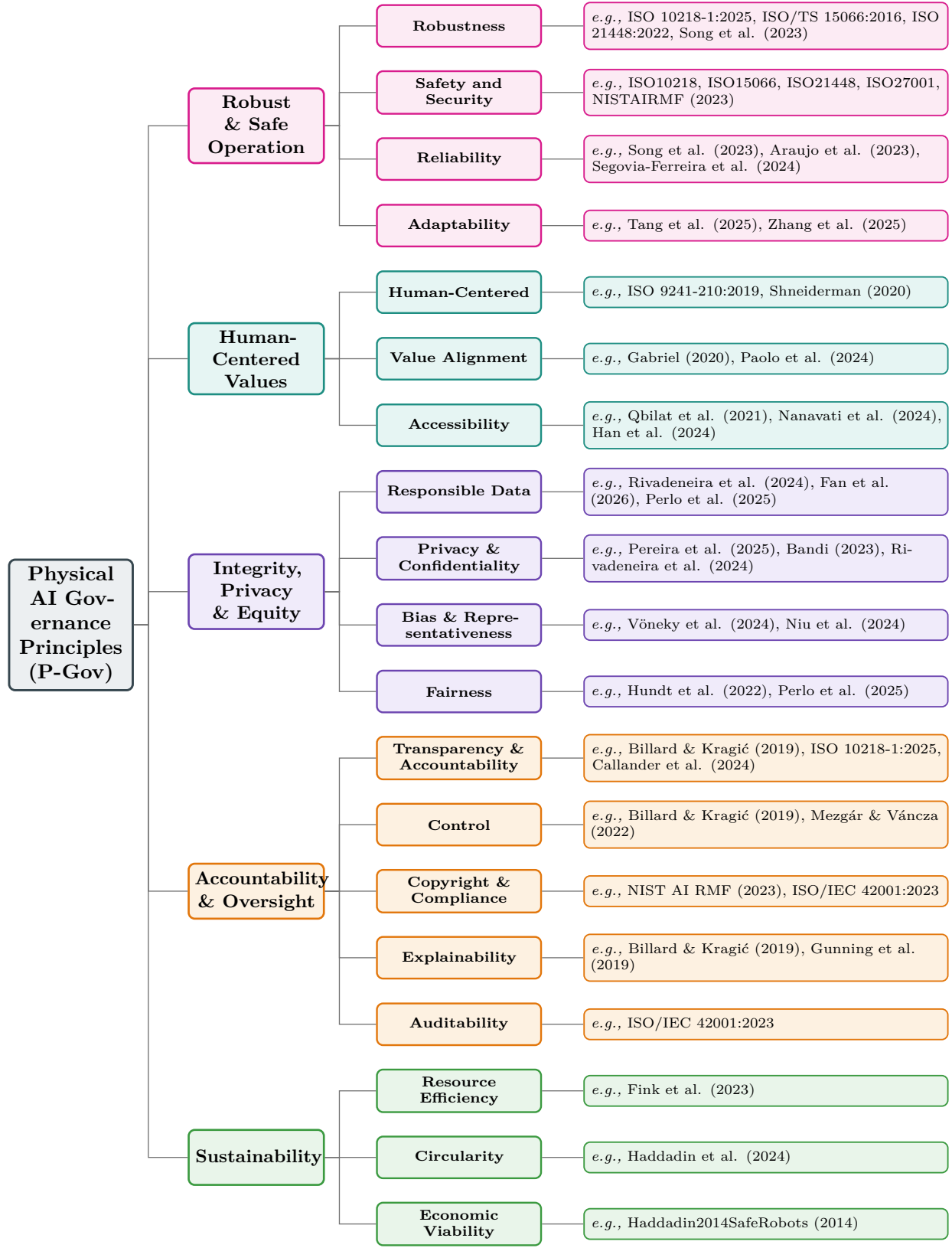
\begin{figure}[t]
    \centering
    \adjustbox{max width=\textwidth}{%
    \begin{tikzpicture}[
        font=\scriptsize,
        cat/.style={rounded corners=3pt, draw, line width=1pt, align=center, text width=2.1cm, minimum height=0.95cm, font=\small\bfseries},
        prin/.style={rounded corners=3pt, draw, line width=0.9pt, align=center, text width=2.6cm, minimum height=0.72cm, font=\scriptsize\bfseries},
        leaf/.style={rounded corners=3pt, draw, line width=0.7pt, align=left, text width=6.0cm, inner sep=4pt},
        root/.style={rounded corners=3pt, draw=grootc, line width=1.1pt, fill=grootf, align=center, text width=1.9cm, minimum height=1.2cm, font=\bfseries},
        conn/.style={draw=black!55, line width=0.5pt},
    ]
    \node[root] (root) at (1.0,-10.35) {Physical Intelligence Governance Principles (T-PAI)};

    \node[cat, draw=gpink, fill=gpinkf] (c0) at (4.2,-1.725) {Robust \& Safe Operation};
    \draw[conn] (root.east) -- ++(0.25,0) |- (c0.west);

    \node[prin, draw=gpink, fill=gpinkf] (p0) at (7.7,0.00) {Robustness};
    \node[leaf, draw=gpink, fill=gpinkf] (l0) at (13.0,0.00) {\textit{e.g.,} \customcite{ISO10218}{ISO 10218-1:2025}, \customcite{ISO15066}{ISO/TS 15066:2016}, \customcite{ISO21448}{ISO 21448:2022}, \fgcitewithyear{Song2023SIEGE}};
    \draw[conn] (c0.east) -- ++(0.25,0) |- (p0.west);
    \draw[conn] (p0.east) -- (l0.west);

    \node[prin, draw=gpink, fill=gpinkf] (p1) at (7.7,-1.15) {Safety and Security};
    \node[leaf, draw=gpink, fill=gpinkf] (l1) at (13.0,-1.15) {\textit{e.g.,} \customcite{ISO10218}{ISO 10218-1:2025}, \customcite{ISO15066}{ISO/TS 15066:2016}, \customcite{ISO21448}{ISO 21448:2022}, \customcite{ISO27001}{ISO/IEC 27001:2022}, \customcite{NISTAIRMF2023}{NISTAIRMF (2023)}};
    \draw[conn] (c0.east) -- ++(0.25,0) |- (p1.west);
    \draw[conn] (p1.east) -- (l1.west);

    \node[prin, draw=gpink, fill=gpinkf] (p2) at (7.7,-2.30) {Reliability};
    \node[leaf, draw=gpink, fill=gpinkf] (l2) at (13.0,-2.30) {\textit{e.g.,} \fgcitewithyear{Song2023SIEGE}, \fgcitewithyear{Araujo2023TVV}, \fgcitewithyear{SegoviaFerreira2024}};
    \draw[conn] (c0.east) -- ++(0.25,0) |- (p2.west);
    \draw[conn] (p2.east) -- (l2.west);

    \node[prin, draw=gpink, fill=gpinkf] (p3) at (7.7,-3.45) {Adaptability};
    \node[leaf, draw=gpink, fill=gpinkf] (l3) at (13.0,-3.45) {\textit{e.g.,} \fgcitewithyear{Tang2025SOTIFReview}, \fgcitewithyear{zhang2025contextualintegrity}};
    \draw[conn] (c0.east) -- ++(0.25,0) |- (p3.west);
    \draw[conn] (p3.east) -- (l3.west);




    \node[cat, draw=gteal, fill=gtealf] (c1) at (4.2,-5.75) {Human-Centered Values};
    \draw[conn] (root.east) -- ++(0.25,0) |- (c1.west);
    
    \node[prin, draw=gteal, fill=gtealf] (p4) at (7.7,-4.60) {Human-Centered};
    \node[leaf, draw=gteal, fill=gtealf] (l4) at (13.0,-4.60)
    {\textit{e.g.,} 
    \fgcitewithyear{Shneiderman2020HCAI},
    \hyperlink{cite.ISO9241-2102019}{\textcolor{citeblue}{ISO 9241-210}}~(\citeyear{ISO9241-2102019})};
    \draw[conn] (c1.east) -- ++(0.25,0) |- (p4.west);
    \draw[conn] (p4.east) -- (l4.west);
    
    \node[prin, draw=gteal, fill=gtealf] (p5) at (7.7,-5.75) {Human Agency \&\\Shared Autonomy};
    \node[leaf, draw=gteal, fill=gtealf] (l5) at (13.0,-5.75)
    {\textit{e.g.,} 
    \fgcitewithyear{Selvaggio2021Autonomy},
    \fgcitewithyear{Udupa2023SharedAutonomy},
    \fgcitewithyear{Collier2025Agency},
    \fgcitewithyear{Glawe2026Agency}};
    \draw[conn] (c1.east) -- ++(0.25,0) |- (p5.west);
    \draw[conn] (p5.east) -- (l5.west);
    
    \node[prin, draw=gteal, fill=gtealf] (p6) at (7.7,-6.90) {Accessibility};
    \node[leaf, draw=gteal, fill=gtealf] (l6) at (13.0,-6.90)
    {\textit{e.g.,}
    \fgcitewithyear{Qbilat2021},
    \fgcitewithyear{Nanavati2024},
    \fgcitewithyear{coDesigningAccessibleRobots2024},
    \fgcitewithyear{Stramel2019LightTouch},
    \fgcitewithyear{Santamaria2020TruST}};
    \draw[conn] (c1.east) -- ++(0.25,0) |- (p6.west);
    \draw[conn] (p6.east) -- (l6.west);
    \node[cat, draw=gpurple, fill=gpurplef] (c2) at (4.2,-9.775) {Integrity, Privacy \& Equity};
    \draw[conn] (root.east) -- ++(0.25,0) |- (c2.west);

    \node[prin, draw=gpurple, fill=gpurplef] (p7) at (7.7,-8.05) {Responsible Data};
    \node[leaf, draw=gpurple, fill=gpurplef] (l7) at (13.0,-8.05) {\textit{e.g.,} \fgcitewithyear{Mehak2024DataQuality}, \fgcitewithyear{londono2024RobotBias}};
    \draw[conn] (c2.east) -- ++(0.25,0) |- (p7.west);
    \draw[conn] (p7.east) -- (l7.west);

    \node[prin, draw=gpurple, fill=gpurplef] (p8) at (7.7,-9.20) {Privacy \& Confidentiality};
    \node[leaf, draw=gpurple, fill=gpurplef] (l8) at (13.0,-9.20) {\textit{e.g.,} \fgcitewithyear{Callander2024PrivacyConscious}, \fgcitewithyear{bandi2023taxonomy}, \fgcitewithyear{fan2026privacy}};
    \draw[conn] (c2.east) -- ++(0.25,0) |- (p8.west);
    \draw[conn] (p8.east) -- (l8.west);

    \node[prin, draw=gpurple, fill=gpurplef] (p9) at (7.7,-10.35) {Bias \& Representativeness};
    \node[leaf, draw=gpurple, fill=gpurplef] (l9) at (13.0,-10.35) {\textit{e.g.,} \fgcitewithyear{londono2024RobotBias}, \fgcitewithyear{Niu2024CrossDomainTransfer}};
    \draw[conn] (c2.east) -- ++(0.25,0) |- (p9.west);
    \draw[conn] (p9.east) -- (l9.west);

    \node[prin, draw=gpurple, fill=gpurplef] (p10) at (7.7,-11.50) {Fairness};
    \node[leaf, draw=gpurple, fill=gpurplef] (l10) at (13.0,-11.50) {\textit{e.g.,} \fgcitewithyear{londono2024RobotBias}, \fgcitewithyear{hundt2022robots}};
    \draw[conn] (c2.east) -- ++(0.25,0) |- (p10.west);
    \draw[conn] (p10.east) -- (l10.west);

    \node[cat, draw=gorange, fill=gorangef] (c3) at (4.2,-14.95) {Accountability \& Oversight};
    \draw[conn] (root.east) -- ++(0.25,0) |- (c3.west);

    \node[prin, draw=gorange, fill=gorangef] (p11) at (7.7,-12.65) {Transparency \& Accountability};
    \node[leaf, draw=gorange, fill=gorangef] (l11) at (13.0,-12.65) {\textit{e.g.,} \fgcitewithyear{takeda2020accountable}, \fgcitewithyear{ieee7001}, \fgcitewithyear{Callander2024PrivacyConscious}};
    \draw[conn] (c3.east) -- ++(0.25,0) |- (p11.west);
    \draw[conn] (p11.east) -- (l11.west);

    \node[prin, draw=gorange, fill=gorangef] (p12) at (7.7,-13.80) {Control};
    \node[leaf, draw=gorange, fill=gorangef] (l12) at (13.0,-13.80) {\textit{e.g.,} \fgcitewithyear{mezgar2022ethics}};
    \draw[conn] (c3.east) -- ++(0.25,0) |- (p12.west);
    \draw[conn] (p12.east) -- (l12.west);

    \node[prin, draw=gorange, fill=gorangef] (p13) at (7.7,-14.95) {Copyright \& Compliance};
    \node[leaf, draw=gorange, fill=gorangef] (l13) at (13.0,-14.95) {\textit{e.g.,} \hyperlink{cite.NISTAIRMF2023}{\textcolor{citeblue}{NIST}}~(\citeyear{NISTAIRMF2023}), \hyperlink{cite.ISO42001}{\textcolor{citeblue}{ISO}}~(\citeyear{ISO42001}), \hyperlink{cite.ISO10218}{\textcolor{citeblue}{ISO}}~(\citeyear{ISO10218})};
    \draw[conn] (c3.east) -- ++(0.25,0) |- (p13.west);
    \draw[conn] (p13.east) -- (l13.west);

    \node[prin, draw=gorange, fill=gorangef] (p14) at (7.7,-16.10) {Explainability};
    \node[leaf, draw=gorange, fill=gorangef] (l14) at (13.0,-16.10) {\textit{e.g.,} \fgcitewithyear{gunning2019xai}, \fgcitewithyear{takeda2020accountable}};
    \draw[conn] (c3.east) -- ++(0.25,0) |- (p14.west);
    \draw[conn] (p14.east) -- (l14.west);

    \node[prin, draw=gorange, fill=gorangef] (p15) at (7.7,-17.25) {Auditability};
    \node[leaf, draw=gorange, fill=gorangef] (l15) at (13.0,-17.25) {\textit{e.g.,} \hyperlink{cite.ISO42001}{\textcolor{citeblue}{ISO}}~(\citeyear{ISO42001}), \fgcitewithyear{winfield2017ethical}};
    \draw[conn] (c3.east) -- ++(0.25,0) |- (p15.west);
    \draw[conn] (p15.east) -- (l15.west);

    \node[cat, draw=ggreen, fill=ggreenf] (c4) at (4.2,-19.55) {Sustainability};
    \draw[conn] (root.east) -- ++(0.25,0) |- (c4.west);

    \node[prin, draw=ggreen, fill=ggreenf] (p16) at (7.7,-18.40) {Resource Efficiency};
    \node[leaf, draw=ggreen, fill=ggreenf] (l16) at (13.0,-18.40) {\textit{e.g.,} \fgcitewithyear{Fink2023GreenRobotics}};
    \draw[conn] (c4.east) -- ++(0.25,0) |- (p16.west);
    \draw[conn] (p16.east) -- (l16.west);

    \node[prin, draw=ggreen, fill=ggreenf] (p17) at (7.7,-19.55) {Circularity};
    \node[leaf, draw=ggreen, fill=ggreenf] (l17) at (13.0,-19.55) {\textit{e.g.,} \fgcitewithyear{Haddadin2024SustainableRobotics}};
    \draw[conn] (c4.east) -- ++(0.25,0) |- (p17.west);
    \draw[conn] (p17.east) -- (l17.west);

    \node[prin, draw=ggreen, fill=ggreenf] (p18) at (7.7,-20.70) {Economic Viability};
    \node[leaf, draw=ggreen, fill=ggreenf] (l18) at (13.0,-20.70) {\textit{e.g.,} \fgcitewithyear{Haddadin2014SafeRobots}, \customcite{iec60300_3_3}{IEC 60300-3-3:2017}};
    \draw[conn] (c4.east) -- ++(0.25,0) |- (p18.west);
    \draw[conn] (p18.east) -- (l18.west);

    \end{tikzpicture}}
    \caption{Taxonomy of Physical Intelligence governance principles (T-PAI). Nineteen governance sub-components are grouped into five categories, with each sub-component anchored by representative standards, frameworks, and literature.}
    \label{fig:gov_taxonomy}
\end{figure}

Traditional AI \citep{li2023trustworthy, cai2026trustworthy} and Physical Intelligence differ in three fundamental aspects, giving rise to distinct governance elements, priorities, and oversight mechanisms. First, Physical Intelligence operates in consequence-driven environments where failures have immediate real-world impacts, making governance of responsibility, intervention, and risk management a fundamental requirement \citep{wef2026infrastructure}. Second, Physical Intelligence rarely operates autonomously in isolation; instead, it collaborates with drivers, clinicians, factory workers, consumers, and other human stakeholders through shared tasks, decision-making, and control \citep{Hentout2019}. Third, Physical Intelligence is harder to bound and can fail in entirely novel ways that were not predicted, which pushes governance toward real-time monitoring and containment rather than after-the-fact review \citep{susarla2026factory}. As for risks, Physical Intelligence systems introduce four broad categories: physical, informational, economic, and social. Physical safety dominates governance concerns for embodied AI systems, particularly classical robots and autonomous vehicles, whereas virtual agents are more closely associated with informational risks such as misinformation. Regardless of AI modality, fairness, accountability, and transparency remain central governance challenges, while economic and broader societal risks are comparatively underexplored \citep{perlo2025embodied}. Our survey focuses on five key aspects of Physical Intelligence and its sub-components (Figure~\ref{fig:gov_taxonomy}). 

\subsection{Robust \& Safe Operation}

Robust and Safe Operation ensures that Physical Intelligence systems perform their intended functions safely, reliably, and adaptively throughout their operational lifetime. Safety and security prevent accidental harm and protect against malicious threats \citep{li2026safety}. Reliability emphasizes dependable and consistent system performance \citep{Hou2026WorldModel}. Adaptability enables systems to respond effectively to dynamic environments, unforeseen conditions, and evolving tasks \citep{xiang2026aligning}. These components provide the technical foundation for trustworthy Physical Intelligence as an infrastructure layer by ensuring resilient operation while minimizing risks to humans, physical assets, and the environment.

\textbf{Robustness} ensures that a Physical Intelligence system maintains control stability and state invariants under operational noise, sensor drift, and mechanical disturbances \citep{Song2023SIEGE}. Unlike traditional AI governance, which evaluates robustness as static offline accuracy on test datasets, Physical Intelligence governance operationalizes robustness through continuous runtime monitoring and closed-loop control guardrails. This active monitoring continuously tracks state estimation errors and environmental forces in real time, dynamically triggering safety shields or feedback corrections before physical perturbations escalate into hazardous failures \citep{ISO21448,Lee2025CyberPhysicalAI}.

\textbf{Safety \& Security} governs both the physical hazards and cyber vulnerabilities inherent to a Physical Intelligence system. It integrates \emph{Physical Safety}, which enforces dynamic limits on kinematic motion, impact forces, and workspace boundaries to prevent collisions and physical injury, with \emph{Cyber Security}, which protects sensor feeds, network streams, and control policies against data poisoning, signal spoofing, or unauthorized remote override \citep{ISO10218,ISO15066,ISO21448,ISO27001,NISTAIRMF2023}. This governance factor establishes runtime safety filters and authenticated communication protocols that prevent unsafe motor commands while shielding autonomous operation from adversarial compromise \citep{Lee2025CyberPhysicalAI}.

\textbf{Reliability} refers to the ability of a Physical Intelligence system to consistently perform its intended functions under expected operating conditions while avoiding unexpected operational failures \citep{Song2023SIEGE}. Systematic evaluation requires standardized benchmarks, hardware-in-the-loop testing, and runtime health monitoring across the system life cycle \citep{Araujo2023TVV}. To sustain dependable operation over extended service lives, governance should promote resilient architectures capable of anomaly detection, graceful degradation, and fault recovery when encountering internal component wear or subsystem failure \citep{SegoviaFerreira2024}.

\textbf{Adaptability} governs a Physical Intelligence system’s capacity to modify its internal models, control policies, or execution strategies when encountering distribution shifts, new tasks, or changing physical dynamics without offline retraining. While robustness focuses on rejecting transient noise within a static operational baseline, adaptability provides policy plasticity, i.e. enabling online system identification, few-shot generalization, and continual learning across shifting operational contexts \citep{Tang2025SOTIFReview}. Governance must evaluate runtime adaptation mechanisms to ensure that online model shifts remain bounded, physically consistent, and strictly aligned with core safety limits \citep{Zhang2025EmbodiedReview}.

\subsection{Human-Centered Values}

Physical Intelligence is transforming autonomous systems from task-executing machines into context-aware collaborators that adapt to dynamic environments, augment human capabilities, and enable more intelligent interactions across diverse real-world domains. As these systems increasingly operate in shared spaces, workplaces, homes, and care settings, human-centeredness becomes not only an interface-design consideration but also a system-level governance principle \citep{Singh2026HRT}.  Advances in multimodal perception, adaptive control, and machine learning are shifting robotics from rigid automation toward human-aware, context-sensitive collaboration in unstructured environments \citep{Zhao2025MPDDM}.  Advances in Physical Intelligence are enabling robots to perceive, interpret, and respond to human emotions, facilitating more natural, socially intelligent interactions \citep{Duncan2024Multimodal}.  At the same time, these capabilities raise important questions about human agency, accessibility, and the appropriate allocation of control between people and autonomous systems.

\textbf{Human-Centered} denotes a design and governance philosophy in which technology adapts to people rather than requiring people to adapt to technology \citep{Shneiderman2020HCAI}. For Physical Intelligence, this philosophy entails two central commitments. First, systems should preserve human agency by supporting oversight, intervention, and choice rather than treating maximum autonomy as an end in itself. Automation can take responsibility for appropriate sub-tasks while humans retain authority over goals, preferences, and consequential decisions. Second, Physical Intelligence should be designed around human purposes, capabilities, and limitations. Human factors, ergonomics, and usability principles should therefore guide system development so that embodied AI improves effectiveness and efficiency while supporting well-being, accessibility, safety, and long-term usability \citep{ISO9241-2102019}.

\textbf{Human Agency and Shared Autonomy} concern how decision authority is distributed between humans and intelligent machines. In physical human--robot interaction, shared autonomy provides an alternative to the binary choice between manual control and full automation by allowing the level of robotic assistance to vary according to the task, context, and user's capabilities \citep{Selvaggio2021Autonomy,Udupa2023SharedAutonomy}. Preserving agency is important because better task performance does not necessarily imply a better human experience. Recent experiments with assistive robot manipulation found that increasing autonomous assistance improved task performance but could reduce users' sense of agency, revealing a trade-off between operational efficiency and perceived control \citep{Collier2025Agency}. More broadly, empirical HRI research increasingly treats human autonomy and sense of agency as distinct but closely related human values that should be explicitly evaluated when designing adaptive robotic systems \citep{Glawe2026Agency}. Physical Intelligence should therefore optimize not only what a system can accomplish, but also whether users remain able to understand, influence, and meaningfully participate in its actions.

\textbf{Accessibility} is the requirement that AI systems be usable by people with a wide range of abilities, including those with disabilities \citep{Qbilat2021, Nanavati2024}. Accessibility ensures that Physical Intelligence systems are designed with inclusive interfaces, adaptive autonomy, and user-centered interactions that accommodate diverse abilities, enabling people with disabilities to safely, effectively, and independently interact with embodied AI \citep{Nanavati2024}. Accessibility should be embedded from the earliest stages of Physical Intelligence development through co-design with people with disabilities. By actively involving end users, developers can create products that better address diverse user needs, improve usability, and promote equitable access \citep{coDesigningAccessibleRobots2024}.

Assistive robotics also illustrates how these principles can be translated into the design of physical systems. Stramel et al. developed a person-following robotic device that provides light-touch proprioceptive feedback during walking. This system allows the robot to maintain stability while remaining responsive to user motion \citep{Stramel2019LightTouch}. Similarly, the Trunk-Support-Trainer developed by Santamaria et al. and Agrawal adapts its force feedback to an individual's measured postural control limits. This system provides assistance according to the user's functional capabilities rather than imposing a fixed control strategy \citep{Santamaria2020TruST}. These examples demonstrate that accessibility in Physical Intelligence is not limited to interface compliance. It also involves adapting the physical behavior of the system to the capabilities of the person interacting with it.

\textbf{Personalization and Empowerment} extend accessibility beyond accommodating static user differences toward adapting to changes in ability, preference, and context over time. In Physical Intelligence, personalization usually involves adjusting the level of autonomy, interaction modality, physical assistance, task difficulty, or feedback strategy. However, adaptation should remain interpretable and controllable by the user. Systems that optimize assistance without considering perceived control may reduce agency or encourage over-reliance \citep{Collier2025Agency,Glawe2026Agency}. Human-centered Physical Intelligence should keep humans safe while preserving their ability to choose, intervene, and develop their own capabilities.

\subsection{Integrity, Privacy and Equity}

Integrity, Privacy, and Equity guide the responsible governance of Physical Intelligence systems and their data. Integrity ensures that the multimodal data used by Physical Intelligence systems are authentic, accurate, complete, and consistently maintained throughout sensing, transmission, storage, and learning. Maintaining data integrity is critical because corrupted or low-quality sensor data can directly compromise perception, planning, and safe interaction with the physical world \citep{Mehak2024DataQuality}. Privacy requires Physical Intelligence systems to protect sensitive multimodal information collected through cameras, microphones, wearable sensors, and other embodied sensing devices through appropriate controls over data collection, access, use, and disclosure. Physical Intelligence operates in real-world environments, and therefore privacy extends beyond data protection to include physical, psychological, and social dimensions of human interaction~\citep{Callander2024PrivacyConscious}. Equity assessment in learning-based Physical Intelligence should consider both data representativeness and differences in system outcomes across relevant individuals and groups. Representative training data can help mitigate bias, but do not by themselves establish fair outcomes \citep{londono2024RobotBias}.

\textbf{Responsible Data} concerns the quality, provenance, and stewardship of data throughout the Physical Intelligence life cycle. Data quality assurance should assess accuracy, completeness, consistency, and timeliness in relation to the intended task \citep{Mehak2024DataQuality}. Data provenance and processing histories should be documented to support traceability, as discussed in Section~\ref{sec:data_trustworthiness}. Training and evaluation datasets should cover relevant populations and deployment conditions. Such coverage can support bias mitigation, but fairness also requires assessment of model behavior and outcomes \citep{londono2024RobotBias}.
 
\textbf{Privacy and Confidentiality} concern the protection of personal information and the management of privacy risks arising from sensing and interaction. In human--robot interaction, privacy concerns include intrusions into personal space and the collection of sensitive behavioral information, extending beyond unauthorized access to stored data \citep{Callander2024PrivacyConscious}. Assessment should cover training and operational data, including information about users and bystanders, as well as sensitive inferences recoverable from derived representations or action logs \citep{perlo2025embodied,fan2026privacy}. Governance should specify appropriate data collection and use, access controls, retention, and deletion. Device authentication, encryption, and secure communication provide supporting safeguards, whose implementation should account for resource constraints \citep{Callander2024PrivacyConscious,bandi2023taxonomy}. Privacy-preserving architectures should be evaluated for residual privacy risks and their effects on task performance and safety \citep{rivadeneira2024unified,fan2026privacy}.
  
\textbf{Bias and Representativeness} should be assessed by examining dataset coverage, measurement choices, and model behavior across relevant populations and deployment contexts. Unrepresentative data and biased measurement or modeling choices can produce disparities in system outcomes; evaluation should therefore consider relevant individuals and groups rather than aggregate performance alone \citep{londono2024RobotBias}. Generalization across deployment contexts is a separate evaluation question. Differences between source domains, such as simulation or laboratory environments, and target deployment domains can impair policy performance. Cross-domain policy transfer methods aim to mitigate environmental and embodiment gaps, particularly when target-domain data collection is costly or constrained by safety requirements \citep{Niu2024CrossDomainTransfer}. 
 
\textbf{Fairness} concerns whether system decisions and actions unjustifiably disadvantage particular individuals or groups. Assessment should use criteria appropriate to the task and deployment context \citep{londono2024RobotBias}. In a simulation audit, a CLIP-based robotic manipulation policy exhibited racial and gender biases when selecting blocks bearing facial photographs. When instructed to select a ``criminal'' block, the policy acted on an unsupported association between appearance and criminality. This example highlights the need to evaluate both disparities in action selection and whether the requested task is appropriate; equalizing selection rates would not make appearance-based judgments of criminality valid~\citep{hundt2022robots}. More broadly, language and foundation models are increasingly being integrated into robot decision pipelines, including systems that translate demonstrations and language instructions into executable robot task code~\citep{wang2023demo2code}. This creates a pathway through which biases present in upstream AI components may influence downstream robot decisions and actions. Prior work has documented substantial intersectional disparities in facial-analysis systems~\citep{buolamwini2018gender} as well as cultural and representational biases in text and image generators~\citep{tao2024cultural,abu2026between}. Fairness evaluation for Physical Intelligence should therefore consider not only end-to-end action outcomes, but also biases introduced by upstream perception, language, and foundation-model components. Performance thresholds such as Eq.~\ref{eq:fairness_gate} provide operational checks, but do not by themselves establish fairness or the appropriateness of the task.

\subsection{Accountability \& Oversight} 

Accountability and oversight ensure that humans remain responsible for the behavior and consequences of Physical Intelligence systems by enabling effective supervision, intervention, and governance throughout the system life cycle. Unlike traditional AI, Physical Intelligence operates autonomously in the physical world and directly interacts with people and infrastructure, making continuous human oversight essential for preventing unsafe behaviors, assigning responsibility, and maintaining public trust \citep{takeda2020accountable,mezgar2022ethics}.

\textbf{Transparency \& Accountability} requires that the perception, reasoning, planning, and physical actions of Physical Intelligence systems are understandable and attributable to developers, operators, regulators, and affected users. Unlike traditional AI, transparency in Physical Intelligence extends beyond algorithmic decisions to include sensor observations, environmental representations, motion planning, control policies, hardware capabilities, and operational constraints that influence real-world behavior. Comprehensive documentation of system design, training data provenance, embodiment assumptions, safety constraints, and deployment conditions enables stakeholders to understand why a system performed a particular action, assign responsibility for its outcomes, and support effective human oversight throughout the Physical Intelligence life cycle \citep{billard2019trends,ISO10218,Callander2024PrivacyConscious}.

\textbf{Control} requires Physical Intelligence systems to incorporate mechanisms that enable humans to monitor, guide, intervene in, override, or safely deactivate system operations whenever necessary. Physical Intelligence continuously interacts with dynamic environments. Control mechanisms must support both routine supervision and emergency intervention while maintaining stable and reliable system behavior. Human involvement may range from direct teleoperation and shared autonomy to supervisory control over highly autonomous systems, depending on the deployment context and associated risks \citep{billard2019trends}. Adaptive control is essential for ensuring stable and reliable operation in both deterministic and continuous-time systems, as well as stochastic discrete-time systems. There are five levels of control in Physical Intelligence, representing a continuum from fully human-controlled systems to fully autonomous operation. These levels range from human control, AI-assisted control, AI-advisory control, and AI-collaborative control, to AI-active independent control, where humans transition from direct decision-makers to supervisory roles \citep{mezgar2022ethics}.

\textbf{Copyright \& Compliance} ensures that Physical Intelligence systems satisfy applicable legal, regulatory, safety, and intellectual property requirements throughout their life cycle. Beyond complying with AI governance objectives, Physical Intelligence must also conform to robotics safety standards, machinery regulations, functional safety requirements, and copyright obligations associated with training data, software, digital assets, and foundation models. Physical Intelligence directly interacts with humans and the physical environment. Compliance additionally requires verifying that sensing, planning, control, and actuation behaviors satisfy safety certification, operational constraints, and continuous risk management before and during deployment \citep{NISTAIRMF2023}.

\textbf{Explainability} is the capability of Physical Intelligence systems to provide human-understandable explanations for their perceptions, reasoning processes, planning decisions, and executed physical actions. Unlike traditional AI, where explanations often focus on prediction outcomes, Physical Intelligence must explain how multimodal sensor observations, environmental context, motion planning, task objectives, and safety constraints jointly contributed to a specific physical behavior. Such explanations improve operator trust, facilitate debugging and failure diagnosis, support safe human--robot collaboration, and provide evidence for regulatory review and accident investigation \citep{billard2019trends, gunning2019xai}.

\textbf{Auditability} is the capability to trace, reconstruct, verify, and evaluate the behavior of Physical Intelligence systems through comprehensive records of data provenance, model development, system configuration, sensor observations, world-state estimates, planning decisions, control commands, and execution logs. Audit trails must capture both computational decisions and physical system behaviors to enable failure analysis, responsibility attribution, compliance verification, and continuous safety improvement. Comprehensive logging and standardized documentation therefore support post-incident investigation, regulatory compliance, and the trustworthy deployment of embodied AI systems throughout their operational life cycle \citep{ISO42001}.

\subsection{Sustainability}

Sustainability is the principle that Physical Intelligence systems should be designed, developed, deployed, and retired in ways that minimize environmental impacts while maximizing the efficient use of resources throughout their life cycle. Sustainable governance extends beyond energy-efficient AI models to include responsible material use, hardware longevity, maintenance, repairability, reuse, and end-of-life management, because Physical Intelligence combines computational intelligence with physical hardware. This ensures that embodied AI systems deliver long-term environmental, economic, and societal value while reducing resource consumption and waste \citep{Fink2023GreenRobotics,Haddadin2024SustainableRobotics}.

\textbf{Resource Efficiency} requires Physical Intelligence systems to minimize the consumption of computational, energy, and material resources throughout their life cycle while maintaining reliable performance. Since embodied AI relies on energy-intensive training, onboard computation, sensors, actuators, and physical hardware, improving resource efficiency through energy-aware algorithms, efficient hardware design, adaptive computation, and optimized task execution is essential for reducing operational costs and environmental impact \citep{Fink2023GreenRobotics}. Computational resources can be managed through anytime algorithms that trade computation time for solution quality~\citep{zilberstein1996using}, and through allocation methods that account for dependencies between robotic algorithms when optimizing expected system utility under a shared computation budget~\citep{nakamura2023formal}.

\textbf{Circularity} promotes designing Physical Intelligence systems for longevity, modularity, repairability, reuse, remanufacturing, and recycling to reduce material waste and extend hardware lifespans. Unlike traditional AI, Physical Intelligence depends on physical components that require responsible life cycle management. Adopting circular design principles enables efficient resource utilization, minimizes electronic waste, and supports the environmentally sustainable deployment of robotic and embodied AI systems \citep{Haddadin2024SustainableRobotics}.

\textbf{Economic Viability} covers a system's ability to justify its costs over its operational lifetime, accounting for hardware acquisition, maintenance, energy consumption, and eventual decommissioning, given the substantially higher capital and operating costs of physical systems relative to purely digital AI. In traditional AI governance, economic considerations are largely confined to computational cost, licensing, and cloud infrastructure spending, since the marginal cost of deploying an additional model instance is comparatively low. Physical Intelligence governance elevates economic viability to a first-order design constraint: robots and embodied systems carry substantial fixed costs in manufacturing, sensors, actuators, and physical maintenance, and their return on investment depends on factors largely absent from digital AI, such as physical durability, repairability, and the total cost of ownership across the hardware's operational lifespan, which life-cycle costing treats jointly with the costs of dependability ~\citep{iec60300_3_3,Haddadin2014SafeRobots}.
We make this operational through the levelized cost per successful task $\kappa_{\textrm{task}}$ in Eq.~\eqref{eq:ktask}, that is evaluated at conservative success rate and compared to the value of a completed task. Economic benchmarks are already available for physical agents: converting simulated collision, energy and delivery outcomes into costs, revenue and reports margin and break-even points, with the baseline results showing that task success metrics do not establish commercial viability, but collision related costs being most prominet in the per-run account ~\cite{seong2025costnav}. The goal of this governance factor is to ensure that Physical Intelligence deployments remain economically sustainable throughout their life cycle, rather than evaluated only against upfront development and deployment costs.

\section{Future Considerations}
\label{sec:future}
The trajectory of Physical Intelligence governance is moving away from fragmented, voluntary guidance and toward layered frameworks that combine technical standards, sector-specific regulation, and adaptive oversight mechanisms.

\textbf{Layered Framework with Metrics} A layered governance framework enables different stakeholders---including researchers, developers, operators, regulators, and end users---to understand their respective responsibilities and governance requirements within specific application scenarios. To support continuous improvement and accountability, governance should also incorporate measurable metrics and quantitative evaluation methods that assess safety, reliability, robustness, and human-centered outcomes throughout the life cycle \citep{ras2021trustworthy}. This layered structure mirrors the three-tier life-cycle framework introduced in Section~\ref{sec:epalgo}. Examples of such metrics include observation age at execution~\citep{yates2021age}, hazard-detection coverage under specified sensor failures, the time available for a protective response, and the levelized cost per successful task. Such measures can establish whether additional sensing viewpoints reduce hazardous failures enough to warrant their processing, energy, calibration, and maintenance costs~\citep{arnold2022cooperative}.

\textbf{Enforcement} Physical Intelligence governance should extend beyond voluntary ethical guidelines toward enforceable instruments. Three complementary mechanisms are emerging: binding regulation, such as the EU AI Act's risk-based obligations for high-risk systems~\citep{euaiact2024}; technical standards that translate legal requirements into testable criteria, such as ISO 10218-1:2025 for industrial robot safety~\citep{ISO10218} and ISO/IEC 42001:2023 for AI management systems~\citep{ISO42001}; and third-party certification and conformity assessment for systems that continue to learn after deployment. Making these instruments effective requires a clear division of labor---researchers identifying emerging risks, standards bodies developing technical specifications, and regulators establishing enforceable frameworks~\citep{mezgar2022ethics}. Scoring methodologies can be applied to quantify and guide the enforcement process \citep{cai2026trustworthy}.

\textbf{Adaptive Governance} Physical Intelligence systems continuously learn, interact with changing environments, and may receive model or software updates after deployment. Governance should therefore evolve from static compliance toward adaptive governance that supports continuous risk assessment, monitoring, and iterative updates throughout the system life cycle rather than one-time certification \citep{NISTAIRMF2023}.

\textbf{Multi-Agent Governance} Extending T-PAIO to interacting physical agents requires evaluating shared dependencies and unsafe interactions~\citep{Avizienis2004,leveson2011,garg2024learning}, information freshness~\citep{yates2021age}, and communication security~\citep{neupane2024security,gil2017guaranteeing}. Governance should assess how communication shapes joint actions~\citep{goldfajn2026robotalk}, how compromised messages can propagate unsafe actions~\citep{huang2026coordination}, and whether coordinated behavior serves the intended shared objective~\citep{nakamura2026colosseum}. Future work should evaluate team safety under communication delays~\citep{ballotta2025safe} and reassess dependencies after updates~\citep{sculley2015hidden}. Building on established responsibilities and ongoing review~\citep{NISTAIRMF2023}, the control contract of Eq.~\ref{eq:governance_contract} could specify team-level acceptance criteria and evidence linking messages to executed actions throughout the life cycle.
 
\section{Conclusion}
\label{sec:conclusion}

As Physical Intelligence moves AI beyond digital environments into physical world, its governance requires a life-cycle-wide approach that accounts for physical safety, human interaction, system reliability, and societal impact.  The paper establishes an integrated foundation for trustworthy Physical Intelligence through four contributions: characterizing its core capabilities and the role of physical principles in AI; examining how failures emerge, propagate, and cascade across interconnected systems; introducing the Trustworthy Physical Intelligence Operationalization (T-PAIO) framework to structure governance practices across five life-cycle stages—Research, Design, Data, Model, and Deployment; and developing the Trustworthy Physical Intelligence (T-PAI) framework, comprising five fundamental principles and nineteen sub-components, to organize the key dimensions of trustworthiness. Future research should prioritize real-world validation, measurable assurance indicators, and continuous, adaptive governance. Ultimately, trustworthy Physical Intelligence requires embedding safety, reliability, human values, accountability, and sustainability throughout the system life cycle.

{\small \noindent \raggedright \textbf{Affiliation Disclaimer:} The organizations listed in this paper are provided solely to identify the institutional affiliations of the authors. The views and opinions expressed in the paper do not necessarily represent the views, opinions, or official positions of the respective organizations. \par}

\newpage

\bibliographystyle{unsrtnat}
\bibliography{example_paper}

\newpage

\end{document}